\documentclass[12pt]{amsart}
\usepackage{amsmath, amssymb, amsthm, amsfonts, enumerate, verbatim, mathtools,bbm,esint,bm}
\usepackage[normalem]{ulem}
\usepackage{pgf,tikz,pgfplots}
\usepackage{tikz-3dplot}
\usetikzlibrary{shapes,decorations,arrows,calc,arrows.meta,fit,positioning,intersections}

\usepackage{physics}
\usepackage{ifthen}
\usepackage[outline]{contour} 
\usetikzlibrary{angles,quotes} 
\usetikzlibrary{patterns}
\tikzset{>=latex} 
\contourlength{1.2pt}
\colorlet{myred}{red!65!black}
\tikzstyle{ground}=[preaction={fill,top color=black!10,bottom color=black!5,shading angle=20},
fill,pattern=north east lines,draw=none,minimum width=0.3,minimum height=0.6]
\tikzstyle{mass}=[line width=0.6,red!30!black,fill=red!40!black!10,rounded corners=1,
top color=red!40!black!20,bottom color=red!40!black!10,shading angle=20]
\tikzstyle{rope}=[brown!70!black,line width=1.2,line cap=round] 

\tikzstyle{force}=[->,myred,thick,line cap=round]
\tikzstyle{Fproj}=[force,myred!40]

\newboolean{showforces}
\setboolean{showforces}{true}

\usetikzlibrary{datavisualization}
\usetikzlibrary{datavisualization.formats.functions}

\tikzset{
	state/.style ={ellipse, draw, minimum width = 0.7 cm}
}
\usepackage[justification=centering]{caption}
\pgfplotsset{compat=1.15}
\usepackage{mathrsfs}
\usetikzlibrary{hobby,arrows.meta,3d}
\usepackage[all]{xy}
\usepackage{tikz}
\usepackage{subfig}
\usepackage{graphicx}
\usepackage{xcolor, soul, todonotes}
\usepackage[hypertexnames=false,debug,linktocpage=true,hidelinks]{hyperref}

\hypersetup{
	colorlinks,
	linktoc=all,
	linkcolor={blue},
	citecolor={red},
	urlcolor={blue}
}
\tikzstyle{punkt}=[circle, fill=black, minimum size=1mm,inner sep=0pt, draw]

\newtheorem{Theorem}{Theorem}[section]
\newtheorem{Lemma}[Theorem]{Lemma}
\newtheorem{Corollary}[Theorem]{Corollary}
\newtheorem{Proposition}[Theorem]{Proposition}
\newtheorem{Remark}[Theorem]{Remark}

\let\epsilon\varepsilon
\let\kappa=\varkappa
\def\pnt{{\raise0.5mm\hbox{\large\bf.}}}

\newif\ifistoreview
\istoreviewtrue

\newcommand{\Supp}{\mathrm{Supp}}

\renewcommand{\P}{\mathbb{P}}
\renewcommand{\d}{\mathrm{d}}
\renewcommand{\div}{\mathrm{div}}
\newcommand{\TV}{\mathrm{TV}}
\newcommand{\oo}{\overline}
\renewcommand{\phi}{\varphi}
\newcommand{\x}{\mathbf{x}}
\newcommand{\X}{\mathbf{X}}
\newcommand{\W}{\mathbf{W}}
\newcommand{\w}{\mathbf{w}}

\newcommand{\rond}{\partial}
\newcommand{\PIEV}{\mathcal{P}\mathcal{I}\mathcal{E}\mathcal{V}}
\newcommand{\PEACE}{\mathbb{P}\mathbb{E}\mathbb{A}\mathbb{C}\mathbb{E}}
\newcommand{\bd}{\mathrm{Bd}}
\newcommand{\inte}{\mathrm{Int}}
\newcommand{\Z}{\mathbf{Z}}
\newcommand{\z}{\mathbf{z}}
\newcommand{\E}{\mathbb{E}}

\begin{document}

	\title{Probabilistic Easy Variational Causal Effect}
\author[Usef Faghihi]{Usef Faghihi\,*}
\address{Usef Faghihi, Department of Mathematics and Computer Science, The University of Quebec at Trois-Rivieres, 3351 Bd des Forges, Trois-Rivières, QC G8Z 4M3, Canada}
\author[Amir Saki]{Amir Saki}
\address{Amir Saki, Department of Mathematics and Computer Science, The University of Quebec at Trois-Rivieres, 3351 Bd des Forges, Trois-Rivières, QC G8Z 4M3, Canada}
\email{amir.saki@uqtr.ca, amir.saki.math@gmail.com}

	
\begin{abstract}
Let $\X$ and $\Z$ be random vectors, and $Y=g(\X,\Z)$. In this paper, on the one hand, for the case that $\X$ and $\Z$ are continuous, by using the ideas from the total variation and the flux of $g$, we develop a point of view in causal inference capable of dealing with a broad domain of causal problems. Indeed, we focus on a function, called \textbf{P}robabilistic \textbf{E}asy V\textbf{a}riationoal \textbf{C}ausal \textbf{E}ffect (PEACE), which can measure the direct causal effect of $\X$ on $Y$ with respect to continuously  and interventionally changing the values of $\X$ while keeping the value of $\Z$ constant. PEACE is  a function of $d\ge 0$, which is a degree managing the strengths of probability density values $f(\x|\z)$. On the other hand, we generalize the above idea for the discrete case and show its compatibility with the  continuous case. Further, we investigate some properties of PEACE using measure theoretical concepts. Furthermore, we provide some identifiability criteria and several examples showing the generic capability of PEACE.  We note that PEACE can deal with the causal problems for which micro-level or just macro-level changes in the value of the input variables are important. Finally,  PEACE is stable under small changes in $\partial g_{in}/\partial \x$  and the joint distribution of $\X$ and $\Z$, where $g_{in}$ is obtained from $g$ by removing all functional relationships defining $\X$ and $\Z$.
	\end{abstract}

	

\maketitle

	\section{Introduction}\label{introduction}
	
Causal reasoning plays an essential role in human cognition for our adaptation to our environment. Among others, it is vital in finding the causes of both observational and non-observational events, planning for actions, and predicting future events \cite{waldmann2017oxford}. 
Causal reasoning is widely used in different domains such as Deep Learning algorithms (DLs). However, advanced DLs, such as ChatGpt can not fully perform causal reasonings (see \cite{bang2023multitask} and \cite{kalantarpourclinical}). 
Current causal reasoning frameworks/theories each have their strengths. For instance, Rubin-Neyman and Pearl causal frameworks \cite{pearl2018book,rubin1974estimating} are two of the most well-known frameworks of causality, which are used with DLs for reasoning. Indeed, these two frameworks are somehow equivalent in concepts but with different points of view. Also, Janzing et al. introduced a new framework in \cite{janzing2013quantifying} by using information theoretic concepts. In \cite{faghihi2022probabilistic},  we discussed  the strengths and the weaknesses of usual causal effect formulas in Rubin-Neyman, Pearl, and Janzing et al. frameworks. Indeed, we clarified  that at a micro level, causal effect formulas associated with Pearl causal framework work well with rare situations, such as rare medical conditions, while  Janzing et al. framework  shows the reverse strengths: works well at a macro level. In this paper, we will develop a new generic causal framework that can handle causal inference both at the micro and macro levels by using the idea of total variation as its core. 

In \cite{faghihi2022probabilistic}, we introduced a framework for causal inference that takes benefits from 1) the concept of intervention as it is in the Pearl framework, and 2) it uses a novel idea called natural availability of changing. The latter says that, given $Y=g(X,Z)$ with $X$ and $Z$ discrete random variables,  for calculating a\footnote{We use ``a'' rather than ``the'', since theoretically different formulas for calculating ``direct causal effect'' could be proposed.} direct causal effect (DCE) of $X$ on $Y$, $\P(x|z)$ might be important for different values of $X=x$ and $Z=z$. To clarify this, assume that in an observational study, the value $Z=z$  rarely occurred. Here, we have two different situations: 1) this rare occurrence has a notable or important impact on a DCE of $X$ on $Y$ (e.g., the causal effect of a rare disease on the blood pressure), and 2) it does not have a notable impact on a DCE of $X$ on $Y$ (e.g., the causal effect of a rare noise on the quality of images). In the first situation, it seems that $\P(x|z)$ should not be involved in a DCE formula of $X$ on $Y$, or if it is involved, it should be somehow strengthened. In contrast, in the second situation, it is reasonable to have $\P(x|z)$ involved in a DCE of $X$ on $Y$. We should note that the aforementioned two situations are such as two endpoints of a segment, and each inner point of this segment could happen in a real-world problem. Indeed, when we move from the first situation to the second one, the importance of involving  $\P(x|z)$ in a DCE of $X$ on $Y$ increases. To deal with this, we used a degree $d\ge 0$ in such a way that $\P(x|z)^d$ could somehow satisfy the above need: smaller and greater degrees $d$ correspond to the situations close to the first situation and the second situation discussed above, respectively. To formalize our framework, we provided some ideas and postulates as discussed in \cite{faghihi2022probabilistic}. Then, we introduced several DCEs, where  each had its  point of view. Hence, the direct causal effect values obtained by these DCEs should be interpreted in the same way they have been defined. One of these DCEs, is called \textbf{P}robabilistic \textbf{E}asy \textbf{V}ariationoal \textbf{C}ausal \textbf{E}ffect (PEACE), which measures causal changes of $Y$ with respect to continuously and interventionally  changing the values of $X$, while keeping $Z$ constant. Indeed, let $\Supp(X)=\{x_0,\ldots,x_l\}$ be the set of all possible values of $X$ with $x_0<\cdots<x_l$. In \cite{faghihi2022probabilistic}, we defined 
\begin{align*}
	\PEACE_d(X\to Y)&:=\E_Z\left(\mathcal{N}\mathcal{P}\mathcal{I}\mathcal{E}\mathcal{V}_d^z(X\to Y)\right),\\
	\mathcal{N}\mathcal{P}\mathcal{I}\mathcal{E}\mathcal{V}_d^z(X\to Y)&:=4^d\sum_{i=1}^l|g_{in}(x_i,z)-g_{in}(x_{i-1},z)|\P(x_i|z)^d\P(x_{i-1}|z)^d,
\end{align*}
where $g_{in}$ is a function that is obtained by removing the functional relationships defining $X$ and $Z$ (i.g., if $X=h(Z,W)$ for some random variable $W$, then this functional relationship should be ignored. The same is true, for a functional relationship such as $Z=q(X,W')$ for some random variable $W'$.) The notation $g_{in}(x,z)$ in the Pearl framework could be interpreted as calculating $Y$ while $\mathrm{Do}(X=x,Z=z)$. The term $\P(x_i|z)^d\P(x_{i-1}|z)^d$ is interpreted as the natural availability  of degree $d$ of changing the value of $X$ from $x_{i-1}$ to $x_i$, while keeping $Z=z$. That is the probability of selecting independently $x_i$ and $x_{i-1}$ from the subpopulation determined by $Z=z$, equipped/weakened/strengthened with a degree $d$. Note that $4^d$ is a normalizer term and comes from the fact that $\P(x_i|z)^d\P(x_{i-1}|z)^d\le 1/4^d$. Further, we note that $\sum_{i=1}^l|g_{in}(x_i,z)-g_{in}(x_{i-1},z)|$ is the total variation of the sequence $g_{in}(x_0,z),\ldots,g_{in}(x_l,z)$.

In this paper, we generalize the idea of PEACE discussed in \cite{faghihi2022probabilistic} for discrete random variables in three ways: 1) to involve continuous random variables,  2) to involve the direct causal effect of a \uline{random vector} on a random variable, and 3) in  dimension 1, to involve the positive and negative direct causal effect values for the continuous case\footnote{By dimension 1, we mean the random vector  that we mentioned before, has just one variable.}. To do so, we use the ideas of the total variation and the flux of a function of several variables. Let $Y=g(\X,\Z)$, where $\X$ and $\Z$ are continuous random vectors such that $\Supp(X)$ is a subset of an open subset $\Omega$ of $\mathbb{R}^n$. We define 
\begin{footnotesize}\begin{align}
	\PEACE_d(X\to Y)&:=\E_{\Z}\left(\mathcal{N}\mathcal{P}\mathcal{I}\mathcal{E}\mathcal{V}_d^{\z}(X\to Y)\right),\nonumber\\
\mathcal{N}\mathcal{P}\mathcal{I}\mathcal{E}\mathcal{V}_d^{\z}(X\to Y)&:=4^d\sup\left\{\int_{\Omega} g_{in}(\x,\z)\div(\varphi)(\x)\,\mathrm{d}\x: \varphi\in C_c^1(\Omega,\mathbb{R}^n),\; |\varphi|\le f_{\X|\Z}^2(\,\cdot\,|\z)\right\},\label{peacedef}
\end{align}\end{footnotesize}
\noindent where $C_c^1(\Omega,\mathbb{R}^n)$ is the set of all compactly supported continuously differentiable functions from $\Omega$ to $\mathbb{R}^n$, $\div(\phi)$ is the divergence of $\phi$, and $f_{\X|\Z}$ is the probability density function of $\X$ given $\Z$.
 While our definition of PEACE seems complicated,  for the case that $g$ is continuously differentiable, in Theorem \ref{maintheorem}, we show that 
\begin{align*}
	\mathcal{N}\mathcal{P}\mathcal{I}\mathcal{E}\mathcal{V}_d^{\z}(X\to Y)&:=4^d\int_{\Omega}\left|\frac{\partial g_{in}}{\partial \x}(\bm{t},\z)\right|f(\x|\z)^{2d}\,\d\bm{t}\d\z.
\end{align*}
It follows that PEACE is stable under small changes in $\partial g_{in} /\partial \x$  and the joint distribution of $\X$ and $\Z$. Further, we generalize the above definition of PEACE for the case that both $\X$ and $\Z$ are discrete random vectors. To do so, first, by generalizing the idea of the flux to involve discrete functions, we provide a new definition of the total variation of a function of several variables in the discrete case.  Then, we use our new definition of the total variation  to define PEACE for discrete random vectors. Furthermore, in dimension 1, namely, for the case that $X$ and $Z$ are random variables and $Y=g(X,Z)$, we define the positive and the negative PEACEs. Indeed, we note that the ordinary PEACE measures the absolute value of the direct causal changes, while positive and  negative PEACEs measure the positive and the negative direct causal changes, respectively. Here, by the direct causal changes, we mean the changes of $Y$ with respect to continuously and interventionally  increasing the values of $X$, while keeping $Z$ constant. Let $\bm{\epsilon}\in\{\bm{\pm} \}$. In \cite{faghihi2022probabilistic}, we defined the positive and the negative PEACEs for the discrete case as follows:
\begin{align*}
	\PIEV_d^{\z}(X\to Y)^{\bm{\epsilon}}&:=\sum_{i=1}^{l}\left(g_{in}(x_i)-g_{in}(x_{i-1})\right)^{\bm{\epsilon}}\P(x_i|\z)^d\P(x_{i-1}|\z)^d,
\end{align*}
where for any $r\in\mathbb{R}$,  $r^{\bm{+}}=\max\{r,0\}$, $r^{\bm{-}}=|r|-r^{\bm{+}}$, and we have assumed that $\Supp(X)=\{x_0,\ldots,x_l\}$ with $x_0<\cdots<x_l$. 
In this paper, for the case that $X$ and $Z$ are continuous random variables and $\Supp(X)\subseteq [a,b]$, we define\footnote{Here, $\Supp(X)\subseteq (-\infty, b]$, $\Supp(X)\subseteq [a,\infty)$ and $\Supp(X)\subseteq (-\infty,\infty)$ work as well.}
\begin{align*}
	\PIEV_d^{z}(X\to Y)^{\bm{\epsilon}}&:=\lim_{\lVert P\rVert\to 0}L_{P,d}^z(X\to Y)^{\bm{\epsilon}},\quad P\in\mathcal{P}([a,b]),\\
	L_{P,d}^z(X\to Y)^{\bm{\epsilon}}&:=\sum_{i=1}^{n_P}\left(g(x_i^{(P)})-g(x_{i-1}^{(P)})\right)^{\bm{\epsilon}}f(x_i^{(P)}|\z)^df(x_{i-1}^{(P)}|\z)^d,
\end{align*}
where $\mathcal{P}([a,b])$ is the set of all partitions of $[a,b]$, and for any $P\in\mathcal{P}([a,b])$, we assume that $P=\{x_0^{(P)},\ldots,x_{n_P}^{(P)}\}$ with  $x_0^{(P)}<\cdots<x_{n_P}^{(P)}$.
In Theorem \ref{NPEV}, we show that
when $g_{in}$ has the continuous partial derivatives with respect to $X$,  then
	\begin{equation}\label{eq-pos-neg}
		\PIEV_d^{\z}(X\to Y)^{\bm{\epsilon}}=\int_{-\infty}^{\infty}\left(\frac{\partial g_{in}}{\partial x}(t,\z)\right)^{\bm{\epsilon}}f_{X|\Z}^{2d}(t|\z)\,\d t.
	\end{equation}
Now, we briefly explain the organization of the paper. In Section \ref{Priliminaries}, we provide some basic concepts required to fully understand the next sections. Section \ref{Relationship Between the Divergence and the Total Variation of a Function} is devoted to discovering the relationships between the divergence of a function $f$ defined on a domain $\Omega$, the flux of $f$ passing through the boundary of $\Omega$, and the total variation of $f$. This section is a preparation to justify the causal sense of the definition given in Equation (\ref{peacedef}). In Section \ref{PEACEs of Continuous Random Vectors }, we define and investigate PEACE, and we discuss why it is causally significant. Let $Y=g(\X,\Z)$ with $\Supp(\X)\subseteq \Omega$, where $\Omega$ is an open subset of $\mathbb{R}^n$.  We show that if we define $\mu$ on the collection of all open subsets of $\Omega$ by setting $\mu(\Gamma):=\PEACE_d(\X|_{\Gamma}\to Y)$, then $\mu$ induces a Borel regular measure, where $\X|_{\Gamma}$ is the restriction of $\X$ on $\Gamma$, and more precisely it is the restricted function $X|_{\Gamma}:X^{-1}(\Gamma)\to\mathbb{R}$. Further, we show that if $\X=h(\W)$ such that $h$ is a restriction of an onto isometry of $\mathbb{R}^n$, then $\PEACE_d(\W\to Y)=\PEACE_d(\X\to Y)$. Section \ref{New Formula for Total Variation} is devoted to defining a new formula for the total variation of a multivariate discrete function compatible with the continuous definition. To do so, first, we discuss why the previous definitions by the other researchers are not suitable. Next, we generalized the concept of the flux of a function for discrete functions. Then, we use this generalization to define the total variation of a multivariate discrete function. In Section \ref{PEACE in Discrete Case}, for a  discrete random vector $\X$, we define  $\PEACE_d(\X\to Y)$   by using the total variation formula defined  in Section \ref{New Formula for Total Variation} and the idea of defining PEACE for the continuous case. Further, in Theorem \ref{dis-con}, we show that our definition of PEACE for the discrete case is compatible with the one for the continuous case. In Section \ref{Identifyability of PEACE}, we provide an identifiability criteria for $\PEACE_d(\X\to Y)$ to deal with unobserved variables such as $U_Y$ in $Y=g(\X,\Z,U_Y)$. Section \ref{Positive and Negative PEACEs} is devoted to the positive and the negative PEACEs. In this section, Equation (\ref{eq-pos-neg}) is proven. Finally, in Section~\ref{Investigating Some Examples}, we provide some examples supporting our framework and its general capability. In Section~\ref{Conclusion}, we provide the conclusion of this paper. Some lemmas, propositions, and theorems required to prove  our results are given and discussed  in Appendix \ref{Some Results on Admissible Compact Covers }.  The  proofs of our results  are provided in  Appendices \ref{appendixmaintheorem} and \ref{other}.
  \section{Preliminaries}\label{Priliminaries}
  In this section, we briefly discuss some basic concepts required for the remainder of the paper.
\subsection*{Cells and Cubes} Let $n$ be a positive integer. By an $n$-\textit{cell}, we mean a Cartesian product of $n$ bounded closed interval (i.e., $[a_1,b_1]\times\cdots\times[a_n,b_n]$, where $a_i,b_i\in\mathbb{R}$ for any $1\le i\le n$). Further, by an $n$-\textit{cube}, we mean the boundary of an $n$-cell\footnote{Here,  by ``boundary" we mean its topological meaning. Indeed, one could see that the boundary of an $n$-cell $C=[a_1,b_1]\times\cdots\times[a_n,b_n]$ is the set of all points $\x=(x_1,\ldots,x_n)$ of $C$ for which $x_i\in\{a_i,b_i\}$ for some $i$ with $1\le i\le n$. }. We may call a  $2$-cube, and a $3$-cube  a \textit{rectangle}, and a \textit{cube}, respectively.
\subsection*{Partitions of a Closed Interval} Let $[a,b]$ be an interval. Then, by a partition for $[a,b]$ we mean a chain $\{x_0,\ldots, x_n\}$, where $a=x_0<x_1<\cdots<x_{n-1}<x_n=b$, and $n$ is a positive  integer. We denote the set of all partitions of $[a,b]$ by $\mathcal{P}([a,b])$. 
\subsection*{Support of a Function}
Let $\varphi:V\subseteq\mathbb{R}^n\to W\subseteq\mathbb{R}^m$ be a function. The \textit{closed support} or simply \textit{support} of $\varphi$, denoted by $\Supp(\varphi)$, is  defined as the topological closure of the set of all points $\x\in V$ with $\varphi(\x)\neq 0$ in $V$.

\subsection*{Some Notations} In this paper, we denote elements of $\mathbb{R}^n$ by bold variables such as $\mathbf{x}=(x_1,\ldots,x_n)$. However, when our focus is on $n=1$, we use the usual variables such as $x$. Further, for $n=3$, we use the standard variables $x,y$ and $z$. 

For any two sets $S_1$ and $S_2$, by $S_1\backslash S_2$ we mean the set of all points of $S_1$ which do not belong to $S_2$.  Also, we denote the powerset of a set $S$  by $\mathcal{P}(S)$. 
We denote the \textit{interior} and the \textit{boundary} of a subset $V$ of $\mathbb{R}^n$ by $\mathrm{Int} (V)$ and $\mathrm{Bd} (V)$, respectively (see \cite{munkres2000topology} for  knowledge on topology).

Let $F:V\to W$ be a function and $U\subseteq V$. Then, we denote the \textit{restriction} of $F$ on $U$ by $F|_U$ (i.e., $F|_U:U\to W$ sending each $x\in U$ to $F(x)$).

Let  $l$ be a non-negative integer. Assume that $V\subseteq \mathbb{R}^n$ and $W\subseteq \mathbb{R}^m$. 
We denote the set of all functions $\varphi:V\to W$, whose derivatives of order $l$ exist and are continuous by $C^l(V,W)$.  Also, we denote the set of all compactly supported functions $\varphi\in C^l(V,W)$  by $C_c^l(V,W)$. 
\subsection*{Diffeomorphisms and Jacobians}
Let $U$ and $V$ be two open subsets of $\mathbb{R}^n$. A function $h:U\to V$ is called a \textit{diffeomorphism} if it is bijective and differentiable, and it has a differentiable inverse. For a function $h$ such as above (not necessarily a diffeomorphism), we define the \textit{jacobian} matrix of $h$ as follows:
\[\mathrm{Jac}(h)(\mathbf{a}):=\left(\frac{\partial h_i}{\partial x_j}(\mathbf{a})\right)_{i,j},\quad h=(h_1,\ldots,h_n).\]
\subsection*{Isometries and Orthogonal Matrices}
A function $h:U\subseteq\mathbb{R}^n\to V\subseteq\mathbb{R}^m$ is called an isometry if $|h(\x)-h(\x')|=|\x-\x'|$ for any $\x,\x'\in U$. An $n\times n$ matrix $A$ whose entries are real numbers is called \textit{orthogonal} if $AA^T=I_n$, where $A^T$ is the \textit{transpose} of $A$. By the Mazur-Ulam theorem \cite[Theorem 1.3.5]{fleming2007isometries}, for each onto isometry $h:\mathbb{R}^n\to\mathbb{R}^n$, there exists an orthogonal matrix $A$ such that $h(\x)=A\x+\mathbf{a}$, where $A$ is an invertible $n\times n$ matrix, and $\mathbf{a}\in\mathbb{R}^n$ (here, we consider the vectors in $\mathbb{R}^n$ as colomn vectors).
\subsection*{Measure Theoretical Preliminaries}
Let $\Omega$ be a set. A subcollection $\Sigma$ of $\mathcal{P}(\Omega)$ is called a $\sigma$-algebra, if $\emptyset\in\Sigma$, $E\in\Sigma$ implies $\Omega\backslash E\in\Sigma$ for any $E\in\Sigma$ , and $\bigcup_{i\in I}E_i\in\Sigma$ for  any family $\{E_i\}_{i\in I}$ of elements of $\Sigma$. Let $\Sigma$ be a $\sigma$-algebra on $\Omega$. Then, a function $\mu:\Sigma\to [0,\infty]$ is called a measure, if $\mu(\emptyset)=0$ and $\mu\left(\bigcup_{i=0}^{\infty}E_i\right)=\sum_{i=0}^{\infty}\mu(E_i)$ for any family $\{E_i\}_{i=1}^{\infty}$ of pair-wise disjoint elements of $\Sigma$. The latter property is called the \textit{countably additive} property. A function $\mu_*:\mathcal{P}(\Omega)\to[0,\infty]$ is called an \textit{outer measure}, if $\mu_*(\emptyset)=0$, $\mu_*(E)\le \mu_*(F)$ for any $E,F\in\mathcal{P}(\Omega)$ with $E\subseteq F$, and $\mu\left(\bigcup_{i=0}^{\infty}E_i\right)\le\sum_{i=0}^{\infty}\mu(E_i)$ for any family $\{E_i\}_{i=1}^{\infty}$ of elements of $\mathcal{P}(\Omega)$. The latter two properties are called \textit{monotonicity} and \textit{countably subadditivity}, respectively. Let $\mu_*$ be an outer measure on $\Omega$. Then, $E\in\mathcal{P}(\Omega)$ is called \textit{$\mu_*$-measurable}, if for any $A\in\mathcal{P}(\Omega)$, we have that $\mu_*(A)=\mu_*(A\cap E)+\mu_*(A\cap (\Omega\backslash E))$. It is well-known that the set of all $\mu_*$-measurable subsets of $\Omega$ is a $\sigma$-algebra, and $\mu_*$ is a measure on this $\sigma$-algebra. We define the $\sigma$-algebra generated by a subcollection $\Theta$ of $\mathcal{P}(\Omega)$ to be the intersection of all $\sigma$-algebras on $\Omega$ which contain $\Theta$. If $\Omega$ is a topological space, then the $\sigma$-algebra generated by open sets in $\Omega$ is called a \textit{Borel $\sigma$-algebra}.  Each element of a Borel $\sigma$-algebra is called a \textit{Borel set}. A measure defined on a Borel $\sigma$-algebra is called a \textit{Borel measure}. In the case that $\Omega\subseteq\mathbb{R}^n$ is open, an outer measure $\mu_*$ on $\Omega$ is called \textit{Borel regular}, if for any $A\subseteq \Omega$, there exists a Borel set $E$ in $\Omega$ with $\mu_*(A)=\mu_*(E)$.  Roughly speaking, the \textit{Lebesgue measure} on $\mathbb{R}^n$ is a generalization of the volume in $\mathbb{R}^n$ (for $n=1$ and  $n=2$, we will have a generalization of the length and the area for the subsets of $\mathbb{R}$ and $\mathbb{R}^2$, respectively).
See \cite{folland1999real} for the precise definitions of the Lebesgue measure, the Lebesgue integration, and other related concepts and results in measure theory.
\subsection*{Integral on an Open set}
Let $F:V\subseteq\mathbb{R}^n\to W\subseteq\mathbb{R}^m$ be a function, where $V$ is bounded.  Assume that $C$ is an $n$-cell containing $V$. Define $\widetilde{F}:C\to W$ by setting $\widetilde{F}|_V=F$ and $\widetilde{F}|_{C\backslash V}\equiv 0$. Following \cite[\S 13]{munkres1991analysis},  we define the  integral of $F$ on $V$ as 
$\int_{V}F\,\d\x:=\int_{C}\widetilde{F}\,\d\x$. 
A bounded subset $V$ of $\mathbb{R}^n$ is called \textit{rectifiable} if $\int_V\;\d\x$ exists (see \cite[\S 14]{munkres1991analysis}). Note that $V$ is rectifiable if and only if the Lebesgue measure of  $\mathrm{Bd}(V)$ is $0$ (\cite[Theorem 14.1]{munkres1991analysis}). Now, we take a close look at integrals over open subsets of $\mathbb{R}^n$. 
Let $F:V\to W\subseteq\mathbb{R}^m$ be a function, where $V$ is an open subset of $\mathbb{R}^n$.   Assume that $\{C_i\}_{i=0}^{\infty}$ is a sequence of rectifiable compact subspaces of $V$ in such a way that
$C_i\subseteq \mathrm{Int}(C_{i+1})$ and the boundary of $C_i$ is a piece-wise manifold of class $C^1$ for any $i\ge 0$, and 
   $V=\bigcup_{i=0}^{\infty}C_i$  (for the existence of such a sequence see Lemma \ref{existenceofthecomapctsequence}).  In addition, assume that $\int_{C_i}|F(\x)|\,\d\x<\infty$ for any $i\ge 0$. Then, $\lim_{i\to\infty}\int_{C_i}F(\x)\,\mathrm{d}\x$ exists, and we have that (see \cite[\S 15]{munkres1991analysis}):
\[\int_VF(\x)\,\mathrm{d}\x=\lim_{i\to\infty}\int_{C_i}F(\x)\,\mathrm{d}\x.\]
In the above, we call the sequence $\{C_i\}_{i=0}^{\infty}$ an \textit{admissible compact cover} for $V$. 
\section{Relationship Between the Divergence and the Total Variation of a Function}\label{Relationship Between the Divergence and the Total Variation of a Function}
The main purpose of this section is to introduce a famous multivariate version of the total variation and show how it intuitively measures the variations of a function. To do so, we need to have some steps. Let $f:V\to\mathbb{R}$ be a function, where $V$ is an open subset of $\mathbb{R}^n$. Also, let $\Omega$ be a compact subspace of $V$, and  $\mathscr{S}$ be the boundary of $\Omega$ of class $C^1$. Moreover, assume that $C$ is an infinitesimal $n$-cell contained in $\Omega$ and $D$ is its boundary. In this section, first, we intuitively discuss  the flux of $f$ passing through $\mathscr{S}$. Then,  we state and clarify the divergence theorem, which makes a relationship between the flux  of  $f$  passing through  $\mathscr{S}$ and the integral of the divergence of $f$ on $\Omega$. Next, the definition of the total variation of $f$ is given, and  we justify that for the univariate functions, this definition of the total variation coincides with the classic total variation of a function under some assumptions. Further,
 we explain how the flux and the total variation of $f$ are related concepts. Indeed, the flux of $f$ on $D$ is approximately equal to the multiplication of the divergence of $f$ on $C$ and the volume of $C$.  However, the total variation of $f$ is obtained similarly to the  latter  with one change: we  replace the divergence of $f$ with the inner product of the divergence of $f$ and a weight function, which is bounded by 1 and makes the above inner product maximum. In this way, on the one hand, under some assumptions,  the absolute values of the partial derivatives of $f$ will appear, which are somehow  related to the variations of $f$. On the other hand, this implies a close relationship between the divergence and the total variation of $f$.
\subsection{Divergence Theorem}\label{Divergence Theorem}
Let $F:U\to\mathbb{R}^n$ be a continuously differentiable function, where $U$ is an open subset of $\mathbb{R}^n$. Then, the divergence of $F$ is defined as:
\[\div(F):U\to\mathbb{R},\qquad\div(\x):=\sum_{i=1}^n\frac{\partial F}{\partial x_i}(\x).\]
The \textit{flux} of $F$ passing through a surface $\mathscr{S}\subseteq U$ is defined as the following surface integral:
\[\mathrm{Flux}(F;\mathscr{S}):=\oint_{\mathscr{S}}F(\x)\cdot\widehat{N}(\x)\,\d S,\]
where $\widehat{N}(\x)$ is the outward unit vector perpendicular to $\mathscr{S}$ at the point $\x$, and $F(\x)\cdot\widehat{N}(\x)$ is the standard inner product of $F(\x)$ and $\widehat{N}(\x)$ as vectors  in $\mathbb{R}^n$ for any $\x\in \mathscr{S}$. We call $\widehat{N}(\x)$ the \textit{outward unit normal vector} for $\mathscr{S}$ at $\x$ as well. 

The following theorem called the \textit{divergence theorem}\footnote{This theorem is famous by two other names: \textit{Gauss's theorem}, and \textit{Ostrogradsky's theorem}}, makes a bridge between the flux of a function passing through a surface and a multiple integral on the region surrounded by that surface (see \cite[Theorem 9.2.4]{willem2023functional}).
 
\begin{Theorem}
	Let $\Omega\subseteq\mathbb{R}^n$ be compact and its boundary is $C^1$. Assume that $V\subseteq\mathbb{R}^n$ be an open neighborhood of $\Omega$. Then, for any continuously differentiable function $F:V\to \mathbb{R}^n$, we have that 
	\[\int_{\Omega} \div(F)\,\mathrm{d}\x=\oint_{\mathscr{S}}F\cdot\widehat{N}\,\mathrm{d}S,\qquad \mathscr{S}=\bd(\Omega).\]
\end{Theorem}

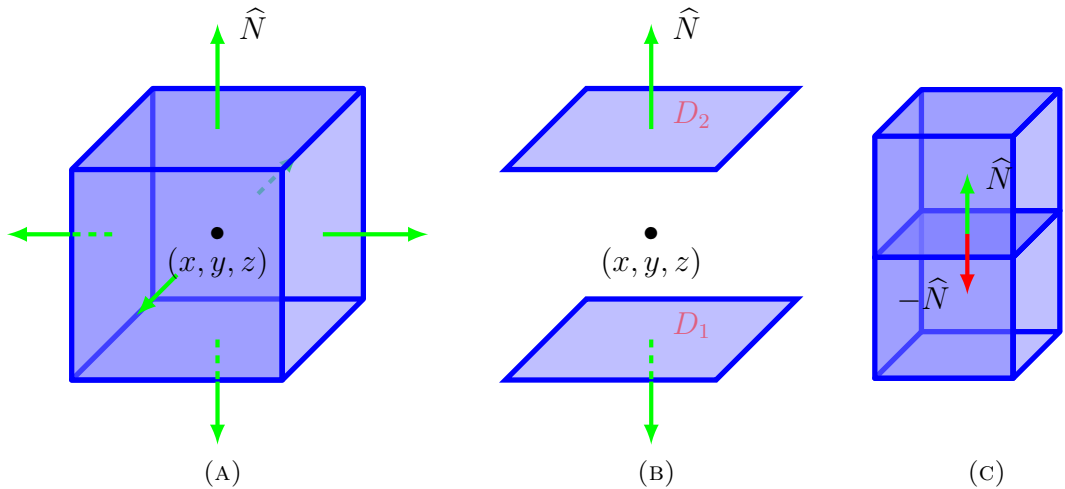
\begin{figure}
\subfloat[]{	\begin{tikzpicture}[scale = 0.7,
		pin distance=3cm,
		every pin/.style={inner sep=1pt},
		every pin edge/.style={black,very thick},
		line width=2pt,
		axisProp/.style={x={(-136:2cm)},
			y={(-15:2cm)},
			z={(90:2cm)}},
		axis/.style={x={(-136:2cm)},
			y={(-15:4.5cm)},
			z={(90:3cm)}}]
		
			\begin{scope}[canvas is xz plane at y=0]

				\coordinate (c01) at (0,0); 
				\coordinate (c02) at (4,0); 
				
				\coordinate (c12) at (4,4);
				\coordinate (c11) at (0,4);
				\coordinate (c) at (2,2);

			\end{scope}
		
			\begin{scope}[canvas is xz plane at y=2]

			\coordinate (cc) at (2,2);

		\end{scope}
	
			\begin{scope}[canvas is xz plane at y=4]
				\coordinate (d01) at (0,0); 
				\coordinate (d02) at (4,0); 
				
				\coordinate (d12) at (4,4);
				\coordinate (d11) at (0,4);
			
			\coordinate (d) at (2,2);
			\end{scope}
			
		\begin{scope}[canvas is xz plane at y=6]

		\coordinate (dd) at (2,2);

	\end{scope}

	\begin{scope}[canvas is xz plane at y=-2]
	\coordinate (dy) at (2,2); 
\end{scope}
	\begin{scope}[canvas is xz plane at y=-0.8]
	\coordinate (dashy) at (2,2); 
\end{scope}
	\begin{scope}[canvas is xy plane at z=4]
	\coordinate (zmax) at (2,2); 
\end{scope}
	\begin{scope}[canvas is xy plane at z=0]
	\coordinate (zmin) at (2,2); 
\end{scope}
	\begin{scope}[canvas is xy plane at z=-2]
	\coordinate (dz) at (2,2); 
\end{scope}
	\begin{scope}[canvas is xy plane at z=6]
	\coordinate (ddz) at (2,2); 
\end{scope}
	\begin{scope}[canvas is yz plane at x=0]
	\coordinate (xmin) at (2,2); 
\end{scope}
	\begin{scope}[canvas is yz plane at x=4]
	\coordinate (xmax) at (2,2); 
\end{scope}
	\begin{scope}[canvas is yz plane at x=6]
	\coordinate (ddx) at (2,2); 
\end{scope}
	\begin{scope}[canvas is yz plane at x=-2]
	\coordinate (dx) at (2,2); 
\end{scope}
	\begin{scope}[canvas is yz plane at x=-0.8]
	\coordinate (dashx) at (2,2); 
\end{scope}
	\begin{scope}[canvas is yz plane at x=2]
	\coordinate (p) at (2,2); 
\end{scope}
\begin{scope}[canvas is xz plane at y=7]
	\coordinate (e01) at (0,0); 
	\coordinate (e02) at (4,0); 
	
	\coordinate (e12) at (4,4);
	\coordinate (e11) at (0,4);
	
\end{scope}

			\draw[->,green,ultra thick,dashed] (zmin) -- (dz);
		\fill[fill=blue!50,fill opacity=0.5]
		(c01) -- (c02) -- (c12) -- (c11)--cycle
		(c01) -- (d01) -- (d02) --(c02)--cycle
		(c01) -- (d01) -- (d11) -- (c11) --cycle;
		
		\draw[blue,join=round] (c01) -- (c02) -- (c12) -- (c11)--cycle;
		\draw[blue] (c01) -- (d01);
		\fill[fill=blue!50,fill opacity=0.5]
		(c12) -- (c11) -- (d11) -- (d12) -- cycle
		(d01) -- (d02) -- (d12) -- (d11) -- cycle;
		\draw[blue,join=round] (d01) -- (d02) -- (d12) -- (d11) -- cycle;
		\draw[blue,join=round] (c02) -- (d02) (c12) -- (d12) (d11) -- (c11) -- (c12) -- (d12) (d02) -- (d12);
		
		\draw[->,green,ultra thick] (d) -- (dd)node[right, black]{$\;\widehat{N}$};
		\draw[dashed,green,ultra thick] (c) -- (dashy);
    	\draw[->,green,ultra thick] (dashy) -- (dy);
    		\draw[->,green,ultra thick] (zmax) -- (ddz);
    		
    			\draw[->,green,ultra thick] (xmax) -- (ddx);
    			\draw[dashed,green,ultra thick] (xmin) -- (dashx);
    			\draw[->,green,ultra thick] (dashx) -- (dx);
    			\draw (p)node{$\bullet$}node[below]{$(x,y,z)$};
	\end{tikzpicture}}\qquad
\subfloat[]{		\begin{tikzpicture}[scale = 0.7,
			pin distance=2cm,
			every pin/.style={inner sep=1pt},
			every pin edge/.style={black,very thick},
			line width=2pt,
			axisProp/.style={x={(-126:2cm)},
				y={(-15:2cm)},
				z={(90:2cm)}},
			axis/.style={x={(-126:2cm)},
				y={(-15:4.5cm)},
				z={(90:2cm)}}]
			
				\begin{scope}[canvas is xz plane at y=-2]

				\coordinate (bb) at (2,2);

			\end{scope}
		
			\begin{scope}[canvas is xz plane at y=0]

				\coordinate (c01) at (0,0); 
				\coordinate (c02) at (4,0); 
				
				\coordinate (c12) at (4,4);
				\coordinate (c11) at (0,4);
				
				\coordinate (c) at (2,2);
				
			\end{scope}
			
			\begin{scope}[canvas is xz plane at y=2]

				\coordinate (cc) at (2,2);

			\end{scope}
			
			\begin{scope}[canvas is xz plane at y=4]
				\coordinate (d01) at (0,0); 
				\coordinate (d02) at (4,0); 
				
				\coordinate (d12) at (4,4);
				\coordinate (d11) at (0,4);
				
				\coordinate (d) at (2,2);
			\end{scope}
			
			\begin{scope}[canvas is xz plane at y=6]

				\coordinate (dd) at (2,2);

			\end{scope}
			
			\begin{scope}[canvas is xz plane at y=7]
				\coordinate (e01) at (0,0); 
				\coordinate (e02) at (4,0); 
				
				\coordinate (e12) at (4,4);
				\coordinate (e11) at (0,4);
				
			\end{scope}
			\begin{scope}[canvas is xz plane at y=-0.8]
			\coordinate (dashy) at (2,2); 
		\end{scope}


			\fill[fill=blue!50,fill opacity=0.5]
			(c12) -- (c11) -- (c01) -- (c02) node[midway, below, red]{$D_1$}-- cycle;
		\draw[blue]
			(c12) -- (c11) -- (c01) -- (c02) -- cycle;
			
			\fill[fill=blue!50,fill opacity=0.5]
			(d12) -- (d11) -- (d01) -- (d02) node[midway, below, red]{$D_2$}-- cycle;
			\draw[blue]
			(d12) -- (d11) -- (d01) -- (d02) -- cycle;
			
				\draw[->,green,ultra thick] (d) -- (dd)node[right, black]{$\;\widehat{N}$};;
					\draw[dashed,green,ultra thick] (c) -- (dashy);
				\draw[->,green,ultra thick] (dashy) -- (bb);
			\draw (cc) node{$\bullet$}node[ below ]{$(x,y,z)$};
		\end{tikzpicture}}\qquad
\subfloat[]{	\begin{tikzpicture}[scale = 0.46, 
		pin distance=3cm,
		every pin/.style={inner sep=1pt},
		every pin edge/.style={black,very thick},
		line width=2pt,
		axisProp/.style={x={(-136:2cm)},
			y={(-15:2cm)},
			z={(90:2cm)}},
		axis/.style={x={(-136:2cm)},
			y={(-15:4.5cm)},
			z={(90:3cm)}}]
		
			\begin{scope}[canvas is xz plane at y=0]

				\coordinate (c01) at (0,0); 
				\coordinate (c02) at (4,0); 
				
				\coordinate (c12) at (4,3.5);
				\coordinate (c11) at (0,3.5);
				\coordinate (c) at (2,2);

			\end{scope}
		\begin{scope}[canvas is xz plane at y=-0.8]
			\coordinate (dashy) at (2,2); 
		\end{scope}
			\begin{scope}[canvas is xz plane at y=1.75]

			\coordinate (cc) at (2,1.75);

		\end{scope}
		\begin{scope}[canvas is xz plane at y=-2.5]
		\coordinate (dy) at (2,2); 
	\end{scope}
			\begin{scope}[canvas is xz plane at y=3.5]
				\coordinate (d01) at (0,0); 
				\coordinate (d02) at (4,0); 
				
				\coordinate (d12) at (4,3.5);
				\coordinate (d11) at (0,3.5);
			
			\coordinate (d) at (2,1.75);
			\end{scope}
			
		\begin{scope}[canvas is xz plane at y=5.25]

		\coordinate (dd) at (2,1.75);

	\end{scope}

\begin{scope}[canvas is xz plane at y=7]
	\coordinate (e01) at (0,0); 
	\coordinate (e02) at (4,0); 
	
	\coordinate (e12) at (4,3.5);
	\coordinate (e11) at (0,3.5);
	
\end{scope}

		
		\fill[fill=blue!50,fill opacity=0.5]
		(c01) -- (c02) -- (c12) -- (c11)--cycle
		(c01) -- (d01) -- (d02) --(c02)--cycle
		(c01) -- (d01) -- (d11) -- (c11) --cycle;
		
		\draw[blue,join=round] (c01) -- (c02) -- (c12) -- (c11)--cycle;
		\draw[blue] (c01) -- (d01);
		\fill[fill=blue!50,fill opacity=0.5]
		(c12) -- (c11) -- (d11) -- (d12) -- cycle
		(d01) -- (d02) -- (d12) -- (d11) -- cycle;
		\draw[blue,join=round] (d01) -- (d02) -- (d12) -- (d11) -- cycle;
		\draw[blue,join=round] (c02) -- (d02) (c12) -- (d12) (d11) -- (c11) -- (c12) -- (d12) (d02) -- (d12);
		
			\fill[fill=blue!50,fill opacity=0.5]
		(d01) -- (d02) -- (d12) -- (d11)--cycle
		(d01) -- (e01) -- (e02) --(d02)--cycle
		(d01) -- (e01) -- (e11) -- (d11) --cycle;
		
		\draw[blue,join=round] (d01) -- (d02) -- (d12) -- (d11)--cycle;
		\draw[blue] (d01) -- (e01);
		\fill[fill=blue!50,fill opacity=0.5]
		(d12) -- (d11) -- (e11) -- (e12) -- cycle
		(e01) -- (e02) -- (e12) -- (e11) -- cycle;
		\draw[blue,join=round] (e01) -- (e02) -- (e12) -- (e11) -- cycle;
		\draw[blue,join=round] (d02) -- (e02) (d12) -- (e12) (e11) -- (d11) -- (d12) -- (e12) (e02) -- (e12);
		\draw[->,green,ultra thick] (d) -- (dd)node[right,black]{$\,\widehat{N}$};
		\draw[->,red,ultra thick] (d) -- (cc)node[left,black]{$-\widehat{N}\,$};
	\draw[->,white,ultra thick] (dashy) -- (dy);
	\end{tikzpicture}}
\caption{(A) A $3$-cell centered at the point $(x,y,z)$ with the outward unit normal vector $\widehat{N}$. (B) The faces of the cell parallel to the $xy$ plane. (C) Two identical $3$-cells, one of which is on  top of the other one. The outward unit normal vectors of  these two cubes on their common face are in  opposite directions.  }\label{cubes}
\end{figure}

\begin{figure}
	\begin{center}
	\input{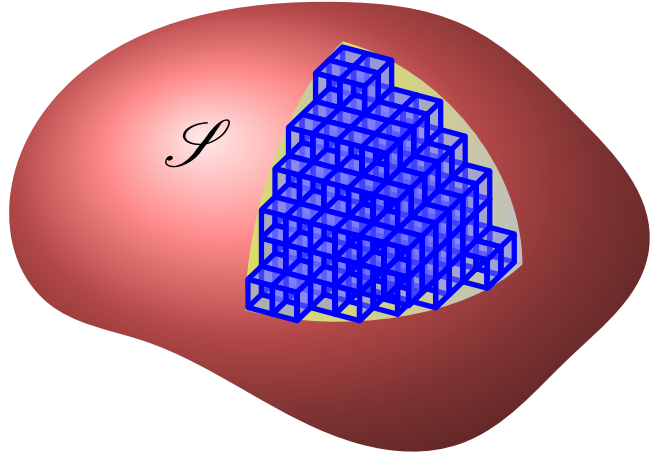}
	\caption{A compact region $\Omega\subseteq\mathbb{R}^3$, whose boundary $\mathscr{S}$ is smooth. The region $\Omega$ could be estimated by the union of the cells belonging to an admissible collection of cells for $\Omega$. }\label{potato}
   \end{center}
\end{figure}

\begin{figure}	
\subfloat[]{	
\begin{tikzpicture}[line width=2pt,scale = 0.5]
	\fill[fill=blue!50,fill opacity=0.5] (0,0) -- (2,0) -- (2,2) -- (0,2)--cycle;
	\draw (1,1)node[font=\large]{$C_1$};
\draw[red] (0,0) -- (2,0) -- (2,2) -- (0,2)--cycle;

\fill[fill=blue!50,fill opacity=0.5] (-2,0) -- (0,0) -- (0,2) -- (-2,2)--cycle;
	\draw (-1,1)node[font=\large]{$C_2$};
\draw[blue] ((-2,0) -- (0,0) -- (0,2) -- (-2,2)--cycle;

\fill[fill=blue!50,fill opacity=0.5] (-2,2) -- (0,2) -- (0,4) -- (-2,4)--cycle;
\draw (-1,3)node[font=\large]{$C_3$};
\draw[blue] ((-2,2) -- (0,2) -- (0,4) -- (-2,4)--cycle;

\fill[fill=blue!50,fill opacity=0.5] (2,0) -- (4,0) -- (4,2) -- (2,2)--cycle;
	\draw (3,1)node[font=\large]{$C_5$};
\draw[blue] ((2,0) -- (4,0) -- (4,2) -- (2,2)--cycle;

\fill[fill=blue!50,fill opacity=0.5] (0,2) -- (2,2) -- (2,4) -- (0,4)--cycle;
	\draw (1,3)node[font=\large]{$C_4$};
\draw[blue] (0,2) -- (2,2) -- (2,4) -- (0,4)--cycle;

\fill[fill=blue!50,fill opacity=0.5] (0,-2) -- (2,-2) -- (2,0) -- (0,0)--cycle;
	\draw (1,-1)node[font=\large]{$C_6$};
\draw[blue] (0,-2) -- (2,-2) -- (2,0) -- (0,0)--cycle;
\end{tikzpicture}
}
\qquad
	\subfloat[]{	
		\begin{tikzpicture}[line width=2pt,scale = 0.5]
		
			\draw[red] (0,0) -- (2,0) -- (2,2) -- (0,2)--cycle;

			\draw[blue] (-2,0) -- (0,0);
			 \draw[red ](0,0)-- (0,2) -- (-2,2);
			\draw[blue] (-2,2)--(-2,0);

			\draw[red] (-2,2) -- (0,2) -- (0,4);
			 \draw[blue] (0,4)-- (-2,4)--(-2,2);

			\draw[blue] (2,0) -- (4,0) -- (4,2) -- (2,2);
			\draw[red] (2,2)--(2,0);

			\draw[red] (0,2) -- (2,2);
			
			 \draw[blue] (2,2) -- (2,4) -- (0,4);
			\draw[red] (0,4)--(0,2);
		
			\draw[blue] (0,-2) -- (2,-2)--(2,0) ;
			\draw[red] (2,0) -- (0,0);
			\draw[blue] (0,0)-- (0,-2);
		\end{tikzpicture}
	}\qquad
	\subfloat[]{	
	\begin{tikzpicture}[line width=2pt,scale = 0.5]

		\draw[blue] (0,0) -- (-2,0) -- (-2,4)--(0,4) ;

		\draw[blue] (2,0) -- (4,0) -- (4,2) -- (2,2);

		\draw[blue] (2,2) -- (2,4) -- (0,4);

		\draw[blue] (0,0)--(0,-2) -- (2,-2) -- (2,0);
	\end{tikzpicture}
}

\caption{\raggedright\Small{Admissible collections could be similarly defined in any dimension. In this figure, for a better visualization, we  consider an admissible collection  $\Gamma=\{C_1,\ldots, C_6\}$ of $2$-cells for a 2-dimensional shape. Note that the subfigures (A), (B), and (C) shows $\lVert\Gamma\rVert$, $\mathscr{S}_{\Gamma}$, and $\mathscr{S}_{\lVert\Gamma\rVert}$, respectively. However, each of the red sides is considered twice as it is the common side of two adjacent squares.} }\label{2dimadmissible}

\end{figure}
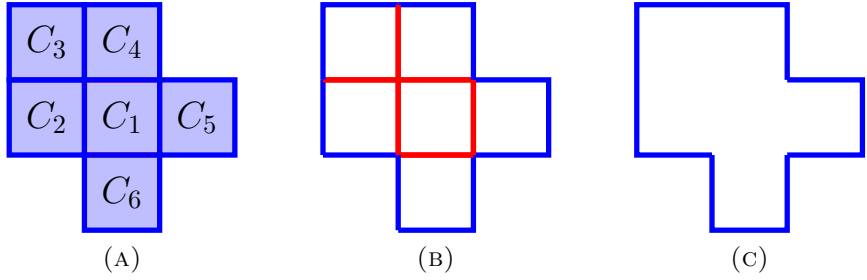

Now, we provide an intuitive justification for the divergence theorem in three dimensions, which makes it clear where $\div(F)$ comes from\footnote{We make it clear that this is not a precise proof!}. To do so, first,  assume that we are given an infinitesimal $3$-cell  as shown in Part (A) of Figure~\ref{cubes}. Also, assume that the boundary of this cell is a cube $D$, and the side lengths of $D$ are $\Delta x, \Delta y$, and $\Delta z$.  To compute $\mathrm{Flux}(F;D)$, we divide $D$ into three pairs of parallel faces. First, as shown in  Part (B) of Figure \ref{cubes}, we consider the  faces $D_1$ and $D_2$  parallel to the $xy$ plane.  
For each $i=1,2$, one could say that $F$ approximately takes the same value on $D_i$. Hence, $\mathrm{Flux}(F;D_1)$ and $\mathrm{Flux}(F;D_2)$ are approximately $\left(F\cdot \widehat{N}\right)(x,y,z-\Delta z/2)\Delta x\Delta y$ and $\left(F\cdot \widehat{N}\right)(x,y,z+\Delta z/2)\Delta x\Delta y$, respectively. We note that 
\[\widehat{N}(x,y,z-\Delta z/2) =- \vec{k}=(0,0,-1),\quad \widehat{N}(x,y,z+\Delta z/2) = \vec{k}=(0,0,1).\]
Now, let $F=(F_1,F_2,F_3)$. Then, the above fluxes are respectively as follows: 
\[ -F_3(x,y,z-\Delta z/2)\Delta x\Delta y, \qquad F_3(x,y,z
+\Delta z/2)\Delta x\Delta y.\]
Hence, we have that
\[ \mathrm{Flux}(F;D_1\cup D_2)\approx \bigl(F_3(x,y,z+\Delta z/2)-F_3(x,y,z-\Delta z/2)\bigr)\Delta x\Delta y.\]
We note that
\begin{align*}
	F_3(x,y,z+\Delta z/2)&\approx F_3(x,y,z)+\frac{\Delta z}{2}\frac{\partial F_3}{\partial z}(x,y,z),\\
	F_3(x,y,z-\Delta z/2)&\approx F_3(x,y,z)-\frac{\Delta z}{2}\frac{\partial F_3}{\partial z}(x,y,z),
\end{align*}
which implies that 
\[F_3(x,y,z+\Delta z/2)-F_3(x,y,z-\Delta z/2)\approx \Delta z \frac{\partial F_3}{\partial z}(x,y,z).\]
Thus,  $ \mathrm{Flux}(F;D_1\cup D_2)\approx \left(\partial F_3/\partial z\right)(x,y,z)\Delta x\Delta y\Delta z$. For the flux of $F$ passing through each of the other two pairs of parallel faces of $D$, by similar arguments, we obtain $ \left(\partial F_1/\partial x\right)(x,y,z)\Delta x\Delta y\Delta z$ and $ \left(\partial F_2/\partial y\right)(x,y,z)\Delta x\Delta y\Delta z$. Therefore, we have that
\[\mathrm{Flux}(F;D)\approx \left(\frac{\partial F_1}{\partial x}+\frac{\partial F_2}{\partial y}+\frac{\partial F_3}{\partial z}\right)(x,y,z)\Delta x\Delta y\Delta z=\div(F)(x,y,z)\Delta x\Delta y\Delta z.\] 
Now, let us consider two identical $3$-cells as shown  in  Part (C) of Figure \ref{cubes}. Since the outward unit normal vectors on the common face of these two cells are in the opposite directions, the effect of the common face disappears  in the fluxes of $F$ passing through the boundaries of these cells\footnote{For the cells in the bottom and  the top, we obtain  $\left(\partial F_3/\partial z\right)(x,y,z+\Delta z/2)\Delta x\Delta y\Delta z$ and $-\left(\partial F_3/\partial z\right)(x,y,z+\Delta z/2)\Delta x\Delta y\Delta z$ for the flux of $F$ passing through the common face of the cells, respectively.}. A similar argument holds for any two adjacent identical $3$-cells. Now, let $\Omega\subseteq\mathbb{R}^3$ be compact as shown in Figure \ref{potato} with the smooth boundary $\mathscr{S}$. 
We call a collection $\Gamma$ of $3$-cells in $\mathbb{R}^3$ \textit{addmissible} for $\Omega$, if we have the following conditions:
\begin{enumerate}
	\item Each cell in $\Gamma$ is contained in $\Omega$, 
	\item Every two cells in $\Gamma$ are identical,
	\item The intersection of every two distinct cells in $\Gamma$ is empty or a common vertex, edge, or  face of both, and
	\item 
	$\Gamma$ is maximal with respect to the above three conditions.
\end{enumerate}
Let $\Gamma$ be an admissible collection of  $3$-cells for $\Omega$. We denote the union of all cells in $\Gamma$ by $\lVert\Gamma\rVert$. We also denote the boundary of $\lVert\Gamma\rVert$ and the disjoint union of the boundaries of the cells in $\Gamma$ by $\mathscr{S}_{\lVert\Gamma\rVert}$ and $\mathscr{S}_{\Gamma}$, respectively (see Figure \ref{2dimadmissible}).
Then, as we discussed above,  we have that $\mathrm{Flux}(F;\mathscr{S}_{\Gamma})=\mathrm{Flux}(F;\mathscr{S}_{\lVert\Gamma\rVert})$.
Thus, we have that 
\[\mathrm{Flux}(F;\mathscr{S}_{\lVert\Gamma\rVert})=\sum\div(F)(x,y,z)\Delta x\Delta y\Delta z,\]
where the summation is over all center points of the cells in $\Gamma$.

Assume that
$\mathrm{mesh}(\Gamma):=\sqrt{\Delta x^2+\Delta y^2+\Delta z^2}$, where $\Delta x$, $\Delta y$, and $\Delta z$ are the side lengths of each cube in $\Gamma$. 
 We note that by tending $\mathrm{mesh}(\Gamma)$ to $0$, $\mathscr{S}_{\lVert \Gamma\rVert }$ tends to $\mathscr{S}$. It follows that 
 \begin{align*}
 	\mathrm{Flux}(F;\mathscr{S})&\approx\lim_{\mathrm{mesh}(\Gamma)\to 0}\mathrm{Flux}(F;\mathscr{S}_{\lVert\Gamma\rVert})\\
 	&\approx\lim_{\mathrm{mesh}(\Gamma)\to 0}\sum\div(F)(x,y,z)\Delta x\Delta y\Delta z\approx\iiint_{\Omega}\div(F)\,\d V.
 \end{align*}
Note that in our intuitive justification of the divergence theorem, we used the linear approximations of $F$.  That is why instead of equality in the theorem  we have an approximation. However, in the precise proof of the theorem, it is shown that the error terms in the approximations tend to 0, and hence we will have  equality in the theorem. 
	
\subsection{Total Variation}
Let $g:[a,b]\to\mathbb{R}$ be a function. The \textit{total variation} of $g$, denoted by $\TV(g)$, is defined as follows:
\[\TV(g):=\max_{P\in\mathcal{P}([a,b])}\sum_{i=1}^{n_P}\left|g\left(x_i^{(P)}\right)-g\left(x_{i-1}^{(P)}\right)\right|,\]
where $P=\{x_0^{(P)},\ldots, x_n^{(P)}\}$ for any $P\in\mathcal{P}([a,b])$.

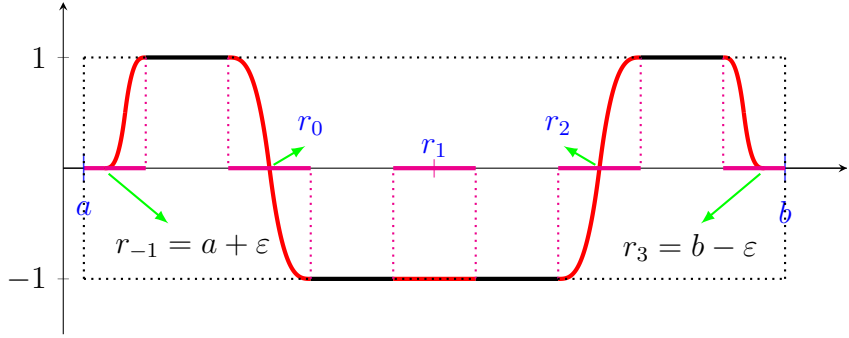
\begin{figure}
	\begin{tikzpicture}
\begin{axis}[axis lines=center,domain=-1:24,
	ymin=-1.5,ymax=1.5,xtick style={draw=none},
	height=6cm,width=12cm, xticklabel=\empty]
	\addplot[domain=0:1,samples=2,color=black] {0};
		\draw[ultra  thick, magenta] (0.5,0)--(1,0);
	\addplot[domain=1:1.5,samples=50,color=red,ultra thick] {4*(x-1)^3};
	\addplot[domain=1.5:2,samples=50,color=red,ultra thick] {4*(x-2)^3+1};
	\addplot[domain=2:4,samples=2,color=black,ultra thick] {1};
	\addplot[domain=4:5,samples=50,color=red,ultra thick] {-(x-4)^3+1};
	\addplot[domain=5:6,samples=50,color=red,ultra thick] {-(x-6)^3-1};
	\addplot[domain=6:8,samples=2,color=black,ultra thick] {-1};
	\addplot[domain=8:10,samples=2,color=red,ultra thick] {-1};
	\addplot[domain=10:12,samples=2,color=black,ultra thick] {-1};
	\addplot[domain=14:16,samples=2,color=black,ultra thick] {1};
	\addplot[domain=12:13,samples=50,color=red,ultra thick] {(x-12)^3-1};
	\addplot[domain=13:14,samples=50,color=red,ultra thick] {(x-14)^3+1};
	\addplot[domain=16:16.5,samples=50,color=red,ultra thick] {-4*(x-16)^3+1};
	\addplot[domain=16.5:17,samples=50,color=red,ultra thick] {-4*(x-17)^3};
	\addplot[domain=17:19,samples=2,color=black] {0};
	\draw[blue] (0.5,0)node{\small{$|$}};
	\draw[blue] (0.5,-.2)node[below]{$a$};
	\draw(1,-.5)node[below right]{$r_{-1}=a+\epsilon$};
	\draw[green,->,thick](1.1,-0.05)--(2.55,-0.5);
	\draw[blue] (6,0.2)node[above]{$r_0$};
	\draw[->,green,thick] (5.1,0.03)--(5.87,0.2);
	
	\draw[ultra  thick, magenta] (4,0)--(6,0);
	\draw[dotted, thick,magenta](4,1)--(4,0);
	\draw[dotted, thick,magenta](2,1)--(2,0);
	
	\draw[blue] (17.5,0)node{\small{$|$}};
	\draw[dotted, thick,magenta](6,-1)--(6,0);
	\draw[dotted, thick,magenta](8,-1)--(8,0);
	\draw[dotted, thick,magenta](10,-1)--(10,0);
	\draw[dotted, thick,magenta](12,-1)--(12,0);
	\draw[dotted, thick,magenta](14,1)--(14,0);
	\draw[dotted, thick,magenta](16,1)--(16,0);
	\draw[blue] (17.5,-.2)node[below]{$b$};
	\draw[ultra  thick, magenta] (1,0)--(2,0);
	\draw[ultra  thick, magenta] (8,0)--(10,0);

	\draw[ultra  thick, magenta] (12,0)--(14,0);
	\draw[ultra  thick, magenta] (16,0)--(17.5,0);
	\draw[blue](9,0)node[above]{$r_1$};
		\draw[magenta] (9,0)node{\Tiny{$|$}};
		\draw[blue](12,0.2)node[above]{$r_2$};
		\draw[->,green,thick](12.9,0.03)--(12.13,0.2);
		\draw(15.2,-0.5)node[below]{$r_{3}=b-\epsilon$};
		\draw[->,green, thick](16.9,-0.05)--(15.46,-0.5);
		\draw[dotted,thick] (0.5,1)--(17.5,1);
		\draw[dotted,thick] (0.5,1)--(0.5,0);
		\draw[dotted,thick] (17.5,1)--(17.5,0);
		
			\draw[dotted,thick] (0.5,-1)--(17.5,-1);
		\draw[dotted,thick] (0.5,-1)--(0.5,0);
		\draw[dotted,thick] (17.5,-1)--(17.5,0);
\end{axis}
	\end{tikzpicture}
\caption{\raggedright\Small{Let $g:[a,b]\to\mathbb{R}$ be continuosly differentiable.  Define $\widetilde{g'}:D\to\mathbb{R}$ sending  each $x\in D$ to $-|g'(x)|/g'(x)$, where $D=\{x\in(a,b):g'(x)\neq 0\}$. Also, assume that $r_0,\ldots, r_k$ are all zeros of $g'$ (in this figure, we  have  that $k=2$).  For a given small enough $\epsilon>0$, this figure shows the graph of a smooth extension  $\phi_{\epsilon}:(a,b)\to\mathbb{R}$  of a restriction of $\widetilde{g'}$, which satisfies the following  properties: 1) the support of $\phi_{\epsilon}$ is compact, 2) $\lVert\phi_{\epsilon}\rVert_{\infty}\le 1$, and 3) the integral of $\phi_{\epsilon}$ is an approximation of the integral of $\widetilde{g'}$. Let us assume that $r_{-1}=a+\epsilon$ and $r_{k+1}=b-\epsilon$. The thick segments on the $x$ axis have the  length $2\epsilon$. Also, $r_{-1}$, $r_0$, $r_1$, $r_2$,  and $r_3$ are the midpoints of the five inner thick segments. The black segments on the graph come from the graph of $\widetilde{g'}$.}}\label{smoothapproximation}
\end{figure}

Now, we talk about the multivariate total variation. Let $g:U\to\mathbb{R}$ be Riemann integrable, where $U$ is an open subset of $\mathbb{R}^n$. Then, the total variation of $g$ is defined as follows:
\[\TV_n(g):=\sup\left\{\int_U g\div(\varphi)\,\mathrm{d}\x: \varphi\in C_c^1(U,\mathbb{R}^n),\;\lVert \varphi\rVert_{\infty}\le 1\right\}.\]
Now, we justify why $\TV_n$ is a generalization of $\TV$. To do so, we have to show that $\TV_1$ and $\TV$ coincide with each other under some assumptions. Assume that $g:[a,b]\to\mathbb{R}$ is continuously differentiable. It is well-known that $\TV(g)=\int_a^b|g'|\,\d x$\footnote{One could easily prove this statement by using the triangle inequality, the mean value theorem, and the definition of the Riemann integration.} .  Consider the extra assumptions that $g\in C^2([a,b],\mathbb{R})$, and $g'$  has finitely many zeros (for the general result without these two assumptions, see Theorem \ref{maintheorem}). We show that $\TV_1(g)=\TV(g)$. Let $\varphi\in C_c^1((a,b),\mathbb{R})$ with $\lVert\varphi\rVert_{\infty}\le 1$. Then, it follows from integrating by part and $\phi(a^+)=\phi(b^-)=0$ that
\[\int_{a^+}^{b^-}g\div(\varphi)\,\d x = \int_{a^+}^{b^-}g\varphi'\,\d x =g\varphi\Big|_{a^+}^{b^-}-\int_{a^+}^{b^-} g'\varphi\,\d x=-\int_{a^+}^{b^-} g'\varphi\,\d x. \]
Hence,
\begin{equation}\label{eq1dim}
	\int_a^bg\div(\varphi)\,\d x = -\int_a^b g'\varphi\,\d x. 
\end{equation}
Now, we have that 
\[	\int_a^bg\div(\varphi)\,\d x =-\int_a^b g'\varphi\,\d x\le \left|\int_a^b g'\varphi\,\d x\right|\le \int_a^b |g'\varphi|\,\d x\le \int_a^b |g'|\,\d x.\]
The latter inequality is due to $\lVert \phi\rVert_{\infty}\le 1$. Thus, $\TV_1(g)\le \int_a^b|g'|\,\d x$.  Assume that $r_0,\ldots,r_k$ are all zeros of $g'$ such that $r_0<\cdots<r_k$. Then, for any small enough $\epsilon>0$, we assume that $r_{-1}=a+\epsilon$ and $r_{k+1}=b-\epsilon$ (see Figure \ref{smoothapproximation}). Also, we define $\phi_{\epsilon}:(a,b)\to\mathbb{R}$ in such a way that it satisfies the following conditions:
\begin{enumerate}
	\item $\phi_{\epsilon}$ equals $-|g'|/g'$ at the points $x$ with $|x-r_i|\ge\epsilon$ for each $0\le i\le k$.
	\item 
	$\phi_{\epsilon}\equiv 0$  on the intervals $[a,r_{-1}]$ and $[r_{k+1},b]$ (hence, the support of $\phi_{\epsilon}$ must be compact). 
	\item   At the other points, we  define $\phi_{\epsilon}$ to be  equal to any  continuously differentiable function $\psi_{\epsilon}$ with the following properties (see the red parts of the graph in Figure \ref{smoothapproximation}):
	\begin{itemize}
		\item $\lVert\psi_{\epsilon}\rVert_{\infty}\le 1$, and
		\item  the values of $\psi_{\epsilon}'$ at the points $r_{-1}$, $r_{k+1}$ and $r_i\pm \epsilon$ equal 0 for $0\le i\le k$. 
	\end{itemize}
\item $\phi_{\epsilon}'$ is continuous\footnote{This is easy to be done. For instance, for each interval $[r_i-\epsilon,r_{i}+\epsilon]$ with $i\neq -1, k+1$,  one could consider a polynomial $P_i$  of degree 4  with $\lVert P_i\rVert_{\infty}\le 1$ as a solution  for the following system of equations:
	\[ \begin{array}{ll}
		P_i(r_i-\epsilon)=-\frac{|g'(r_i-\epsilon)|}{g'(r_i-\epsilon)},&P_i(r_i+\epsilon)=-\frac{|g'(r_i+\epsilon)|}{g'(r_i+\epsilon)},\\
		P_i'(r_i-\epsilon)=0, &P_i'(r_i+\epsilon)=0.
	\end{array}\] 
Also, for $i=-1$ or $i=k+1$, we consider the following system of equations to define $P_i$ on the intervals $[r_{-1}, r_{-1}+\epsilon]$ and $[r_{k+1}-\epsilon, r_{k+1}]$, respectively:
	\begin{align*}
	&P_{-1}(r_{-1})=
	P_{-1}'(r_{-1})=P_i'(r_{-1}+\epsilon)=0, \quad P_i(r_{-1}+\epsilon)=-\frac{|g'(r_{-1}+\epsilon)|}{g'(r_{-1}+\epsilon)}.\\
		&P_{k+1}(r_{k+1})=
	P_{k+1}'(r_{k+1})=P_i'(r_{k+1}-\epsilon)=0, \quad P_i(r_{k+1}-\epsilon)=-\frac{|g'(r_{k+1}-\epsilon)|}{g'(r_{k+1}-\epsilon)}
\end{align*} 
By the above setting, $\phi_{\epsilon}'$ is continuous.}.
\end{enumerate}  
 Now, we have that 
 \begin{align*}
 	\int_a^b g'\phi_{\epsilon}\,\d x&=\sum_{i=0}^{k+1}\int_{r_{i-1}+\epsilon}^{r_i-\epsilon}g'\phi_{\epsilon}\,\d x+\sum_{i=0}^{k+1}\int_{r_{i}-\epsilon}^{r_i+\epsilon}g'\phi_{\epsilon}\,\d x,
 \end{align*}
while we have that
\begin{align*}
	\sum_{i=0}^{k+1}\int_{r_{i-1}+\epsilon}^{r_i-\epsilon}g'\phi_{\epsilon}\,\d x &= \sum_{i=0}^{k+1}\int_{r_{i-1}+\epsilon}^{r_i-\epsilon}-g'\frac{|g'|}{g'}\,\d x=-\sum_{i=0}^{k+1}\int_{r_{i-1}+\epsilon}^{r_i-\epsilon}|g'|\,\d x,\\
	\left|\sum_{i=0}^{k+1}\int_{r_{i}-\epsilon}^{r_i+\epsilon}g'\phi_{\epsilon}\,\d x\right|&\le \sum_{i=0}^{k+1}\int_{r_{i}-\epsilon}^{r_i+\epsilon}|g'\phi_{\epsilon}|\,\d x\le \sum_{i=0}^{k+1}\int_{r_{i}-\epsilon}^{r_i+\epsilon}|g'|\,\d x.
\end{align*}
It follows that
\begin{align*}
	\left|\int_a^b g'\phi_{\epsilon}\,\d x+\int_a^b |g'|\,\d x\right|&=\left|-\sum_{i=0}^{k+1}\int_{r_{i-1}+\epsilon}^{r_i-\epsilon}|g'|\,\d x+\sum_{i=0}^{k+1}\int_{r_{i}-\epsilon}^{r_i+\epsilon}g'\phi_{\epsilon}\,\d x+\int_a^b |g'|\,\d x\right|\\
	&= \left|\sum_{i=0}^{k+1}\int_{r_{i}-\epsilon}^{r_i+\epsilon}g'\phi_{\epsilon}\,\d x+\sum_{i=0}^{k+1}\int_{r_{i}-\epsilon}^{r_i+\epsilon}|g'|\,\d x\right|\\
	&\le \left|\sum_{i=0}^{k+1}\int_{r_{i}-\epsilon}^{r_i+\epsilon}g'\phi_{\epsilon}\,\d x\right|+\sum_{i=0}^{k+1}\int_{r_{i}-\epsilon}^{r_i+\epsilon}|g'|\,\d x\\
	&\le 2\sum_{i=0}^{k+1}\int_{r_{i}-\epsilon}^{r_i+\epsilon}|g'|\,\d x\le 4M(k+2)\epsilon,
\end{align*}
where $M$ is a bound for $g'$. 
It follows that 
\[\lim_{\epsilon\to 0}\left(-\int_a^b g'\phi_{\epsilon}\,\d x\right)=\int_a^b|g'|\,\d x.\]
Therefore, $\TV_1(g)=\TV(g)$. 
	
Now, we obtain an equality  similar to Equation (\ref{eq1dim}) in the multivariate case. To do so, let  $g\in C^1(V,\mathbb{R})$, where $V$ is an open subset of $\mathbb{R}^n$. Also, assume that $U$ is a bounded open subset of $\mathbb{R}^n$ satisfying $\oo{U}\subseteq V$, and  the boundary of $\oo{U}$ is a $C^1$ surface $\mathscr{S}$. Fix $\phi\in C_c^1(U,\mathbb{R}^n)$ with $\lVert\phi\rVert_{\infty}\le 1$. Let $\{C_i\}_{i=0}^{\infty}$ be an admissible compact cover for $U$. Then, there exists a positive integer $N_0$ such that $i\ge N_0$ implies that $\Supp(\phi)\cap \mathscr{S}_i= \emptyset$, where $\mathscr{S}_i=\bd(C_i)$ (see Proposition \ref{supp(phi)bd(c_i)}). For an $i\ge N_0$, we define $R:C_i\to\mathbb{R}^n$ by setting $R(\x)=\phi(\x)g(\x)$. Then, by the divergence theorem, we have that 
\[\int_{C_i}\div(R)\,\d \x=\int_{\mathscr{S}_i}R\cdot \widehat{N}\,\d S=0,\]
since $\phi$ and consequently $R$ are 0 on $\mathscr{S}_i$ due to $\Supp(\phi)\cap \mathscr{S}_i= \emptyset$.
Assume that $\phi=(\phi_1,\ldots,\phi_n)$. Then, we have that
\begin{align}
	\div(R)&=\sum_{i=1}^n\frac{\partial (\phi_i g)}{\partial x_i}=\sum_{i=1}^n\left(\frac{\partial \phi_i }{\partial x_i}g+\phi_i\frac{\partial g }{\partial x_i}\right)\nonumber\\
	&= g\sum_{i=1}^n\frac{\partial \phi_i }{\partial x_i}+\sum_{i=1}^n\phi_i\frac{\partial g }{\partial x_i}=g\div(\phi)+\phi\cdot\nabla g\label{divergenceofR}.
\end{align}
Now, it follows from $\int_{C_i}\div(R)\,\d \x=0$ that 
\begin{equation}\label{multivariateequation}
	\int_{C_i}g\div(\phi)\,\d \x = -\int_{C_i}\phi\cdot \nabla g\,\d \x,
\end{equation}
which implies that
\[\int_{U}g\div(\phi)\,\d \x=\lim_{i\to\infty}\int_{C_i}g\div(\phi)\,\d \x =\lim_{i\to\infty}\left(-\int_{C_i}\phi\cdot \nabla g\,\d \x\right)=-\int_{U}\phi\cdot \nabla g\,\d \x. \]
Now, we justify why $\TV_n$ somehow measures the variations of real-valued  functions with $n$ variables. 
To do so, let $n=3$. First, assume that $C$ is an infinitesimal $3$-cell. We have that 
\[\phi\cdot\nabla g\,\d x\d y\d z =\left( \phi_1\frac{\partial g}{\partial x}+\phi_2\frac{\partial g}{\partial y}+\phi_3\frac{\partial g}{\partial z}\right)\d x\d y\d z.\]
As we already discussed in Subsection \ref{Divergence Theorem}, $(\partial g/\partial x) \d x\d y\d z$, $(\partial g/\partial y) \d x\d y\d z$, and $(\partial g/\partial z) \d x\d y\d z$ are the approximated fluxes passing through the three pairs of  faces of $C$ parallel to the $yz$, $xz$, and $xy$ planes, respectively. Hence, $\phi\cdot\nabla g\, \d x\d y\d z$ is a weighted sum of the above three fluxes. Here, by considering an admissible collection $\Gamma$ of $3$-cells for $C_i$, we can say that $-\int_{C_i}\phi\cdot\nabla g\,\d x\d y\d z$ is approximately the sum of the above weighted sums varying on different cells in $\Gamma$. Note that since $\lVert\phi\rVert_{\infty}\le 1$, we have that $\phi_i(\partial g/\partial x_i)\le |\partial g/\partial x_i|$. Hence, when we maximize $-\int_{U}\phi\cdot\nabla g\,\d x\d y\d z$ to obtaint $\TV_n(g)$, in fact we are somehow measuring variations of $g$ while  moving slowly  through $U$.

\section{PEACEs of Continuous Random Vectors }\label{PEACEs of Continuous Random Vectors }
Assume that $\Omega\subseteq\mathbb{R}^n$ is open, and $W=g(\X)$, where $g:\Omega\to \mathbb{R}$ is integrable.
In this section, we define the PEACE of degree $d$ of a random vector on an output variable. Then, we justify why our definition of PEACE  is causally significant. Furthermore,  assume that $\lambda(\Gamma)$ is the PEACE of degree $d$ of $X|_{\Gamma}$ on $W$ for any open subset $\Gamma$ of $\Omega$. We show that $\lambda$  extends to a Borel regular measure on $\Omega$. 

\vspace*{0.3cm}

First of all, we define the PEACE of $\X$ on $W$ as follows:
\[\PEACE(\X\to W):=\sup\left\{\int_{\Omega} g\div(\varphi)\,\mathrm{d}\x: \varphi\in C_c^1(\Omega,\mathbb{R}^n),\; |\varphi|\le f_{\X}^2\right\},\]
where $f_{\X}$ is the probability density function of $\X$. 

Now we justify why the above formula captures causality. To do so, let  $n=3$, $\X=(X,Y,Z)$, and let $g:\oo{U}\to\mathbb{R}$ be continuously differentiable, where $U$ is an open subset of $\mathbb{R}^n$ . Then, Equation (\ref{multivariateequation}) holds, where $\{C_i\}_{i=0}^{\infty}$ is an admissible compact cover for $U$. Hence, it is enough to justify our formula  when $C_i$ is an infinitesimal  $3$-cell centered at $(x,y,z)$ whose side lengths are $\Delta  x,\Delta y$ and $\Delta z$. In this case, it follows from Equation (\ref{multivariateequation}) that
\[\int_{C_i}g\div(\phi)\,\d x\d y\d z \approx  -\phi\cdot\nabla g\,\Delta x\Delta y\Delta z,\]
while we have that 
\[\phi\cdot\nabla g\,\Delta x\Delta y\Delta z =\left( \phi_1\frac{\partial g}{\partial x}+\phi_2\frac{\partial g}{\partial y}+\phi_3\frac{\partial g}{\partial z}\right)\Delta x\Delta y\Delta z.\]
For any $y-\Delta y/2\le y'\le y+\Delta y/2$ and $z-\Delta z/2\le z'\le z+\Delta z/2$,   $(\rond g/\rond x)(x,y,z)\Delta x$ is approximately the difference between the values of $g$ at the points $(x-\Delta x/2, y',z')$ and $(x+\Delta x/2, y',z')$. Therefore, $(\rond g/\rond x)(x,y,z)\Delta x \Delta y\d z$ is the total difference considered on these two faces of $C_i$. A similar interpretation holds for the other two terms (i.e., $(\rond g/\rond y)(x,y,z)\Delta x\Delta y\Delta z$ and $(\rond g/\rond z)(x,y,z)\Delta x\Delta y\Delta z$).  Here, we have considered $X$, $Y$, and $Z$ as independent variables. Thus, all dependencies between these variables have been cut, which  is an intervention. Consequently, each of  $(\rond g/\rond x)(x,y,z)\Delta  x\Delta  y\Delta  z$, $(\rond g/\rond y)(x,y,z)\Delta  x\Delta  y\Delta  z$, and $(\rond g/\rond z)(x,y,z)\Delta  x\Delta  y\Delta  z$ reflects the causal effect obtained by intervening on the values of $X,Y$ and $Z$ with respect to three different interventional changes as follows:
\begin{enumerate}
	\item keeping $y$ and $z$ constant, and changing $x$ from $x-\Delta  x/2$ to $x+\Delta  x/2$ (corresponded to $(\rond g/\rond x)(x,y,z)\Delta  x\Delta  y\Delta  z$),
	\item 
	keeping $x$ and $z$ constant, and changing $y$ from $y-\Delta  y/2$ to $y+\Delta  y/2$ (corresponded to $(\rond g/\rond y)(x,y,z)\Delta  x\Delta  y\Delta  z$), and
	\item keeping $x$ and $y$ constant, and changing $z$ from $z-\Delta  z/2$ to $z+\Delta  z/2$ (corresponded to $(\rond g/\rond z)(x,y,z)\Delta  x\Delta  y\Delta  z$).
\end{enumerate} 
Also,   $-\phi_1,-\phi_2$ and $-\phi_3$ are probabilistic weights for these three single causal effects.  We note that
\begin{small}\[\nabla g\cdot \phi\, \Delta  x\Delta  y\Delta  z \le |\nabla g||\phi|\Delta  x\Delta  y\Delta  z=\sqrt{\left(\frac{\rond g}{\rond x}|\phi|\right)^2+\left(\frac{\rond g}{\rond y}|\phi|\right)^2+\left(\frac{\rond g}{\rond z}|\phi|\right)^2}\Delta  x\Delta  y\Delta  z.\]\end{small}
Hence, $(\rond g/\rond x) \Delta  x\Delta  y\Delta  z\approx \left(g(x+\Delta  x/2,y,z)-g(x-\Delta  x/2,y,z)\right)\Delta  y \Delta  z$.
 Selecting each component of this difference has approximately the density $f_X(x,y,z)$, and hence when we randomly and independently select both components, we have the density $f_X^2$. Hence,  we can assume that $|\phi|\le f_X^2$ (in the proof of Theorem \ref{maintheorem}, we will see that the aforementioned supremum is obtained when $\phi$ equals $(\nabla g/|\nabla g|)f_X^2$ on the points with $\nabla g\neq 0$).  

We define the PEACE of degree $d$ of $\X$ on $W$ as follows:
\[\PEACE_d(\X\to W):=\sup\left\{\int_{\Omega} g\div(\varphi)\,\mathrm{d}\x: \varphi\in C_c^1(\Omega,\mathbb{R}^n),\; |\varphi|\le f_{\X}^{2d}\right\}.\]
Here, when $d$ increases, the  points with higher density have a greater contribution in determining the value of PEACE. Hence, we will have a spectrum for the values of PEACE. Since, $\phi\equiv 0$ is in $C_c^1(\Omega,\mathbb{R}^n)$, the value 0 belongs to the above set, and hence $\PEACE_d(\X\to W)\ge 0$.  In the following proposition, by considering countable families of pair-wise disjoint open subsets of $\Omega$, we show that PEACE is countably additive. 
\begin{Proposition}\label{finiteadditive}
	Let $\{\Omega_i\}_{i=0}^{\infty}$ be a family of pair-wise disjoint open subsets of $\mathbb{R}^n$, and $\Omega=\bigcup_{i=0}^{\infty}\Omega_i$. Then, we have that
\[\PEACE_d(\X\to Y)=\sum_{i=0}^{\infty}\PEACE_d(\X|_{\Omega_i}\to Y).\]
\end{Proposition}
\begin{proof}
See Appendix \ref{other}.
\end{proof}
In the following,  by considering countable families of open subsets of $\Omega$, we show that PEACE is countably subadditive
\begin{Proposition}\label{subadditivityofopensets}
		Let $\{\Omega_i\}_{i=0}^{\infty}$ be a family of  open subsets of $\mathbb{R}^n$, and $\Omega=\bigcup_{i=0}^{\infty}\Omega_i$. Then, we have that
	\[\PEACE_d(\X\to Y)\le\sum_{i=0}^{\infty}\PEACE_d(\X|_{\Omega_i}\to Y).\]
\end{Proposition}
\begin{proof}
See Appendix \ref{other}.
\end{proof}

Now, we show that PEACE is monotone.
\begin{Proposition}\label{monoton}
Let $\Omega$ and $\Gamma$ be two open subsets of $\mathbb{R}^n$ with $\Gamma\subseteq\Omega$.  Then, \[\PEACE_d(\X|_{\Gamma}\to Y)\le\PEACE_d(\X\to Y).\]
\end{Proposition}
\begin{proof}
	See Appendix \ref{other}.
\end{proof}
In the following theorem, we show that PEACE on open subsets of $\Omega$ can be extended to a Borel regular measure.
\begin{Theorem}\label{outermeasure}
Denote the set of all open subsets of $\Omega$ by $\tau$. Define $\lambda:\tau\to[0,\infty]$ by $\lambda(\Gamma)=\PEACE_d(\X|_{\Gamma}\to Y)$. Then, define $\mu:\mathcal{P}(\Omega)\to [0,\infty]$ by
\[\mu(E):=\inf\{\lambda(\Gamma): E\subseteq\Gamma,\;\Gamma\in \tau\}.\]
Then, $\mu$ is an outer measure. Moreover, $\mu$ is a Borel regular measure. 
\end{Theorem}
\begin{proof}
See Appendix \ref{other}.
\end{proof}
\begin{Remark}\label{mu=lambda}
	Considering Theorem \ref{outermeasure}, one could use Proposition \ref{monoton} to show that $\mu=\lambda$ on $\tau$.
\end{Remark}
In the following theorem, we show that how continuously differentiability of $g$ could lead us to a handy and easily computable formula for  PEACE.
\begin{Theorem}\label{maintheorem}
	Let $Y=g(\X)$ with $g\in C^1(\oo{\Omega},\mathbb{R})$,  where $\Omega$ is a bounded open subset of $\mathbb{R}^n$. 
	Then, we have that
	\[\PEACE_d(\X\to Y)=\int_{\Omega}\left|\nabla g(\x)\right|f_X^{2d}(\x)\,\d \x.\]
\end{Theorem}
\begin{proof}
	See Appendix \ref{appendixmaintheorem}.
\end{proof}
\begin{Corollary}\label{unboundedmain}
Theorem \ref{maintheorem} holds even when $\Omega$ is not bounded. 
\end{Corollary}
\begin{proof}
See Appendix \ref{appendixmaintheorem}.
\end{proof}
The following theorem supports that PEACE is causally significant.
\begin{Corollary}\label{lastcor}
	Let $Y=g(\X)$ with $g\in C^1(\overline{\Omega},\mathbb{R})$, where $\Omega$ is an open subset of $\mathbb{R}^n$. Also, let $\Gamma$ be the the set of all points $\x\in\Omega$ with $f_{\X}(\x)\neq 0$. Then, $\PEACE_d(\X\to Y)=0$ if and only if $g|_{\Gamma}$ is locally constant. Consequently, if in addition $\Omega$ is connected, then $\PEACE_d(\X\to Y)=0$ if and only if $g|_{\Gamma}$ is constant.
\end{Corollary}
\begin{proof}
	See Appendix \ref{appendixmaintheorem}.
\end{proof}
Now, we introduce the notation $g_{in}$.  In fact, when we use the notation $g_{in}$ against $g$, we mean that the domain of $g_{in}$ is an $n$-cell $C$, while the domain of $g$ is an open subset $\Omega$ of $C$ defined by some functional relationships. Hence, all variables in $C$ are independent, while this might not be true for the variables in $\Omega$. In other words, $g_{in}$ is obtained from $g$ after removing all functional relationships defining $\X$ and $\Z$. The index $in$ here comes from the word ``intervention''. 

Now, let $Y=g(\X,\Z)$ with $g:\Omega\to\mathbb{R}$, where $\X=(X_1,\ldots,X_n)$ and $\Z=(Z_1,\ldots,Z_m)$, and $\Omega\subseteq \mathbb{R}^{n+m}$ is open. Assume that $\pi:\mathbb{R}^{n+m}\to\mathbb{R}^n$ is the projection of $\mathbb{R}^{n+m}$ on $\mathbb{R}^n$ (i.e., $\pi(\x,\z)=\x$). 
For a fixed $\Z=\z$, we denote the probability density function of $\X$ given $\Z$ by $f_{\X|\Z}$. We intend to formulate the PEACE of degree $d$ of $\X$ on $Y$. To do so, we define the \textit{probabilistic interventional easy variation of degree} $d$ of $\X$ on $Y$ keeping $\Z=\z$ as follows:
\[\PIEV_d^{\z}(\X\to Y):=\sup\left\{\int_U g_{in}(\x,\z)\div(\varphi)(\x)\,\mathrm{d}\x: \varphi\in C_c^1(U,\mathbb{R}^n),\; |\varphi|\le f_{\X|\Z}^{2d}(\,\cdot\,,\z)\right\},\]
where $U=\pi(\Omega)$. 
Then, we define the PEACE of degree $d$ of $\X$ on $Y$ as follows:
\[\PEACE_d(\X\to Y):=4^d\E_{\Z}(\PIEV_d^{\z}(\X\to Y)).\]
We note that all previous results about PEACE for the functions of the form of $Y=g(\X)$ hold for $\PIEV_d^{\z}(\X\to Y)$ as well. The changes required in the aforementioned results are to replace $f_{\X}^{2d}$ with $f_{\X|\Z}^{2d}(\,\cdot\,,\z)$, $\Omega$ with the domain of $\X$, and  $g$ with $(g_{in})_{\z}:U\to\mathbb{R}$ defined by setting $(g_{in})_{\z}(\x)=g_{in}(\x,\z)$.

In the following proposition, we investigate the effect of the change of variables on the PEACE formula. Especially, we show that isometry change of variables do not affect the value of PEACE. 
\begin{Proposition}\label{iso}
	Let $Y=g(\X,\Z)$  with $g\in C^1(\overline{\Omega},\mathbb{R})$ and $\X=h(\W)$, where 
	$\Omega$ is an open subset of $\mathbb{R}^{n+m}$,  $\mathbf{W}\in \mathbb{R}^n$, and $h$ is a diffeomorphism. Then 
	\begin{enumerate}
		\item In general, we have that
	\begin{small}	\[
		\PIEV_d^{\z}(\mathbf{W}\to Y)=\int_{\Gamma}\left|\mathrm{Jac}(h)(h^{-1}(\x))\frac{\partial g_{in}}{\partial \x}(\x,\z) \right|\left(\left|\det\left(\mathrm{Jac}(h)(h^{-1}(\x))\right)\right|f(\x|\z)\right)^{2d}\,\d\w.
		\]\end{small}
		\item
		 If $h$ is an affine function defined by $\X=h(\W)=A\mathbf{W}+\mathbf{a}$, where $A$ is an $n\times n$ invertible matrix whose entries are real numbers, then 
		 \[	\PIEV_d^{\z}(\mathbf{W}\to Y)=\int_{\Gamma}\left|A^T\frac{\partial g_{in}}{\partial \x}(\x,\z) \right|\left(\left|\det(A)\right|f(\x|\z)\right)^{2d}\,\d\w.\]
	\item If $h$ is a restriction of an onto isometry of $\mathbb{R}^n$, then \[\PEACE_d(\mathbf{W}\to Y)=\PEACE_d(\mathbf{X}\to Y).\]
			\end{enumerate}
\end{Proposition}
\begin{proof}
	See Appendix \ref{other}.
\end{proof}

\section{A New Formula for Total Variation}\label{New Formula for Total Variation}
To define the PEACE of an outcome random variable with respect to a  discrete random vector, first, we need a definition for the total variation of a multivariate discrete function. 
The total variation of a discrete function of two variables has been under a hot discussion in the recent decades (see \cite{ abergel2017shannon, chambolle2005total,rudin1992nonlinear}), since each  photo, in the classic sense, could be seen as such a function. Let $\Omega\subseteq\mathbb{Z}^2$ and $g:\Omega\to\mathbb{R}$, where $\mathbb{Z}$ is the set of integers. Then, the total variation of $g$ could be naturally defined as follows:\footnote{These two formulas are used for image denoising.}
\begin{align*}
	\TV(g)&:=\sum_{(k,l)\in\Omega}\sqrt{\rond _1g(k,l)^2+\rond_2g(k,l)^2},\quad \TV_{\text{ani}}(g):=\sum_{(k,l)\in\Omega}|\rond _1g(k,l)|+|\rond_2g(k,l)|,
\end{align*}
where
\begin{align*}
	\rond _1g(k,l)&=g(k+1,l)-g(k,l),\quad \rond _2g(k,l) = g(k,l+1)-g(k,l).
\end{align*}
Note that if we try to reformulate the above definitions which could be used for any domain $\Omega\subseteq\mathbb{R}^2$ and a continuously differentiable function $g:\Omega\to \mathbb{R}$, then we cannot reach the true formula, which is $\int_{\Omega}|\nabla g(\x)|\,\d\x$. To see this, for a partition $\{(x_k,y_l):0\le k\le n,\;0\le l\le n'\}$ for $\Omega$, by the mean value theorem for $g$, for any $k$ and $l$, there exist $\theta_k\in(x_{k},x_{k+1})$ and $\eta_l\in (y_{l},y_{l+1})$ in such a way that
\[\sqrt{\rond _1g(x_k,y_l)^2+\rond_2g(x_k,y_l)^2}=\sqrt{\left(\frac{\partial g}{\partial x}(\theta_k,y_l)\right)^2\Delta x_k+\left(\frac{\partial g}{\partial y}(x_k,\eta_l)\right)^2\Delta y_l}.\]
Hence, $\Delta x_k,\Delta y_l\to 0$, does not yield that $\TV(g)$ is the integral mentioned above. A similar argument shows that $\TV_{\text{ani}}(g)$ does not equal  the above integral as well. In \cite{abergel2017shannon}, a complicated alternative definition has been proposed for the total variation of a discrete function applicable in image denoising. This definition comes from an interpolation of a discrete function, and hence it is a continuous point of view for the discrete total variation of a function. Note that this definition   yields an integral such as the above. 

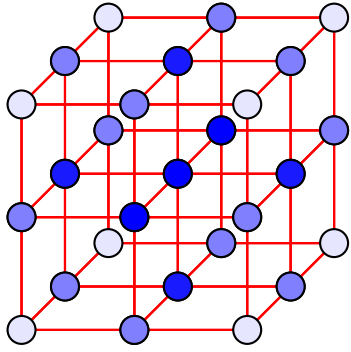
\begin{figure}
\begin{tikzpicture}[scale=1.5, thick]  
	
	\foreach \x in {0,1,2}
	\foreach \y in {0,1,2}
	\foreach \z in {0,1,2}
	{
		\draw[thick,red,opacity=1] (\x,0,\z) -- (\x,2,\z);
		\draw[thick,red,opacity=1] (0,\y,\z) -- (2,\y,\z);
		\draw[thick,red,opacity=1] (\x,\y,0) -- (\x,\y,2);
	}
	\foreach \x in {0,1,2}
	\foreach \y in {0,1,2}
	\foreach \z in {0,1,2}
	{\draw[fill=blue!10] (\x,\y,\z) circle (0.3em);}    
	
	\foreach \x/\y in {1/0,2/1,1/2,0/1}
	\foreach \z in {0,2}
	{\draw[fill=blue!50] (\x,\y,\z) circle (0.3em);}
	
	\foreach \x/\y in {2/0,2/2,0/2,0/0}
	\foreach \z in {1}
	{\draw[fill=blue!50] (\x,\y,\z) circle (0.3em);}
	
	\foreach \x/\y in {1/0,2/1,1/2,0/1}
	\foreach \z in {1}
	{\draw[fill=blue!90] (\x,\y,\z) circle (0.3em);}
	
	\foreach \x/\y in {1/1}
	\foreach \z in {0,1,2}
	{\draw[fill=blue] (\x,\y,\z) circle (0.3em);}

\end{tikzpicture}
\caption{The set of the vertices of each of the small cubes and their faces are a  cube-like and a face-like, respectively. }\label{face-like}
\end{figure}

 In this section, we provide a new formula for the total variation of a multivariate discrete function compatible with the continuous definition. 
To do so, we use the idea of the flux of a function that we previously used in the continuous case. 
\uline{In this section and Section \ref{PEACE in Discrete Case}, we study the PEACE of $\X$ on $Y$ when $Y=g(\X)$. However, all the content we discuss is satisfied when we have that $Y=g(\X,\Z)$. Here, we  need to replace $\PEACE_d(\X\to Y)$ with $\mathcal{N}\PIEV_d^{\z}(\X\to Y)=4^d\PIEV_d^{\z}(\X\to Y)$, and define $\PEACE_d(\X\to Y)=\E_{\Z}\left(\mathcal{N}\PIEV_d^{\z}(\X\to Y)\right)$. Also, we need to replace $g$ with $g_{in}$ and naturally modifying some notations (e.g., replacing $\nabla g$ with $\rond g_{in}/\rond \x$)}.
 Let $Y=g(\X)$, where $\X=(X_1,\ldots,X_n)$, and $\Supp(X_i)=\{x_{i1}, \ldots, x_{in_i}\}$, where $x_{i1}<\cdots<x_{in_i}$ for any $i$.  Let $2\le j_i\le n_{i}$ for any $i$. Then, by a \textit{discrete-like} $n$-cube $C(x_{1j_1},\ldots,x_{nj_n})$, we mean the set of points $(\alpha_1,\ldots,\alpha_n)$, where $\alpha_i\in\{x_{ij_i}, x_{i,j_i-1}\}$ for any $i$. For any $1\le i\le n$, a \textit{face-like} $F(x_{1j_1},\ldots,x_{nj_n};x_{ij_i})$ of $C(x_{1j_1},\ldots,x_{nj_n})$ is obtained when we fix $\alpha_i=x_{ij_i}$. Similarly, the face-like $F(x_{1j_1},\ldots,x_{nj_n};x_{i,j_i-1})$ is defined, when we fix $\alpha_i=x_{i,j_i-1}$. We call the above two face-likes  the \textit{parallel } face-likes (see Figure \ref{face-like}).
Let $\Gamma$ be the set of all discrete-like $n$-cubes obtained from $\prod_{i=1}^n\Supp(X_i)$.  Let $\phi:\Gamma\to\mathbb{R}^n$ be an arbitrary function with  $|\phi|\le1$. For simplicity, assume that $\phi_{j_1,\ldots,j_n}=\phi\left(C(x_{1j_1},\ldots,x_{nj_n})\right)$, and $\phi_{j_1,\ldots,j_n}=\left(\phi_{j_1,\ldots,j_n}^{(1)},\ldots, \phi_{j_1,\ldots,j_n}^{(n)}\right)$. Now, we define the $\phi$-flux of $g$ passing through these two parallel face-likes, denoted by $\mathrm{Flux}_{\phi}^{(i)}(x_{1j_1},\ldots,x_{nj_n})$ as follows:
\[\mathrm{Flux}_{\phi}^{(i)}(x_{1j_1},\ldots,x_{nj_n}):=\mathrm{dif}_g^{(i)}(C(x_{1j_1},\ldots,x_{nj_n}))\phi_{j_1,\ldots,j_n}^{(i)}\mathrm{Vol}(F(x_{1j_1},\ldots,x_{nj_n};x_{ij_i})),\]
 where
 \[\mathrm{dif}_g^{(i)}(C(x_{1j_1},\ldots,x_{nj_n})):=\frac{1}{2^{n-1}}\left(\sum_{\substack{\x\in F(x_{1j_1},\ldots,x_{nj_n};x_{ij_i})\\x'_i=x_{i,j_i-1},\;\forall\,k\neq i\;x'_k=x_k}}\bigl(g(\x)-g(\x')\bigr)\right)\]
 is the mean of the differences of $g$ on corresponding points of  a pair of parallel face-likes, 
 $\mathrm{Vol}(F(x_{1j_1},\ldots,x_{nj_n};x_{ij_i}))=\left(\prod_{k=1}^n\Delta x_{kj_k}\right)/\Delta x_{ij_i}$, and $\Delta x_{kj_k}=x_{kj_k}-x_{k,j_k-1}$ for any $1\le k\le n$. Next, we define the $\phi$-flux of $g$ passing through $C(x_{1j_1},\ldots,x_{nj_n})$ as the sum of $\phi$-fluxes of $g$ passing through all pairs of parallel face-likes of  $C(x_{1j_1},\ldots,x_{nj_n})$:
 \begin{equation*}
 	\mathrm{Flux}_{\phi}(x_{1j_1},\ldots,x_{nj_n}):=\sum_{i=1}^n \mathrm{Flux}_{\phi}^{(i)}(x_{1j_1},\ldots,x_{nj_n}).
  \end{equation*}
 Finally, we define the $\phi$-variation and the total variation of $g$ as follows, respectively:
 \[\TV_{\phi}(g):=\sum_{j_1,\ldots,j_n}\mathrm{Flux}_{\phi}(x_{1j_1},\ldots,x_{nj_n}),\quad\TV(g):=\sup_{\phi}\TV_{\phi}(g).\]
\begin{Proposition}\label{totalvariation-discrete}
	Considering the above assumptions, define 
	\[\mathrm{Flux}^{(i)}(x_{1j_1},\ldots,x_{nj_n}):=\mathrm{dif}_g^{(i)}(C(x_{1j_1},\ldots,x_{nj_n}))\mathrm{Vol}(F(x_{1j_1},\ldots,x_{nj_n};x_{ij_i})).\]
	Then, if
	\begin{align*}	\mathrm{Flux}(x_{1j_1},\ldots,x_{nj_n})&:=\sqrt{\sum_{i=1}^n \mathrm{Flux}^{(i)}(x_{1j_1},\ldots,x_{nj_n})^2},
\end{align*}
we have that
	\begin{align*} \TV(g)&=\sum_{j_1,\ldots,j_n}^n\mathrm{Flux}(x_{1j_1},\ldots,x_{nj_n}).
\end{align*}
\end{Proposition}
\begin{proof}
	See Appendix \ref{other}.
\end{proof}
 \begin{Remark}
 	One could see that for $n=1$, this definition of total variation coincides with the previous definition given in \cite{faghihi2022probabilistic}\footnote{Let $Y=g(X)$, where $X$ is a discrete random variable for which $\Supp(X)=\{x_0,\ldots,x_l\}$ with $x_0<\cdots<x_l$. Then, the total variation of $g$ is defined to be the total variation of the sequence $g(x_0),\ldots,g(x_l)$, namely, $\sum_{i=1}^l|g(x_i)-g(x_{i-1})|$.}.
 \end{Remark}
The above definition of total variation could be generalized to the continuous case as well. Let $Y=g(\X)$ with $g:\Omega\to\mathbb{R}$, where $X_1,\ldots,X_n$ are continuous random variables, and $\Omega$ is an open subset of $\mathbb{R}^n$. Assume that $\{C_j\}_{j=0}^{\infty}$ is an admissible compact cover for $\Omega$. Let $\phi\in C_c^1(C,\mathbb{R}^n)$ with $\lVert\phi\rVert_{\infty}\le 1$ and $\phi=(\phi^{(1)},\ldots,\phi^{(n)})$. There exists $N_0$ such that $j\ge N_0$ implies that $\Supp(\phi)\subseteq C_{j}$. Fix $j\ge N_0$, and let $C_j\subseteq  C=[a_1,b_1]\times\cdots\times[a_n,b_n]$. A partition for $C$ is obtained as $P_1\times\cdots\times P_n$, where $P_i\in\mathcal{P}([a_i,b_i])$ for any $i$. We denote the set of all partitions of $C$ by $\mathcal{P}(C)$. For any $P\in\mathcal{P}(C)$, set
$P=\prod_{i=1}^n\{x_{i0}^{(P)},\ldots, x_{in_i}^{(P)}\}$.  Now, we define $\mathrm{Flux}_{\phi}^{(i)}(x_{i0}^{(P)},\ldots, x_{in_i}^{(P)}; \x_{j_1,\ldots,j_n}^{(P)})$ as follows
\[\mathrm{dif}_g^{(i)}(C(x_{i0}^{(P)},\ldots, x_{in_i}^{(P)}))\phi^{(i)}(\x_{j_1,\ldots,j_n}^{(P)})\mathrm{Vol}(F(x_{i0}^{(P)},\ldots, x_{in_i}^{(P)};x_{ij_i}^{(P)})),\]
where $\x_{j_1,\ldots,j_n}^{(P)}$ is an arbitrary point in $C(x_{i0}^{(P)},\ldots, x_{in_i}^{(P)})$. Next, we define
\[\mathrm{Flux}_{\phi}(x_{i0}^{(P)},\ldots, x_{in_i}^{(P)};\x_{j_1,\ldots,j_n}^{(P)}):=\sum_{i=1}^n\mathrm{Flux}_{\phi}^{(i)}(x_{i0}^{(P)},\ldots, x_{in_i}^{(P)};\x_{j_1,\ldots,j_n}^{(P)})\]
as before. Finally, we define
the \textit{discrete-like}  $\phi$-variation and  total variation of $g$ as follows, respectively:
\begin{align*}
	\TV^{\mathrm{dis}}_{\phi}(g)&:=\lim_{\lVert P\rVert\to 0}\sum_{j_1,\ldots,j_n}\mathrm{Flux}(x_{1j_1}^{(P)},\ldots,x_{nj_n}^{(P)};\x_{j_1,\ldots,j_n}^{(P)}),\\
		\TV^{\mathrm{dis}}(g)&:=\sup_{\phi}	\TV^{\mathrm{dis}}_{\phi}(g),
\end{align*}
where $\lVert P\rVert$ is the maximum of the diameters of the discrete-like $n$-cubes of $P$. 
One could see that if $g$ is continuously differentiable or a Lipschitz function\footnote{$g$ is called a Lipschitz function if there exists $M>0$ in such a way that for any $\x,\x'\in\Omega$, we have that $|g(\x)-g(\x')|\le M|\x-\x'|$.}, then the above definitions of $\phi$-variation and total variation are well-defined (they do not  depend on the choice of $\x_{j_1,\ldots,j_n}$ and the admissible compact cover for $\Omega$). 
\section{PEACE in Discrete Case}\label{PEACE in Discrete Case}
In this section, we define the PEACE formula for discrete random vectors. To do so, we use the total variation formula that we defined in Section \ref{New Formula for Total Variation}, which is based on the idea of the flux of a function and the definition of the PEACE formula in the continuous case. 

Let $\phi:\Gamma\to\mathbb{R}^n$ with $|\phi|\le w_{\X,d}$, where $\Gamma$ is the set of all discrete-like $n$-cubes of $\prod_{i=1}^n\Supp(X_i)$, and $w_{\X,d}:\Gamma\to\mathbb{R}$ is defined by  \[\omega_{\X,d}(C(x_{1j_1},\ldots,x_{nj_n}))=\frac{4^d}{n}\sum_{i=1}^n\frac{1}{2^{n-1}}\left(\sum_{\substack{\x\in F(x_{1j_1},\ldots,x_{nj_n};x_{ij_i})\\x'_i=x_{i,j_i-1},\;\forall\,k\neq i\;x'_i=x_i}}\P_{\X}^d(\x)\P_{\X}^d(\x')\right).\]
Intiuitively, $\omega_{\X,d}(C(x_{1j_1},\ldots,x_{nj_n}))=4^d\E\left(\E_{\X_{\mathrm{Ufm}}}\left(\P_{\X}(\X)^d\P_{\X}(\X')^d\right)\right)$, where the random variable $\X_{\mathrm{Ufm}}$ is $\X|_{F(x_{1j_1},\ldots,x_{nj_n};x_{ij_i})}$ that uniformly takes its values. We note that $\P_{\X}(\x)\P_{\X}(\x')$ is the probability of randomly and independently selecting $\x$ and $\x'$, respectively, that could be thought of the availability of the interventional change $g(\x)-g(\x')$. Also, the outer expected value in the above formula for $\omega_{\X,d}$ is a uniform average over the set of parallel face-likes of $C(x_{1j_1},\ldots,x_{nj_n})$. Now, considering the above notations, we define the \textit{probabilistic $\phi$-flux ($\mathrm{PFlux}_{\phi})$ of $g$} passing through the parallel face-likes $F(x_{1j_1},\ldots,x_{nj_n};x_{ij_i})$ and $F(x_{1j_1},\ldots,x_{nj_n};x_{i,j_i-1})$, as follows:
\[\mathrm{PFlux}_{\phi}^{(i)}(x_{1j_1},\ldots,x_{nj_n}):=\mathrm{dif}_g^{(i)}(C(x_{1j_1},\ldots,x_{nj_n}))\phi_{j_1,\ldots,j_n}^{(i)}\mathrm{Vol}(F(x_{1j_1},\ldots,x_{nj_n};x_{ij_i})),\]
 We also define
 \begin{equation*}
	\mathrm{PFlux}_{\phi}(x_{1j_1},\ldots,x_{nj_n}):=\sum_{i=1}^n \mathrm{PFlux}_{\phi}^{(i)}(x_{1j_1},\ldots,x_{nj_n}).
\end{equation*}

 Finally, we define the $\phi$-PEACE and the PEACE of degrees $d$ of $\X$ on $Y$ as follows, respectively:
 \begin{align*}
 	\PEACE_d^{\phi}(\X\to Y)&:=\sum_{j_1,\ldots,j_n}\mathrm{PFlux}_{\phi}(x_{1j_1},\ldots,x_{nj_n}),\\
 	\PEACE_d(\X\to Y)&:=\sup_{\phi}\PEACE_d^{\phi}(\X\to Y).
 \end{align*}
 The following proposition could be shown similarly to the proof of Proposition \ref{totalvariation-discrete}.
 \begin{Proposition}\label{peace-discerete}
 	Considering the above assumptions, define $\mathrm{PFlux}^{(i)}(x_{1j_1},\ldots,x_{nj_n})$ as follows:
 	\[\omega_{\X,d}(C(x_{1j_1},\ldots,x_{nj_n}))\mathrm{dif}_g^{(i)}(C(x_{1j_1},\ldots,x_{nj_n}))\mathrm{Vol}(F(x_{1j_1},\ldots,x_{nj_n};x_{ij_i})).\]
 	Then, if
 	\begin{align*}	\mathrm{PFlux}(x_{1j_1},\ldots,x_{nj_n})&:=\sqrt{\sum_{i=1}^n \mathrm{PFlux}^{(i)}(x_{1j_1},\ldots,x_{nj_n})^2},
 	\end{align*}
 	we have that
 	\begin{align*} \PEACE_d(\X\to Y)&=\sum_{j_1,\ldots,j_n}^n\mathrm{PFlux}(x_{1j_1},\ldots,x_{nj_n}).
 	\end{align*}
 \end{Proposition}
 In the case that $\X$ and $Y$ are continuous random variables, using an admissible compact cover for $\Omega$ (as in Section \ref{New Formula for Total Variation}), we define the discrete-like $\phi$-PEACE and PEACE as follows, respectively:
\begin{align*}
	\PEACE_d^{\phi,\mathrm{dis}}(\X\to Y)&:=\lim_{\lVert P\rVert\to 0}\sum_{j_1,\ldots,j_n}\mathrm{PFlux}_{\phi}(x_{1j_1}^{(P)},\ldots,x_{nj_n}^{(P)};\x_{j_1,\ldots,j_n}^{(P)}),\\
	\PEACE_d^{\mathrm{dis}}(\X\to Y)&:=\sup_{\phi}\PEACE_d^{\phi,\mathrm{dis}}(\X\to Y),
\end{align*}
where $\x_{j_1,\ldots,j_n}^{(P)}$ is an arbitrary point in $C(x_{1j_1}^{(P)},\ldots,x_{nj_n}^{(P)})$. Similar to the discussion in Section \ref{New Formula for Total Variation}, one could see that if $g$ is continuously differentiable or a Lipschitz function, then the above definitions of $\phi$-PEACE and PEACE are well-defined. 
Note that, for the continuous case, we assume that $\phi\in C_c^1(C,\mathbb{R}^n)$ with $|\phi|\le f_{\X}^{2d}$. The previous assumption is a replacement for the assumption $|\phi|\le \omega_{\X,d}$ in the discrete case\footnote{When we decrease the diameter of a cube to 0, then we can assume that all probability values in the definition of $\omega_{\X,d}$ are equal to $f_{\X}^d(\x_{j_1,\ldots,j_n})$, and hence in the continuous case, $\omega_{\X,d}\approx f_{\X}^{2d}$.}.

In the following theorem, we show that our definitions of PEACE in discrete and continuous cases are compatible. 
\begin{Theorem}\label{dis-con}
	Let $Y=g(\X)$ with $g\in C^1(\oo{\Omega},\mathbb{R})$, where $\Omega$ is a bounded open subset of $\mathbb{R}^n$.  Then, 
	\[\PEACE^{\mathrm{dis}}_d(\X\to Y)=\PEACE_d(\X\to Y)=\int_{\Omega}|\nabla g|f_{\X}^{2d}\,\d\x.\]
\end{Theorem}
\begin{proof}
	See Appendix \ref{other}.
\end{proof}
\section{Identifiability of PEACE}\label{Identifyability of PEACE}
Let $\mathcal{F}$ be a formula associating a quantity $\mathcal{F}(\mathcal{S})$ to an SEM $\mathcal{S}$. Roughly speaking, we say that $\mathcal{F}$  is identifiable under the assumption $\mathcal{A}$, if $\mathcal{F}(\mathcal{S})$ could be computed only by using observed variables for any SEM $\mathcal{S}$ under the assumption $\mathcal{A}$ (see \cite{faghihi2022probabilistic}). Let $Y=g(\X,\Z,U_Y)$, where $U_Y$ is an unobserved  random variable. We say that $g_{in}$ is separable with respect to $\Z$ if we can write $g(\X,\Z,U_Y)=g^{(1)}(\X,\Z)+g^{(2)}(\Z,U_Y)$. One could see that if $g$ has the partial dervative with respect to $\X$, then $g_{in}$ is separable with respect to $\Z$ if and only if $\left(\rond g_{in}/\rond\X\right)(\X,\Z,U_Y)$ is a function of $\X$ and $\Z$. From this fact, and the proof of \cite[Theorem 4.19]{faghihi2022probabilistic}, one could show the following theorem and its corollary:
\begin{Theorem}
	Let $Y=g(\X, \Z, U_Y )$, and $g_{in}$ has the partial derivative with respect to $\X$, and the following conditions are satisfied for any $\x\in\Supp(\X)$ and $\z\in\Supp(\Z)$:
	\begin{enumerate}
		\item
	 $g_{in}(\X, \Z, U_Y )$ is separable with respect to $\Z$,  and
	 	\item 
	 Given $U_Y$, the random variables $\X$ and $\Z$ are independent.
	\end{enumerate}
Then,  $\PEACE_d(\X\to Y)$ is identifiable. 
Besides, if we have the following additional assumptions for any $\x\in\Supp(\X)$ and $\z\in\Supp(\Z)$:
\begin{enumerate}
	\setcounter{enumi}{2}
		 \item Given $\Z$, $Y_{\x,\z}$ and $\X$ are independent,
	 \item $Y_{\x,\z}$ is a one-to-one function of $(U_Y )_{\x,\z}$, and 
	\item $Y_{\x,\z}$ and $\Z$ are independent,
\end{enumerate}
then, we have that
\begin{align*}
	\PIEV_d^{(\z,u_y)}(\X\to Y)&=\int_{\Supp(\X)}\left|\frac{\rond \E(Y|\X,\Z)}{\rond\X}(\x,\z)\right|f^{2d}(\x|\z)\,\d\x,\\
	\PEACE_d(\X\to Y)&= \E_{\Z}\left(	\PIEV_d^{(\z,u_y)}(\X\to Y)\right).
\end{align*}
\end{Theorem}
\begin{Corollary}
	Let $Y=\X\boldsymbol{\alpha}+\Z\boldsymbol{\beta}+\gamma U_Y$, where $\gamma\neq 0$.  If $\Z$ and $\X$ are independent given $U_Y$, then $\PEACE_d(\X\to Y)$ is identifiable, and we have that
	\[\quad\PIEV_d^{(\z,u_Y)}(\X\to Y)=|\boldsymbol{\alpha}|\int_{\Supp(\X)}f^{2d}(\x|\z)\,\d\x.\]
	Further, if the following assumptions hold for any $\x\in\Supp(\X)$ and $\z\in\Supp(\Z)$:
	\begin{itemize}
		\item Given $\Z$, $Y_{\x,\z}$ and $\X$ are independent, and
		\item $Y_{\x,\z}$ and $\Z$ are independent,
	\end{itemize}
then, we have that
\[ |\boldsymbol{\alpha}|=\left|\frac{\rond \E(Y|\X,\Z)}{\rond\X}\right|.\]
\end{Corollary}
\section{Positive and Negative PEACEs}\label{Positive and Negative PEACEs}
In \cite[Section 4.6]{faghihi2022probabilistic}, we defined the positive and the negative PEACEs in the discrete case. Now, we discuss these concepts in the continuous case. 

Let $X$ and $\Z$ be a random variable and a random vector, respectively. Also, let $Y=g(X,\Z)$, where $\Supp(X)\subseteq [a,b]$. Similar to the discrete case,  by the positive/negative interventional variation of $X$ on $Y$, we mean only to  account for the positive interventional changes of $Y$  due to the increase in the value of $X$ and keeping $\Z$ constant. The \textit{positive/negative probabilistic  interventional easy variation}  of $Y$ with respect to $X$  has a similar interpretation. Indeed, for $\bm{\epsilon}\in\{\bm{\pm} \}$, we define
\begin{align*}
	\PIEV_d^{\z}(X\to Y)^{\bm{\epsilon}}&:=\lim_{\lVert P\rVert\to 0}L_{P,d}^z(X\to Y)^{\bm{\epsilon}},\quad P\in\mathcal{P}([a,b]),\\
	L_{P,d}^z(X\to Y)^{\bm{\epsilon}}&:=\sum_{i=1}^{n_P}\left(g(x_i^{(P)})-g(x_{i-1}^{(P)})\right)^{\bm{\epsilon}}f(x_i^{(P)}|\z)^df(x_{i-1}^{(P)}|\z)^d.
\end{align*}
\begin{Theorem}\label{NPEV}
Assume that $Y=g(X,\Z)$ has the continuous partial derivatives with respect to $X$, and $\Supp(X)\subseteq[a,b]$. Then, we have that 
	\[\PIEV_d^{\z}(X\to Y)^{\bm{\epsilon}}=\int_a^b\left(\frac{\partial g_{in}}{\partial x}(t,\z)\right)^{\bm{\epsilon}}f^{2d}(t|\z)\,\d t,\quad \bm{\epsilon}\in\{\bm{\pm} \}.\]
\end{Theorem}
\begin{proof}
See Appendix \ref{other}.
	\end{proof}
	Now, we generalize our definition to include the random variables $X$ whose supports are not necessarily bounded. To do so, 
	assume that $\Supp(X)\subseteq [a,\infty)$. Then, we define 
	\[\PIEV_d^{\mathbf{z}}(X\to Y)^{\bm{+}}:=\lim_{b\to \infty} \lim_{\substack{P\in\mathcal{P}([a,b]) \\\lVert P\rVert\to 0}}L_{P,d}^{\mathbf{z}}(X|_{[a,b]}\to Y)^{\bm{+}}.\]
	Similarly, if $\Supp(X)\subseteq (-\infty,b]$, we define 
	\[\PIEV_d^{\mathbf{z}}(X\to Y)^{\bm{+}}:=\lim_{a\to -\infty} \lim_{\substack{P\in\mathcal{P}([a,b]) \\\lVert P\rVert\to 0}}L_{P,d}^{\mathbf{z}}(X|_{[a,b]}\to Y)^{\bm{+}}.\]
	Finally, if $\Supp(X)\subseteq (-\infty,\infty)$, we define $\PIEV(X\to Y)^{\bm{+}}$ as follows:
	\begin{align*}
	&\lim_{a\to - \infty} \lim_{\substack{P\in\mathcal{P}([a,0]) \\\lVert P\rVert\to 0}}L_{P,d}^{\mathbf{z}}(X|_{[a,0]}\to Y)^{\bm{+}}
	+\lim_{b\to \infty} \lim_{\substack{P\in\mathcal{P}([0,b]) \\\lVert P\rVert\to 0}}L_{P,d}^{\mathbf{z}}(X|_{[0,b]}\to Y)^{\bm{+}}.
	\end{align*}
	\begin{Corollary}\label{corNPEV}
	Theorem \ref{NPEV} holds also for any random variable $X$ which does not necessarily satisfy $\Supp(X)\subseteq[a,b]$.
	\end{Corollary}
	\begin{proof}
	It is straightforward.
\end{proof}
\begin{Corollary}\label{cor2NPEV}
	Under the assumptions of Corollary \ref{corNPEV}, we have that
	\[\PIEV_d^{\z}(X\to Y)=\PIEV_d^{\z}(X\to Y)^{\bm{+}}+\PIEV_d^{\z}(X\to Y)^{\bm{-}}.\]
\end{Corollary}
\begin{Remark}
Let $Y=g(X,\Z)$ with $\Supp(X)\subseteq (a,b)$. 	Another alternative definition for $	\PIEV^{\z}(X\to Y)^{\bm{+}}$ (resp. 	$\PIEV^{\z}(X\to Y)^{\bm{-}}$) is as follows:
	\[\sup\left\{\int_{\Gamma}g_{in}\phi'\,\d x: \phi\in C_c^1(\Gamma,\mathbb{R}),|\phi|\le f^{2d}(\,\cdot\,|\z)\right\},\]
	where $\Gamma$ is an open subset of $(a,b)$ consisting of all points $x\in(a,b)$ for which there exist $r_x$ in such a way that the restriction of  $g_{in}(\,\cdot\,,\z)$ to $(x-r_x,x+r_x)$ is increasing (resp. decreasing). In this case, one could see that if $g(\,\cdot\,,\z)$ is continuously differentiable, then Theorem ~\ref{NPEV} holds. In the situations that $\Supp(X)$ is unbounded,  Corollary \ref{corNPEV} holds. Furthermore, in general, Corollary \ref{cor2NPEV} holds as well. 
\end{Remark}
Now, naturally such as before, by normalizing and then taking the expected value with respect to $\Z$, we have the $\PEACE_d(X\to Y)^{\bm{\epsilon}}$  formula for $\bm{\epsilon}\in\{\bm{\pm}  \}$. One could see that
\[\PEACE_d(X\to Y)=\PEACE_d(X\to Y)^{\bm{+}}+\PEACE_d(X\to Y)^{\bm{-}}.\] 
\section{Investigating Some Examples}\label{Investigating Some Examples}
In this section, we investigate some causal problems to show the general capability of our framework. 
\subsection*{PEACEs Corresponding to Independent Uniform Discrete Random Variables}
Let $Y=\X\bm{\alpha}+\Z\bm{\beta}$ in such a way that $\X$ given $\Z$  has a uniform distribution. Assume that $\Supp(X_i)=\{1,2,\ldots,n_i\}$ for any $i$. Then, by using the notations of Proposition \ref{peace-discerete}, we have that $\omega_{\X|\Z,d}\equiv 1/(n_1\cdots n_m)^{2d}$, and $\mathrm{dif}^{(i)}\equiv \alpha_i$. It follows that $\mathrm{PFlux}^{(i)}\equiv \alpha_i/(n_1\cdots n_m)^{2d}$, and hence $\mathrm{PFlux}\equiv |\bm{\alpha}|/(n_1\cdots n_m)^{2d}$. Therefore, 
\[\PIEV_d(\X\to Y)=\frac{(n_1-1)\cdots(n_m-1)|\bm{\alpha}|}{(n_1\cdots n_m)^{2d}}.\]
Hence,
\[\PEACE_d(\X\to Y)=\frac{4^d(n_1-1)\cdots(n_m-1)|\bm{\alpha}|}{(n_1\cdots n_m)^{2d}}.\]
\subsection*{Newton's Second Law}
Assume that a force $f$ is applied to an object of mass $m$ as shown in Figure \ref{bodymassm}. Consequently, a friction force $f_0$ resists the movement of the object. Assume that $f$ overcomes $f_0$, and the object starts moving.  Then, by Newton's second law of motion, we have that $f-f_0=ma$, where $a$ is the acceleration of the object.  Consider a specific moment when the object is moving, and the friction force $F_0$ due to environmental factors has a random nature and acts against a random force  $F>>F_0$. Hence, in this specific moment, we can assume that  $F$ and $F_0$ are independent, and $F\sim \mathrm{N}(\mu,\sigma^2)$ and $F_0\sim\mathrm{N}(\mu_0,\sigma_0^2)$. It follows that $a$ has a random nature as well. Let $A$ be the random variable describing the value of $a$. Then, we have that $A=g(F,F_0)=(F-F_0)/m$. Therefore, for $d>0$, we have that 
\begin{align*}
	\PIEV_d^{F_0=\beta}(F\to A)&\approx \int_{-\infty}^{\infty}\left|\frac{\rond g}{\rond F}(\alpha,\beta)\right|f_{F}^{2d}(\alpha)\,\d \alpha=\frac{1}{m}\int_{-\infty}^{\infty}f_{F}^{2d}(\alpha|\beta)\,\d \alpha\\
	&=\frac{1}{m(2\pi\sigma^2)^d}\int_{-\infty}^{\infty}e^{-d\left(\frac{\alpha-\mu}{\sigma}\right)^{2}}\,\d\alpha=\frac{1}{m\sqrt{2d}(2\pi)^d\sigma^{2d-1}}\int_{-\infty}^{\infty}e^{-\frac{\alpha^2}{2}}\,\d\alpha\\
	&=\frac{\sqrt{2\pi}}{m\sqrt{2d}(2\pi)^d\sigma^{2d-1}}=\frac{1}{m\sqrt{d}2^d\pi^{d-\frac{1}{2}}\sigma^{2d-1}}.
\end{align*}
Hence, 
\begin{align*}
	\PEACE_d(F\to A)\approx \E_{F_0}\biggl(\PIEV_d^{F_0=\beta}(F\to A)\biggr)=\frac{1}{m\sqrt{d}2^d\pi^{d-\frac{1}{2}}\sigma^{2d-1}}.
\end{align*}
Similarly, we have that
\begin{align*}
	\PEACE_d(F_0\to A)\approx \E_{F}\biggl(\PIEV_d^{F=\beta}(F_0\to A)\biggr)=\frac{1}{m\sqrt{d}2^d\pi^{d-\frac{1}{2}}\sigma_0^{2d-1}}.
\end{align*}
Further, 
\begin{align*}
	\PEACE_d((F,F_0)\to A)&=\int_{-\infty}^{\infty}\int_{-\infty}^{\infty}\left|\nabla g(\alpha,\beta)\right|f_{(F,F_0)}^{2d}(\alpha,\beta)\,\d \alpha\d \beta\\
	&=\frac{\sqrt{2}}{m}\left(\int_{-\infty}^{\infty}f_{F}^{2d}(\alpha)\,\d \alpha\right)\left(\int_{-\infty}^{\infty}f_{F_0}^{2d}(\beta)\,\d \beta\right)\\
	&=\frac{\sqrt{2}}{m}\left( \frac{1}{\sqrt{d}2^d\pi^{d-\frac{1}{2}}\sigma^{2d-1}}\right)\left( \frac{1}{\sqrt{d}2^d\pi^{d-\frac{1}{2}}\sigma_0^{2d-1}}\right)\\
	&=\frac{1}{m2^{2d-\frac{1}{2}}d\pi^{2d-1}(\sigma\sigma_0)^{2d-1}}.
\end{align*}
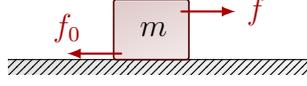
\begin{figure}
	\begin{center}
		\begin{tikzpicture}
			\def\W{4}  
			\def\D{0.2}  
			\def\h{0.8}  
			\def\w{1.0}  
			\def\mx{-0.15*\W} 
			
			\draw[ground] (-0.5*\W,0) rectangle++ (\W,-\D);
			\draw (-0.5*\W,0) --++ (\W,0);
			\draw[mass] (\mx,0) rectangle++ (\w,\h) ;
			\node at (-.07,0.4){$m$};

			\ifthenelse{\boolean{showforces}}{
				\draw[force] (\mx+0.9*\w,0.80*\h) --++ ( 0.9*\h,0) node[right] {$f$};
				\draw[force] (\mx+0.1*\w,0.10*\h) --++ (-0.9*\h,0) node[right=6,above=0] {$f_0$};
			}{}
			
		\end{tikzpicture}
		\caption{An object of mass $m$ that is affected by a force $f$ and consequently the friction force $f_0$.}\label{bodymassm}
	\end{center}
\end{figure}

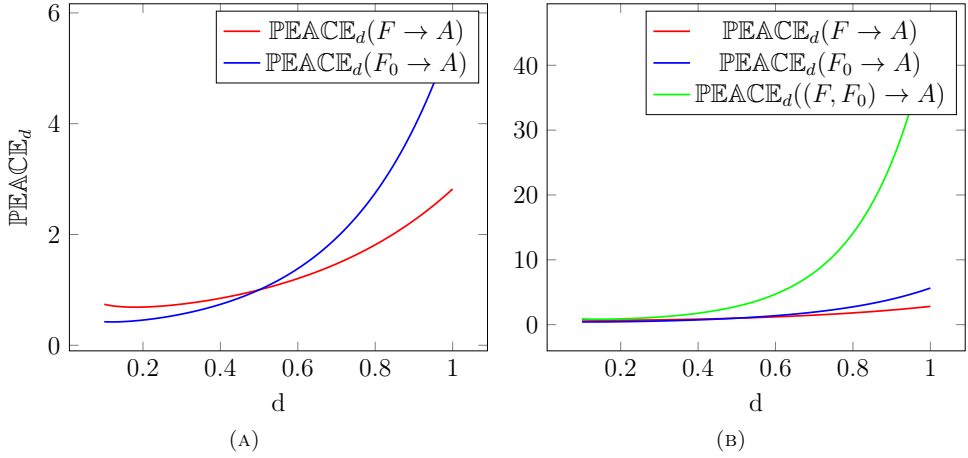
\begin{figure}
	\resizebox{0.87\textwidth}{!}{
		\subfloat[]
		{
			\begin{tikzpicture}
				\begin{axis}[ xlabel=d, ylabel = $\PEACE_d$]
					\addplot [no markers,domain=0.1:1, samples=100,color=red, thick]{(2^(x-1))*(5^(2*x-1))/(sqrt(x)*(pi^(x-0.5)))};
					\addlegendentry{$\PEACE_d(F\rightarrow A)$}
					\addplot [no markers,domain=0.1:1, samples=100,color=blue,thick]{(2^(x-1))*(10^(2*x-1)/(sqrt(x)*(pi^(x-0.5)))};
					\addlegendentry{$\PEACE_d(F_0\rightarrow A)$}
				\end{axis}
			\end{tikzpicture}
		}
		\subfloat[]{
			\begin{tikzpicture}
				\begin{axis}[ xlabel=d]
					\addplot [no markers,domain=0.1:1, samples=100,color=red, thick]{(2^(x-1))*(5^(2*x-1))/(sqrt(x)*(pi^(x-0.5)))};
					\addlegendentry{$\PEACE_d(F\rightarrow A)$}
					\addplot [no markers,domain=0.1:1, samples=100,color=blue,thick]{(2^(x-1))*(10^(2*x-1)/(sqrt(x)*(pi^(x-0.5)))};
					\addlegendentry{$\PEACE_d(F_0\rightarrow A)$}
					\addplot [no markers,domain=0.1:1, samples=100,color=green,thick]{sqrt(2)*(100^(2*x-1))/(x*(pi^(2*x-1)))};
					\addlegendentry{$\PEACE_d((F,F_0)\rightarrow A)$}
				\end{axis}
		\end{tikzpicture}}
	}
	\caption{The PEACEs of degree $d$ of $F$, $F_0$, and $(F,F_0)$ on $A$ for $\sigma=0.1$, $\sigma_0=0.05$, and $m=1\mathrm{kg}$.}
\end{figure}

\subsection*{PEACEs Corresponding the joint of Random Variables vs PEACEs Corresponding to Each of the Variables  }
Let $\X$ and $\Z$ be continuous random vectors,  $y=g(\X,\Z)$, and $\X=(X_1,\ldots,X_m)$. Assume that $g$ has continuous partial derivatives with respect to variables in $\X$. We can generalize the definition of PEACE in the following sense:
\[\PEACE_d^{(r)}(\X\to Y):=\E_{\Z^{(r)}}(\X\to Y):=\int\PIEV_d^{\z}(\X\to Y)f_{\Z}^r(\z)\,\d\x.\]
Hence, for greater $r$, the above mean is more dependent on the greater density values of $\Z$ rather than the smaller ones. Now, first, we get a lower and an upper bound for $\PEACE_d^{(2d)}(\X\to Y)$ in terms of the values $\PEACE_d^{(2d)}(X_i\to Y)$.
To do so, by setting $\X^{i}=(X_1,\ldots, X_{i-1},X_{i+1},\ldots,X_m)$ and $\W^i=(\X^i,\Z)$, we have that 
\begin{align*}
	\PEACE_d^{(2d)}(\X\to Y)&=\iint\left|\frac{\partial g}{\partial \x}(\x,\z)\right|f(\x|\z)^{2d}f(\z)^{2d}\,\d\x \d\z\\
	&\le \iint\sum_{i=1}^m \left|\frac{\partial g}{\partial x_i}(x_i,\w^i)\right|f(\x|\z)^{2d}f(\z)^{2d}\,\d\x \d\z\\
	&= \sum_{i=1}^m \iint \left|\frac{\partial g}{\partial x_i}(x_i,\w^i)\right|f(x_i|\x^i,\z)^{2d}f(\x^i|\z)^{2d}f(\z)^{2d}\,\d x_i  \d\w^i\\
	&= \sum_{i=1}^m \iint \left|\frac{\partial g}{\partial x_i}(x_i,\w^i)\right|f(x_i|\w^i)^{2d}f(\w^i)^{2d}\,\d x_i  \d\w^i\\
	&=\sum_{i=1}^m\PEACE_d^{(2d)}(X_i\to Y).
\end{align*}
Similarly, by using the well-known fact that the quadratic mean of some non-negative numbers is not less than the arithmetic means of them, we have that
\begin{align*}
	\PEACE_d^{(2d)}(\X\to Y)\ge 
	&\frac{1}{\sqrt{n}}\sum_{i=1}^m\PEACE_d^{(2d)}(X_i\to Y).
\end{align*}
Now, consider the case that $g$ is linear and for any $i$, $X_i$ and $\X^i$ are independent given $\Z$. Then, if $Y=\X\bm{\alpha}+\Z\bm{\beta}$, where $\bm{\alpha}$ and $\bm{\beta}$ are column vectors, we have that 
\begin{align*}
	\PEACE_d(\X\to Y)&= \iint |\bm{\alpha}|f(\x|\z)^{2d}f(\z)\,\d\x \d\z= \prod_{i=1}^m \iint |\bm{\alpha}|f(x_i|\z)^{2d}f(\z)\,\d x_i \d\w^i\\
	&=\prod_{i=1}^m \iint |\bm{\alpha}|f(x_i|\w^i)^{2d}f(\w^i)\,\d x_i \d\w^i=\frac{|\bm{\alpha}|\prod_{i=1}^m \PEACE_d(X_i\to Y)}{|\alpha_1\cdots\alpha_n|}.
\end{align*}

\subsection*{Effect of Sodium Intake and Age on the Blood Peasure}
\begin{figure}
	\begin{tikzpicture}
		\node[state] (S) at (0,0) {$S$};
		
		\node[state] (B) at (3,0) {$B$};
		\node[state] (P) at (1.5,-1.5) {$P$};
		\node[state] (A) at (1.5,1.5) {$A$};
		\path (S) edge (B);
		\path (B) edge (P);
		\path (S) edge (P);
		\path (A) edge (B);
		\path (A) edge (S);
	\end{tikzpicture}
	\caption{The causal graph associated to the sodium intake ($S$), age ($A$), the proteinuria ($P$), and the blood pressure ($B$).}\label{blood}
\end{figure}
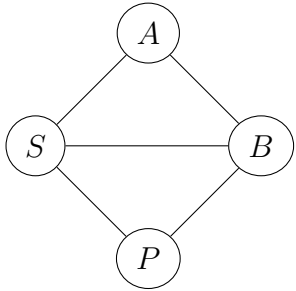

This example has been discussed in \cite{schomaker2018educational} using the Pearl graphical framework.  Let $S$, $A$, $P$, and $B$ denote the sodium intake, the age, the proteinuria, and the blood pressure, respectively. Then, there is a  causal graph as shown in Figure \ref{blood}. For this example, we use the data and the python code provided in \href{https://github.com/joseffaghihi/PEACE-General-.git}{PEACE-General}). to obtain the plots shown in Figure~\ref{sodium}.
We note that Subfigures (\ref{a}, \ref{b}) shows that when we increase the degree $d$ from 0 to 1, $\PEACE_d(A\to B)$ loses its strength against $\PEACE_d(S\to B)$ and becomes less than $\PEACE_d(S\to B)$. This means that the ``\textit{average causal change of $B$ with respect to $A$ for more probable subpopulations determined by $S$}'' becomes less than the ``\textit{average causal change of $B$ with respect to $S$ in more probable subpopulations determined by $A$}''. However,  by Subfigures (\ref{c}, \ref{d}), when we increase the degree $d$ from 0 to 1, $\PEACE_d^{(2d)}(A\to B)$ remains superior against $\PEACE_d^{(2d)}(S\to B)$. Indeed, in $\PEACE_d^{(2d)}(*\to B)$, compared to $\PEACE_d(*\to B)$, the density values $f(\#)$ appeared in the final average,  have the same power ($2d$) as the density values $f(*|\#)$ for $*,\#\in \{S,A\}$ with $*\neq \#$. As we see in Figures (\ref{e},\ref{f}), by increasing $d$, $\PEACE_d((S,A)\to B)$ becomes less than both $\PEACE_d(S\to B)$ and $\PEACE_d(A\to B)$, while  Figures~(\ref{g}, \ref{h}) say that $\PEACE_d^{(2d)}((S,A)\to B)$ remains superior to both $\PEACE_d^{(2d)}(S\to B)$ and $\PEACE_d^{(2d)}(A\to B)$. 
\definecolor{ao}{rgb}{0.0, 0.5, 0.0}
\begin{figure}
	\vspace*{-0.5cm}
	\subfloat[\label{a}]
	{\includegraphics[scale=0.4]{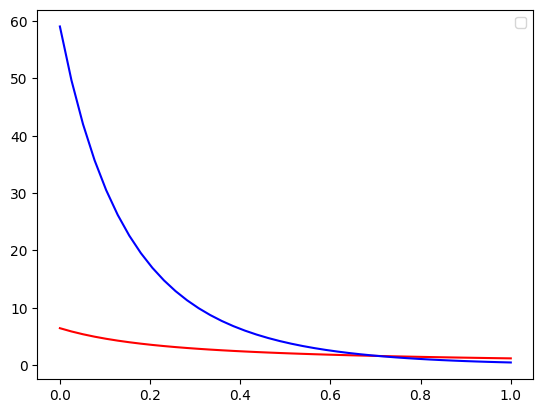}}
		\subfloat[\label{b}]
	{\includegraphics[scale=0.4]{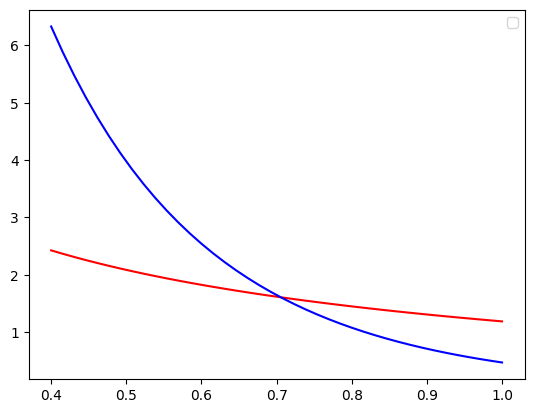}}\\
		\subfloat[\label{c}]
	{\includegraphics[scale=0.4]{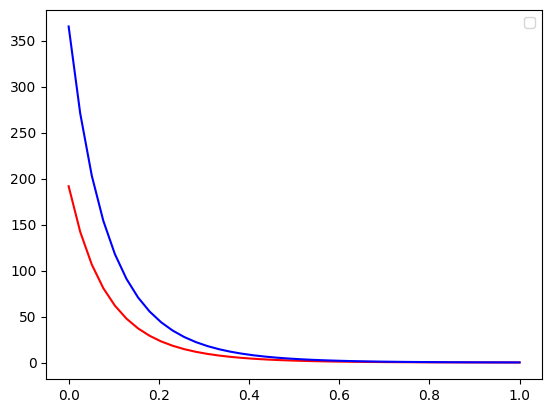}}
	\subfloat[\label{d}]
	{\includegraphics[scale=0.4]{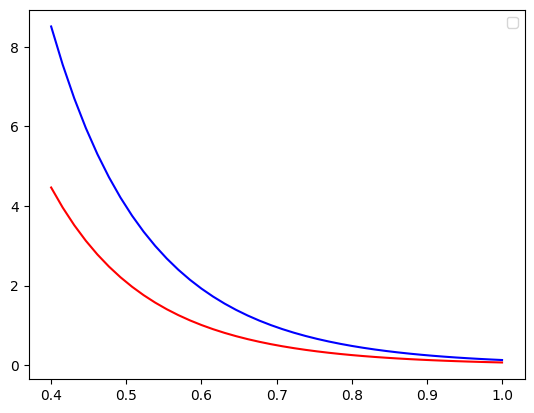}}\\
		\subfloat[\label{e}]
	{\includegraphics[scale=0.4]{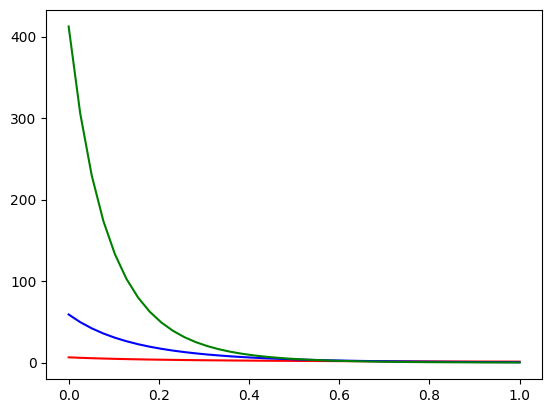}}
	\subfloat[\label{f}]
	{\includegraphics[scale=0.4]{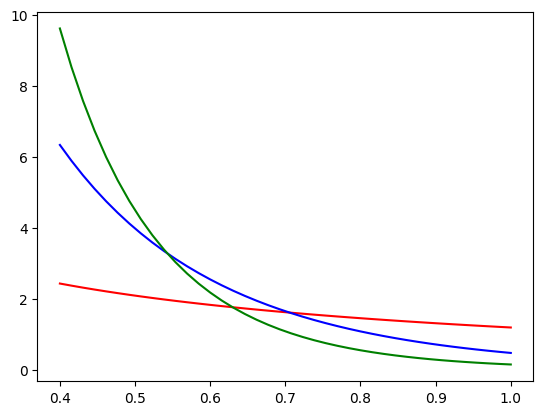}}\\
		\subfloat[\label{g}]
	{\includegraphics[scale=0.4]{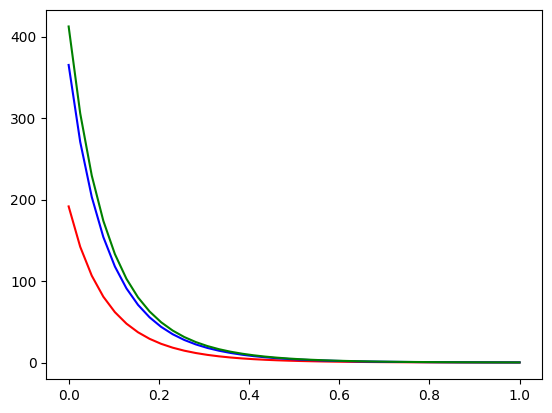}}
	\subfloat[\label{h}]
	{\includegraphics[scale=0.4]{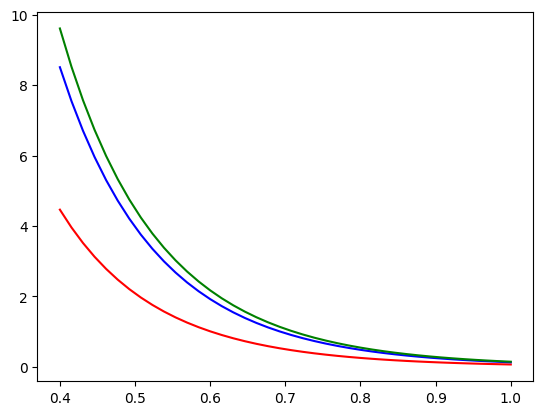}}
	\vspace*{-.3cm}
	\caption{\small{The PEACEs of \textcolor{red}{sodium intake},  \textcolor{blue}{age}, and the \textcolor{ao}{joint of sodium intake and age} on blood pressure. The horizontal axis denotes the degree $d$. Also,  the vertical axis of Plots (A), (B), (E), and (F) denotes the $\PEACE_d$, while in the other plots, it denotes $\PEACE_d^{(2d)} $. }}\label{sodium}
\end{figure}
\section{Conclusion}\label{Conclusion}
In this paper, we extended and developed a framework, introduced in \cite{faghihi2022probabilistic}, for measuring the direct causal effect of a random vector on an outcome variable in both discrete and continuous cases. To formalize our framework, we used several concepts and tools from functional analysis and measure theory such as integrals on open sets, total variation of a multivariate function, and the flux of a function passing through a surface. Our framework has a general capability to deal with different types of causal problems compared to the other well-known  frameworks such as the Rubin-Neyman, the Pearl, and the Janzing et al. frameworks. Indeed, we introduced and justified a function called Probabilistic Easy Variational Causal Effect (PEACE), which measures the total direct causal changes of an  outcome $Y$ with respect to interventionally and continuously changing the value of the exposure/treatment $\X$, while keeping other variables $\Z$ unchanged. PEACE is a function of a degree $d\ge 0$.  By considering small and high values for $d$, one could have  direct causal effect values for which  the probability/density of $\X$ given $\Z$ is strengthened and weakened, respectively. Hence,  in an observational study, when  rare subpopulations determined by $\Z$ do not have  important impacts on $Y$ (e.g., the problem of the effect of a rare noise on the quality of an image), then higher values of $d$ are suitable. Otherwise, smaller values of $d$ could be used (e.g., the problem of the effect of a rare disease on blood pressure). PEACE reflects the total absolute direct causal changes of $Y$ with respect to $\X$ regardless of being positive or negative. Thus, to measure separately the positive and the negative direct causal changes of $Y$, we introduced the positive and negative PEACEs. Further, we showed that the PEACE of $\X$ on $Y$ is stable under small changes of the joint density of $\X$ and $\Z$, and the partial derivative of $Y_{in}$ with respect to $\X$, where $Y_{in}$ is obtained from $Y$ by removing all functional relationships defining $\X$ and $\Z$. Furthermore, in the presence of unobserved variables, we provided an identifiability criterion. Moreover, we supported the general capability of PEACE by investigating some examples. 

\bibliographystyle{amsplain}
\bibliography{References}
\appendix
\section{Some Results on Admissible Compact Covers }\label{Some Results on Admissible Compact Covers }
	In the following lemma, we show that each open subset of $\mathbb{R}^n$ has an admissible compact cover. 
\begin{Lemma}\label{existenceofthecomapctsequence}
	Let $V$ be an open subset of $\mathbb{R}^n$. Then, there exists a sequence $\{C_i\}_{i=0}^{\infty}$ of rectifiable compact  subspaces of $V$ such that $C_i\subseteq \mathrm{Int}(C_{i+1})$,  the boundary of $C_i$ is piece-wise $C^1$ for any $i\ge 0$, and $V=\bigcup_{i=0}^{\infty}C_i$.
\end{Lemma}

\begin{proof}
Let $\{a_i\}_{i=0}^{\infty}$ be a strictly decreasing sequence of positive real numbers converging to 0. For any non-negative integer $i$, assume that
	\[B_i=\{\x\in V: \inf_{\x'\in \mathbb{R}^n\backslash V}|\x-\x'|\ge a_i,\;|\x|\le \frac{1}{a_i}\}.\]
Then, for any $i$, $B_i$ is a compact subsapce of $\mathbb{R}^n$,  $B_i\subseteq \mathrm{Int}(B_{i+1})$, and $V=\bigcup_{i=0}^{\infty}B_i$. 	However, the sets $B_i$ might  not be  rectifiable, and their boundaries might not be piece-wise $C^1$. To solve this issue, for any $i$ and $\x\in B_i$, assume that $V_{i,\x}\subseteq \mathrm{Int}(B_{i+1})$ is a bounded closed $n$-cell including $\x$. In the rest of this proof, by ``$i$'' we mean any $i$. Now, assume that $U_{i,\x}\subseteq V_{i,\x}$ is an open subset of $\mathbb{R}^n$ including $x$. Then, $\{U_{i,\x}:\x\in B_i\}$ is an open cover for $B_i$, which has a finite subcover $\{U_{i,\x_0},\ldots, U_{i,\x_{n_i}}\}$ for $B_i$ by the compactness of $B_i$.  Set $C_i=\bigcup_{j=0}^{n_i}V_{i,\x_j}$. Then, $C_i$ is compact as a finite union of compact subsets of $\mathbb{R}^n$. Further, clearly $B_i\subseteq C_i\subseteq \mathrm{Int}(B_{i+1})\subseteq \mathrm{Int}(C_{i+1})$, and $\bigcup_{i=0}^{\infty}B_i\subseteq\bigcup_{i=0}^{\infty} C_i$, and hence $\bigcup_{i=0}^{\infty} C_i=V$. Since each $n$-cell is rectifiable, $C_i$ is also rectifiable as a finite union of $n$-celles. Finally, the boundary of $C_i$ is piece-wise linear, and consequently, it is a piece-wise manifold of class $C^1$. 
\end{proof}
\begin{Remark}\label{remarkforexistenceofthecomapctsequence}
	In the proof of Lemma \ref{existenceofthecomapctsequence}, we could define
		\[B_i=\{\x\in V: h(\x)\ge a_i,\;|\x|\le \frac{1}{a_i}\},\]
		where $h:V\to\mathbb{R}$ is a positive continuous function.
		The rest of the proof is the same as before, and suitable $C_i$'s could be found. Thus, we have that 
		\[C_i\subseteq \mathrm{Int}(B_{i+1})=\left\{\x\in V: h(\x)>a_{i+1},\;|\x|< \frac{1}{a_{i+1}}\right\}.\]
\end{Remark}
The following result  is directly used in the proof of Theorem \ref{maintheorem}.
The following proposition is required to prove Theorem \ref{maintheorem}.
\begin{Proposition}\label{supp(phi)bd(c_i)}
	Let $V$ be an open subset of $\mathbb{R}^n$ and $\{C_i\}_{i=0}^{\infty}$ be an admissible compact cover for $V$. Also, let $C\subseteq V$ be compact. Then, there exists a positive integer $N_0$ such that $i\ge N_0$ implies that $C\cap\bd(C_i)=\emptyset$. 
\end{Proposition}
\begin{proof}
	It is enough to show that there exists $N_0$ with $C\subseteq C_{N_0-1}$, since it follows from $C_{N_0-1}\subseteq \inte(C_{i})$ that $C\cap \bd(C_i)=\emptyset$ for any $i\ge N_0$. On the contrary, assume that  $C\nsubseteq C_i$ for any $i$. Choose $\x_i\in C\backslash C_i$ for any $i$. It follows from the compactness of $C$ that $\{\x_i\}_{i=0}^{\infty}$ has a convergent subsequence $\{\x_{i_j}\}_{j=0}^{\infty}$. Assume that $\x=\lim_{j\to\infty}\x_{i_j}$. There exists an index $l$ with $\x\in C_{l}$, which implies that $\x\in\inte(C_{l+1})$. Let $i_{j_0}\ge l+1$. Then, $\x\in \inte(C_{i_{j_0}})$, and hence there exists a positive integer $N\ge j_0$ such that $j\ge N$ implies that $\x_{i_j}\in C_{i_{j_0}}$. Now, it follows from $C_{i_{j_0}}\subseteq C_{i_N}$ that $\x_{i_N}\in C_{i_N}$, a contradiction!
\end{proof}

\section{Proof of Theorem \ref{maintheorem} and Its Corollaries}\label{appendixmaintheorem}

To prove Theorem \ref{maintheorem}, roughly speaking, the idea is to build a sequence of functions belonging to $C_c^1(\Omega,\mathbb{R}^n)$ such that their integrals on $\Omega$ tend to the integral of $(|\nabla g|/\nabla(g))f_{\X}^{2d}$ on the set of points with $\nabla g\neq 0$.
To do so, first, we need some lemmas and their corollaries. We start with the following lemma that is used to prove Lemma 
\ref{stack}.
\begin{Lemma}\label{ReLU}
	Let $f(x)=\max\{x,0\}$ for any $x\in\mathbb{R}$ (this function is called the \textit{positive part } or \textit{ReLU} function). Then, for any $\epsilon>0$, there exists a non-negative function $\psi\in C^1(\mathbb{R},\mathbb{R})$ with  $|\psi(x)-f(x)|<\epsilon$ for any $x\in\mathbb{R}$. 
\end{Lemma}
\begin{proof}
Equivalently, we show that for any $\epsilon>0$, there exists a non-negative function $\psi\in C^1(\mathbb{R},\mathbb{R})$ with $\psi(x)=0$ when $x\le 0$, and  $|\psi(x)-x|<\epsilon$ when $x\ge 0$. 	Let $\epsilon>0$. Define
	\[\psi(x)=\left\{\begin{array}{ll}0,&x\le 0\\[2pt] \frac{2x^2}{\epsilon},& 0\le x\le \frac{\epsilon}{4}\\ [2pt]x-\frac{\epsilon}{8},& x\ge \frac{\epsilon}{4}\end{array}\right..\]
	One could see that $\psi\in C^1(\mathbb{R},\mathbb{R})$.  For $0\le x\le \epsilon/4$, we have that
	\[|\psi(x)-x|<\epsilon \iff \frac{2x^2}{\epsilon}-x-\epsilon<0\;\;\;\&\;\;\;\frac{2x^2}{\epsilon}-x+\epsilon>0. \]
	The first and the second inequalities are true for $-\epsilon/2<x<\epsilon$ and  any $x\in\mathbb{R}$, respectively. Hence, both  inequalities are true for $0\le x\le \epsilon/4$. Clearly, for $x\ge \epsilon/4$, we have that $|\psi(x)-x|=\epsilon/8<\epsilon$. 
\end{proof}
The following lemma is  required to prove Corrolary \ref{corr}.
\begin{Lemma}\label{keylemma}
	Let $U$ and $V$ with $\emptyset\neq U\subseteq V$ be  open subsets of  $\mathbb{R}^n$, and let $\x_0\in U$. Then, there exists $q\in C^1_c(V,\mathbb{R})$ with $\Supp(q)\subseteq U$ and $q(\x_0)\neq 0$.
\end{Lemma}
\begin{proof}
	Define 
	\[q_0:\mathbb{R}\to\mathbb{R},\qquad q_0(x)=\left\{\begin{array}{ll}x^2,&x> 0\\ 0,&x\le 0\end{array}\right..\]
	One could see that $q_0\in C^1(\mathbb{R},\mathbb{R})$. Now, for any closed interval $[a,b]$ with $a< b$, we define $q_{a,b}(x)=q_0(x-a)q_0(b-x)$. Then, $q_{a,b}\in C^1_c(\mathbb{R},\mathbb{R})$ with $\Supp(q_{a,b})\subseteq [a,b]$. Now, for an $n$-cell $C=[a_1,b_1]\times\cdots\times [a_n,b_n]$, we define $q_C:\mathbb{R}^n\to\mathbb{R}$ by setting $q_C(\x)=q_{a_1,b_1}(x_1)\cdots q_{a_n,b_n}(x_n)$. Obviously, $q_C\in C^1_c(\mathbb{R}^n,\mathbb{R})$ with $\Supp(q_C)\subseteq C$. Now, let $C_0$ be  an $n$-cell  with $\x_0\in C_0\backslash \mathrm{Bd}(C_0)\subseteq C_0\subseteq U$. Then, the desired function $q$ could be selected as the restriction of  $(q)_{C_0}$ on $V$. 
\end{proof}
The following lemma is known as  Urysohn's lemma for locally compact Hausdorff topological spaces (see \cite[Lemma 2.12]{rudin1974real}). 
\begin{Lemma}\label{uryhsonlemma}
	Let $X$ be a Hausdorff topological space and $K\subseteq U\subseteq X$, where $K$ is compact and $U$ is an open subset of $X$. Then, there exists $q\in C_c(X,\mathbb{R})$ in such a way that $0\le q\le 1$, $q|_{K}\equiv 1$ and $\Supp(q)\subseteq U$. 
\end{Lemma}
The following corollary is directly used in the proof of Theorem \ref{maintheorem}.
\begin{Corollary}\label{need}
	Let $X$ be a Hausdorff topological space and $K\subseteq U\subseteq X$, where $K$ is compact and $U$ is an open subset of $X$. Also, let $f:X\to\mathbb{R}$ be a  continuous function. Then, there exists $\widetilde{f}\in C_c(X,\mathbb{R})$ with $\widetilde{f}|_K=f$, $|\widetilde{f}|\le |f|$, and $\Supp(\widetilde{f})\subseteq U$. Especially, if $f$ is non-negative, then we have the additional property  $0\le \widetilde{f}\le f$.
\end{Corollary}
\begin{proof}
	It follows from Lemma \ref{uryhsonlemma} that there exists $q\in C_c(X,\mathbb{R})$ with  $0\le q\le 1$, $q|_{K}\equiv 1$ and $\Supp(q)\subseteq U$. Now, it is enough to assume that $\widetilde{f}=qf$. 
\end{proof}
Now, we provide some definitions and concepts that are used in Theorem \ref{Stone–Weierstrass}.  Let $X$ be a topological space. We say that a continuous function $f:X\to \mathbb{R}$ \textit{vanishes at infinity} if for any $\epsilon >0$, there exists a compact subspace $K$ of $X$ with $|f|<\epsilon$ on $X\backslash K$.  We denote the set of all continuous functions $f:X\to\mathbb{R}$ that vanish at infinity by $C_0(X,\mathbb{R})$.  Note that $C_0(X,\mathbb{R})$ is an algebra. In other words,  it is non-empty, and for any $f,g\in C_0(X,\mathbb{R})$ and $\alpha\in\mathbb{R}$, we have that $f+\alpha g\in C_0(X,\mathbb{R})$ and $fg\in C_0(X,\mathbb{R})$. Further,  $C_0(X,\mathbb{R})$ is equipped with the $L^{\infty}$-norm (i.e., $\lVert f\rVert_{\infty}=\sup_{x\in X}|f(x)|$).
Let $A$ be a subalgebra of $C_0(X,\mathbb{R})$. We say that $A$ \textit{separates the points} of $X$ if for any $x_1,x_2\in X$ with $x_1\neq x_2$, there exists $f\in A$ with $f(x_1)\neq f(x_2)$. Further, we say that $A$ \textit{vanishes nowhere} if for any $x\in X$ there exists $f\in A$ with $f(x)\neq 0$.  The following  is known as  the Stone–Weierstrass theorem for locally compact spaces (see \cite{de1959stone}).
\begin{Theorem}\label{Stone–Weierstrass}
	Let $X$ be a locally compact Hausdorff topological space and $A$ be a subalgebra of $C_0(X,\mathbb{R})$. Then, $A$ is dense in $C_0(X,\mathbb{R})$ if and only if $A$ separate the points of $X$ and it vanishes nowhere. 
\end{Theorem}
We use the following corollary in Lemma \ref{stack} and also directly in the proof of Theorem \ref{maintheorem}.
\begin{Corollary}\label{corr}
	Let $U$ be an open subset of $\mathbb{R}^n$. Then, $C^1_c(U,\mathbb{R})$ is dense in $C_0(U,\mathbb{R})$. 
\end{Corollary}
\begin{proof}
	Each open subset of $\mathbb{R}^n$ is  a locally compact Hausdorff topological space. Let $x_1,x_2\in U$ with $x_1\neq x_2$. Since $U$ is an open subset of $\mathbb{R}^n$, there exists $r>0$ with $B(x_1,r)=\{y\in \mathbb{R}^n:|x_1-y|<r\}\subseteq U$. Let $r'=\min\{r, |x_1-x_2|\}$. Then, by Lemma~\ref{keylemma}, there exists $q\in C^1_c(U,\mathbb{R})$ with $\Supp(q)\subseteq B(x_1,r'/2)$ and $q(x_1)\neq 0$. Also, since $x_2\notin B(x_1,r'/2)$, we have that $q(x_2)=0$. 
	Thus,  $C_c^1(U,\mathbb{R})$ vanishes nowhere, and it separates the points of $U$. Therefore, it follows from Theorem \ref{Stone–Weierstrass} that $C_c^1(U,\mathbb{R})$ is dense in $C_0(U,\mathbb{R})$.
\end{proof}
In the proof of Theorem \ref{maintheorem}, we use the following lemma to approximate $f_{\X}^{2d}$ with a compactly supported continuously differentiable function (see \href{https://math.stackexchange.com/questions/4670844}{Approximations of continuous functions by compactly supported smooth functions with a criterion}).
\begin{Lemma}\label{stack}
	Let $f:U\to \mathbb{R}$ be a non-negative continuous function which vanishes at infinity. Then, for any $\epsilon >0$, there exists  a non-negative function $g\in C_c^1(U,\mathbb{R})$ with $0\le f-g<\epsilon$.
\end{Lemma}
\begin{proof}
Let $\epsilon >0$. 	It follows from Corollary \ref{corr} that there exists $g_1\in C_c^1(U,\mathbb{R})$ with $|f-g_1|<\epsilon/4$. Assume that $g_2=g_1-\epsilon/2$. Then, we have that $\epsilon/4< f-g_2<3\epsilon/4$. It follows from Lemma \ref{ReLU} that there exists a non-negative function $\psi\in C_c^1(\mathbb{R},\mathbb{R})$ in such a way that $\psi(x)=0$ for $x\le 0$, and $|\psi(x)-x|<\epsilon/4$ for $x>0$. Set $g=\psi\circ g_2$. Then, $g\ge 0$. Now, we show that $0\le f-g<\epsilon$. Let $x\in U$. First, assume that $g_2(x)> 0$. Then, it follows from $|\psi(g_2(x))- g_2(x)|<\epsilon/4$ that $|g(x)-g_2(x)|<\epsilon/4$, and hence 
\begin{align*}
	0=\frac{\epsilon}{4}-\frac{\epsilon}{4}<\overbrace{(f-g_2)(x)+(g_2-g)(x)}^{f(x)-g(x)}<\frac{3\epsilon}{4}+\frac{\epsilon}{4}=\epsilon.
\end{align*}
Next, assume that $g_2(x)\le 0$. Then, $g(x)=\psi(g_2(x))=0$, and hence $g_2(x)\le g(x)$. Thus, \[0\le f(x)= f(x)-g(x)\le f(x)-g_2(x)<\frac{3\epsilon}{4}<\epsilon.\]
Therefore,  we have that $0\le f-g<\epsilon$.
\end{proof}
\subsection*{Proof of Theorem \ref{maintheorem}}	Let $\phi\in C_c^1(\Omega,\mathbb{R}^n)$ with $|\phi|\le f_X^{2d}$. We define $R:\Omega\to \mathbb{R}$ by setting $R(\x)=\phi(\x)g(\x)$ for any $\x\in\Omega$. Then, by Equation (\ref{divergenceofR}),
$
\div(R)=\nabla g\cdot \phi +g\div(\phi)$.
Such as before, let $\{C_i\}_{i=0}^{\infty}$ be an admissible compact cover for $\Omega$. Then, by Proposition \ref{supp(phi)bd(c_i)}, there exists a positive integer $N_0$ in such a way that $i\ge N_0$ implies that $\phi_{|_{\mathscr{S}_i}}=0$, where $\mathscr{S}_i=\bd(C_i)$. For any $i\ge N_0$, 
it follows from the divergence theorem that
\[\int_{C_i} \div(R)(\x)\,\d \x=\int_{\mathscr{S}_i} R(\x)\cdot \widehat{N}(\x)\,\d S.\]

Now, since $\phi$ and consequently $R$ are $0$ on $\mathscr{S}_i$, the right side of the above equality is 0. Hence, we have that $\int_{C_i}\div(R)\,\d \x=0$, which implies that
\[ \int_{C_i} g\div(\phi)\,\d x = -\int_{C_i}\nabla g\cdot \phi\,\d x.\]
Now, we have that 
\[\left|\int_{C_i}\nabla g\cdot \phi\,\d \x\right|\le \int_{C_i}|\nabla g\cdot \phi|\,\d \x\le \int_{C_i}|\nabla g|| \phi|\,\d \x\le \int_{C_i}|\nabla g| f_{\X}^{2d}\,\d \x.\]
It follows that 
\[\int_{\Omega}g\div(\phi)\,\d \x=\lim_{i\to\infty}\int_{C_i}g\div(\phi)\,\d \x\le \lim_{i\to\infty}\int_{C_i}|\nabla g|f_{\X}^{2d}\,\d\x=\int_{\Omega}|\nabla g|f_{\X}^{2d}\,\d\x.\]
Therefore, $\PEACE_d(\X\to Y)\le\int_{\Omega}\left|\nabla g(x)\right|f_X^{2d}(\x)\,\d \x$. 

Conversely, define
\[\theta_k:=-\frac{\nabla g}{|\nabla g|_k}f_X^{2d},\qquad k=1,2,3,\ldots,\]
where for any  $\mathbf{x}\in\mathbb{R}^n$, by $|\x|_k$ we mean $\sqrt{x_1^2+\cdots+x_n^2+1/k}$. One could see that $|\x|_k=\sqrt{|\x|^2+1/k}$, and hence $1/\sqrt{k}\le|\x|_k\le |\x|+1/\sqrt{k} $. 
Now, fix a positive integer $k$. Obviously, $|\theta_k|\le f_X^{2d}$ but it might $\theta_k\notin C_c^1(\Omega,\mathbb{R}^n)$. 
In the rest of this proof, we overcome this issue. 

For each $1\le i\le n$, let $p_i=-\rond g/\rond x_i$.  Now, by Corollary~\ref{need}, there exist $h_k, q_k^{(i)}\in C_c(\Omega,\mathbb{R})$ with $0\le h_k\le f_{\X}^{2d}$ and $ |q_k^{(i)}|\le |p_i|$ satisfying $h_k|_{C_k}=f_{\X}^{2d}|_{C_k}$ and $q_k|_{C_k}=p_i|_{C_k}$ for each $1\le i\le n$. Next, it follows from Lemma~\ref{stack} that there exists a non-negative function $\widetilde{h}_k\in C_c^1(\Omega,\mathbb{R})$ with $0\le h_k-\widetilde{h}_k<1/k$. Also, by Corrolary \ref{corr}, there exists $g_k^{(i)}\in C^1_c(\Omega,\mathbb{R})$ with $|g_k^{(i)}-q_k^{(i)}|<1/k^2$ for each $1\le i\le n$.  Now, let $g_k=(g_k^{(1)},\ldots, g_k^{(n)})$ and $\widetilde{g}_k= g_k/| g_k|_k$. We note that $\widetilde{g}_k\in C_c^1(\Omega,\mathbb{R}^n)$ and 
$0\le \widetilde{h}_k\le h_k\le f_{\X}^{2d}$. Now, we define $\widetilde{\theta}_k= \widetilde{g}_k\widetilde{h}_k$. Clearly, we have that $\widetilde{\theta}_k\in C_c^1(\Omega,\mathbb{R}^n)$ and $|\widetilde{\theta}_k|\le f_{\X}^{2d}$. 

By Remark \ref{remarkforexistenceofthecomapctsequence}, let $\{C_k\}_{k=1}^{\infty}$ be an admissible compact cover for $\Omega\backslash B$, where
\[C_k\subseteq \left\{\x\in\Omega: |\nabla g(\x)|>\frac{1}{\sqrt[4]{k}}\right\},\quad B=\{\x\in\Omega: \nabla g(\x)=0\},\quad k=1,2,3,\ldots\] 
Assume that
$S_k=\{\x\in\Omega:|\nabla g|<1/k\}$ for any $k$. Let $M_1$ and $M_2$ be bounds for $|\nabla g|$ and $f_{\X}^{2d}$, respectively. Also, let $A=\int_{\Omega}\,\d\x$.
Assume that $H_k=-\nabla g\cdot \widetilde{\theta}_k$. Then, $|H_k|\le M_2/k$ on $S_k$. We note that
\begin{align*}
	\left|\int_{\Omega}H_k\,\d\x-\int_{\Omega\backslash B}H_k\,\d\x\right|&=	\left|	\int_{S_k}H_k\,\d\x-	\int_{(\Omega\backslash B)\cap S_k}H_k\,\d\x\right|\\
	&\le	\left|	\int_{S_k}H_k\,\d\x\right|+	\left|\int_{(\Omega\backslash B)\cap S_k}H_k\,\d\x\right|\le2	\int_{S_k}	\left|H_k\right|\,\d\x\le \frac{2M_2A}{k}.
\end{align*}
Similarly, we have that 
\[	\left|\int_{\Omega}|\nabla g|f_{\X}^{2d}\,\d\x-\int_{\Omega\backslash B}|\nabla g|f_{\X}^{2d}\,\d\x\right|\le \frac{2M_2A}{k}.\]
There exists a positive integer $N_0$ in such a way that $k\ge N_0$ implies that 
\[\left|\int_{\Omega\backslash B}-\nabla g\cdot \widetilde{\theta}_k\,\d\x-\int_{C_k}-\nabla g\cdot \widetilde{\theta}_k\,\d\x\right|<\frac{1}{k},\quad \left|\int_{\Omega\backslash B}|\nabla g|f_{\X}^{2d}\,\d\x- \int_{C_k}|\nabla g|f_{\X}^{2d}\,\d\x\right|<\frac{1}{k}. \]
 Assume that 
\[L_k=\left|\int_{\Omega}-\nabla g\cdot \widetilde{\theta}_k\,\d\x-\int_{\Omega}|\nabla g|f_{\X}^{2d}\,\d\x\right|,\qquad k=1,2,3,\cdots\]
To prove the theorem, it is enough to show that $\lim _{k\to \infty}L_k=0$. We have that $L_k\le L_k^{(1)}+L_k^{(2)}+L_k^{(3)}+L_k^{(4)}+L_k^{(5)}$, where 
\begin{align*}
		L_k^{(1)}&=\left|\int_{\Omega}-\nabla g\cdot \widetilde{\theta}_k\,\d\x-\int_{\Omega\backslash B}-\nabla g\cdot \widetilde{\theta}_k\,\d\x\right|<\frac{2M_2A}{k},\\
		L_k^{(2)}&=\left|\int_{\Omega}|\nabla g|f_{\X}^{2d}\,\d\x- \int_{\Omega\backslash B}|\nabla g|f_{\X}^{2d}\,\d\x\right|<\frac{2M_2A}{k},\\
	L_k^{(3)}&=\left|\int_{\Omega\backslash B}-\nabla g\cdot \widetilde{\theta}_k\,\d\x-\int_{C_k}-\nabla g\cdot \widetilde{\theta}_k\,\d\x\right|<\frac{1}{k},\\ L_k^{(4)}&=\left|\int_{\Omega\backslash B}|\nabla g|f_{\X}^{2d}\,\d\x- \int_{C_k}|\nabla g|f_{\X}^{2d}\,\d\x\right|<\frac{1}{k},\\ L_k^{(5)}&=\left|\int_{C_k}-\nabla g\cdot \widetilde{\theta}_k\,\d\x-\int_{C_k}|\nabla g|f_{\X}^{2d}\,\d\x\right|.
\end{align*}
 Note that on $C_k$, $q_k^{(i)}=p_i=-\rond g/\rond x_i$ for each $1\le i\le n$. Hence, $|g_k+\nabla g|\le \sqrt{n}/k^2$. 
 Furthermore, one could see that $\left||g_k|_k-|\nabla g|_k\right|\le \left||g_k|-\nabla g\right|\le |g_k+\nabla g|$, and hence, on $C_k$, we have that
\begin{align*}
	\left|\widetilde{g}_k+\frac{\nabla g}{|\nabla g|_k}\right|&=\left|\frac{|\nabla g|_k g_k+| g_k|_k\nabla g}{|g_k|_k|\nabla g|_k}\right|\le\left|\frac{|\nabla g|_k g_k+| g_k|_k\nabla g}{\frac{1}{k}}\right|\\
	&=k\left||\nabla g|_k (g_k+\nabla g)+\nabla g(| g_k|_k-|\nabla g|_k)\right|\\
	&\le k\left(\left(|\nabla g|+\frac{1}{\sqrt{k}}\right)|g_k+\nabla g|+|\nabla g|| g_k+\nabla g|\right)\\
	&\le k\left(2|\nabla g|+1\right)|g_k+\nabla g|\le \frac{M_3}{k},
\end{align*}
where $M_3=(2M_1+1)\sqrt{n}$. Now, we note that
$L_k^{(5)}\le L_k^{(5,1)}+L_k^{(5,2)}+L_k^{(5,3)}$, where 
\begin{align*}L_k^{(5,1)}&=\left|\int_{C_k}-\nabla g\cdot \widetilde{\theta}_k\,\d\x-\int_{C_k}\nabla g\cdot \left(\frac{\nabla g}{|\nabla g|_k}\widetilde{h}_k\right)\,\d\x\right|,\\
L_k^{(5,2)}&=\left|\int_{C_k}\nabla g\cdot \left(\frac{\nabla g}{|\nabla g|_k}\widetilde{h}_k\right)\,\d\x-\int_{C_k}\nabla g\cdot \left(\frac{\nabla g}{|\nabla g|_k}f_{\X}^{2d}\right)\,\d\x\right|,\\
L_k^{(5,3)}&=\left|\int_{C_k}\nabla g\cdot \left(\frac{\nabla g}{|\nabla g|_k}f_{\X}^{2d}\right)\,\d\x-\int_{C_k}|\nabla g|f_{\X}^{2d}\,\d\x\right|,\\
\end{align*}
We have that
\begin{align*}
	L_k^{(5,1)}&=\left|\int_{C_k}-\nabla g\cdot\left(\widetilde{g}_k+\frac{\nabla g}{|\nabla g|_k}\right)\widetilde{h}_k\,\d\x\right|\le \frac{M_4}{k},\quad M_4=M_1M_2M_3A,\\
	L_k^{(5,2)}&=\left|\int_{C_k}\frac{|\nabla g|^2}{|\nabla g|_k}(\widetilde{h}_k-f_{\X}^{2d})\,\d\x\right|\le \frac{M_1A}{k},\\
	L_k^{(5,3)}&=\left|\int_{C_k}|\nabla g|f_{\X}^{2d}\left(\frac{|\nabla g|}{|\nabla g|_k}-1\right)\,\d\x\right|.
\end{align*}
For any $\x\in\mathbb{R}$ with $|\x|>1/\sqrt[4]{k}$, 
\[\frac{|\x|_k-|\x|}{|\x|_k}=\frac{|\x|_k^2-|\x|^2}{|\x|_k(|\x|_k+|\x|)}=\frac{\frac{1}{k}}{|\x|_k(|\x|_k+|\x|)}\le \frac{\frac{1}{k}}{2|\x|^2}< \frac{\frac{1}{k}}{\frac{2}{\sqrt{k}}}=\frac{1}{2\sqrt{k}}.\]
Thus, on $C_k$,  $(|\nabla g|_k-|\nabla g|)/|\nabla g|_k<1/(2\sqrt{k})$. Hence, 
$L_k^{(5,3)}\le (M_1M_2A)/(2\sqrt{k})$. 
Therefore, $L_k\le \left((4M_2A+2+M_4+M_1A)/k\right)+ (M_1M_2A)/(2\sqrt{k})$, which implies that $\lim_{k\to\infty} L_k=0$, and hence the proof is complete. 
\subsection*{Proof of Corollary \ref{unboundedmain} }
	Let $\{\Omega_k\}_{k=0}^{\infty}$ be an increasing sequence of bounded open subsets of $\mathbb{R}^n$ with $\Omega=\bigcup_{k=0}^{\infty}\Omega_k$ (for instance, one could assume that $\Omega_k=B(0,k)\cap \Omega$ for any $k$). First, assume that $\PEACE_d(\X\to Y)<\infty$, and  let $\epsilon>0$. Then, there exists $\phi\in C_c^1(\Omega,\mathbb{R}^n )$ with $|\phi|\le f_{\X}^{2d}$ and
\[\PEACE_d(\X\to Y)-\epsilon<\int_{\Omega}g\div(\phi)\,\d\x.\]
Now, we show that there exists $k_0$ with $\Supp(\phi)\subseteq \Omega_{k_0}$. On the contrary, assume that $\Supp(\phi)\nsubseteq \Omega_k$ for any $k$. Then, there exists $\x_k\in \Supp(\phi)\backslash\Omega_k$ for any $k$. It follows from the comnpactness of $\Supp(\phi)$ that there exists a subsequence $\{\x_{k_i}\}_{i=0}^{\infty}$ convergent to a point of $\Supp(\phi)$ such as $\x$. Since, $\Supp(\phi)\subseteq \Omega$, there exists $N_0$ with $\x\in \Omega_{N_0}$. Since $\{\x_{k_i}\}_{i=0}^{\infty}$ converges to $\x$ and $\Omega_{N_0}$ is open, there exists a positive integer $N\ge N_0$ such that $i\ge N$ implies that $\x_{k_i}\in\Omega_{N_0}$. Therefore, $\x_{k_N}\in\Omega_{N_0}\subseteq\Omega_N\subseteq \Omega_{k_N}$, a contradiction! Hence, there exists $k_0$ with $\Supp(\phi)\subseteq \Omega_{k_0}$. It follows from the above discussion and Theorem \ref{maintheorem} that
\begin{align*}
	\PEACE_d(\X\to Y)-\epsilon&<\int_{\Omega}g\div(\phi)\,\d\x=\int_{\Omega_{k_0}}g\div(\phi)\,\d\x\\
	&\le\PEACE_d(\X|_{\Omega_{k_0}}\to Y)=\int_{\Omega_{k_0}}|\nabla g|f_{\X}^{2d}\,\d\x\le \int_{\Omega}|\nabla g|f_{\X}^{2d}\,\d\x,
\end{align*}
which implies that 
\[\PEACE_d(\X\to Y)\le \int_{\Omega}|\nabla g|f_{\X}^{2d}\,\d\x.\]
Conversely, there exist $k_1$ and $\psi\in C_c^1(\Omega_{k_1},\mathbb{R}^n)$ with $|\psi|\le f_{\X}^{2d}$ such that
\[\int_{\Omega}|\nabla g|f_{\X}^{2d}\,\d\x<\int_{\Omega_{k_1}}|\nabla g|f_{\X}^{2d}\,\d\x+\frac{\epsilon}{2}<\int_{\Omega_{k_1}}g\div(\psi)\,\d\x+\epsilon.\]
Define $\widetilde{\psi}\in C_c^1(\Omega,\mathbb{R}^n)$ with $|\widetilde{\psi}|\le f_{\X}^{2d}$ by setting  $\widetilde{\psi}|_{\Omega_{k_1}}=\psi|_{\Omega_{k_1}}$ and $\widetilde{\psi}|_{\Omega\backslash \Omega_{k_1}}\equiv 0$. 
Then, we have that 
\[\int_{\Omega_{k_1}}g\div(\psi)\,\d\x=\int_{\Omega}g\div(\widetilde{\psi})\,\d\x\le \PEACE_d(\X\to Y),\]
which implies that 
\[\int_{\Omega}|\nabla g|f_{\X}^{2d}\,\d\x<\PEACE_d(\X\to Y)+\epsilon,\]
and hence 
\[\int_{\Omega}|\nabla g|f_{\X}^{2d}\,\d\x\le\PEACE_d(\X\to Y).\]
Consequently, we have that 
\[\PEACE_d(\X\to Y)=\int_{\Omega}|\nabla g|f_{\X}^{2d}\,\d\x.\]
Now, assume that $\PEACE_d(\X\to Y)=\infty$. We show that  $\int_{\Omega}|\nabla g|f_{\X}^{2d}\,\d\x=\infty$. For any $M>0$ there exists $\phi_M\in C_c^1(\Omega,\mathbb{R}^n)$ with $|\phi_M|\le f_{\X}^{2d}$ such that $\int_{\Omega}g\div(\phi_M)\,\d\x>M$. It follows that there exists a positive integer $N_0$ for which $k\ge N_0$ implies that $\Supp(\phi_M)\subseteq \Omega_k$ and  $\int_{\Omega_k}g\div(\phi_M)\,\d\x>M$, and hence $\int_{\Omega_k}|\nabla g|f_{\X}^{2d}\,\d\x=\PEACE_d(\X|_{\Omega_k}\to Y)>M$. It follows that 
\[\int_{\Omega}|\nabla g|f_{\X}^{2d}\,\d\x=\lim_{k\to\infty}\int_{\Omega_k}|\nabla g|f_{\X}^{2d}\,\d\x=\infty.\]
\subsection*{Proof of Corollary \ref{lastcor}}
The first part of the corollary follows from the fact that the gradient of a function is 0  if and only if that function is locally constant. To prove the second part, first we note that $\Gamma$ is an open subset of $\Omega$, and hence it is connected by the connectedness of $\Omega$. Now, on the contrary, assume that $g|_{\Gamma}$ is not constant and $\alpha$ is a value of $g|_{\Gamma}$. Then, $U=\{\x\in\Gamma: g(\x)=\alpha\}$ is both open and close in $\Gamma$, which contradicts with connectedness of $\Gamma$. Therefore, $g|_{\Gamma}$ is constant.
\section{Proofs of Other Results}\label{other}
\subsection*{Proof of Proposition \ref{finiteadditive}}
	Let $\phi\in C_c^1(\Omega,\mathbb{R}^n)$ with $|\phi|\le f_{\X}^{2d}$. Then, one could see that
$\Supp(\phi|_{\Omega_i})=\Supp(\phi)\cap \Omega_i$ for any $i$. Thus, $\phi|_{\Omega_i}\in C_c^1(\Omega_i,\mathbb{R}^n)$ for any $i$. It follows that
\begin{align*}
	\int_{\Omega}g\div(\phi)\,\d\x&=\sum_{i=0}^{\infty}\int_{\Omega_i}g\div(\phi)\,\d\x\le\sum_{i=0}^{\infty} \PEACE_d(\X|_{\Omega_i}\to Y),
\end{align*}
which implies that 
\[\PEACE_d(\X\to Y)\le \sum_{i=0}^{\infty} \PEACE_d(\X|_{\Omega_i}\to Y).\]
Conversely, first, assume that for any $i$,  $\PEACE_d(\X|_{\Omega_i}\to Y)<\infty$, and let $\epsilon>0$. Then, there exists $\phi_i\in C_c^1(\Omega_i,\mathbb{R}^n)$ with $|\phi_i|\le f_{\X}^{2d}|_{\Omega_i}$  in such a way that
\begin{equation}\label{eq-finiteadditive}
	\PEACE_d(\X|_{\Omega_i}\to Y)-\frac{\epsilon}{2^{i+1}}<\int_{\Omega_i}g\div(\phi_i)\,\d\x,\qquad i=0,1,2,\cdots
\end{equation}
Now, we define $\psi:\Omega\to\mathbb{R}^n$ by setting $\psi|_{\Omega_i}=\phi_i$ for any $i$.  Then, $|\psi|\le f_{\X}^{2d}$ and $\Supp(\psi)= \bigcup_{i=0}^{\infty}\Supp(\phi_i)\subseteq \bigcup_{i=0}^{\infty}\Omega_i=\Omega$. Hence, we have that
\begin{small}\begin{align*}
		\sum_{i=0}^{\infty}\PEACE_d(\X|_{\Omega_i}\to Y)&<\sum_{i=0}^{\infty}\int_{\Omega_i}g\div(\phi_i)\,\d\x=\sum_{i=0}^{\infty}\int_{\Omega_i}g\div(\psi|_{\Omega_i})\,\d\x+\epsilon\\
		&=\int_{\Omega}g\div(\psi)\,\d\x+\epsilon\le\PEACE_d(\X\to Y)+\epsilon,
	\end{align*}
\end{small}
which implies that 
\[	\sum_{i=0}^{\infty}\PEACE_d(\X|_{\Omega_i}\to Y)\le \PEACE_d(\X\to Y). \]
Now, assume that one of the values $\PEACE_d(\X|_{\Omega_i}\to Y)$  is $\infty$. Without loss of generality, assume that $\PEACE_d(\X|_{\Omega_0}\to Y)=\infty$. Then, for any $M>0$ there exists $\phi_M\in C_c^1(\Omega_0,\mathbb{R}^n)$ with $|\phi_M|\le f_{\X}^{2d}$ in such a way that $\int_{\Omega_0}g\div(\phi_M)\,\d\x>M$. Now, we define $\psi_M:\Omega\to\mathbb{R}$ by setting $\psi_M|_{\Omega_0}=\phi_M$, and $\psi_M|_{\Omega\backslash \Omega_0}\equiv 0$. Then, $\psi_M\in C_c^1(\Omega,\mathbb{R}^n)$ with $|\psi_M|\le f_{\X}^{2d}$. Note that 
\[\int_{\Omega}g\div(\psi_M)\,\d\x=\int_{\Omega_0}g\div(\phi_M)\,\d\x>M,\]
which implies that $\PEACE_d(\X\to Y)=\infty$. 
\subsection*{Proof of Proposition \ref{subadditivityofopensets}}
	Let $\phi\in C_c^1(\Omega,\mathbb{R}^n)$ with $|\phi|\le f_{\X}^{2d}$.  By compactness of $\Supp(\phi)$, it is covered by finitely many of the aforementioned open sets. Assume that $\Supp(\phi)\subseteq\bigcup_{i=0}^{m} \Omega_i$. Then, by the \textit{partition of unity theorem} (see \cite[Theorem  16.3]{munkres1991analysis}), there exists $\alpha_i\in C_c^1(\Omega,\mathbb{R})$ with $\Supp(\alpha_i)\subseteq \Omega_i$ and $\alpha_i\ge 0$ for any $0\le i\le m$, in such a way that $\sum_{i=0}^m\alpha_i=1$. Thus, $\phi=\sum_{i=0}^m\phi_i$, where $\phi_i=\phi\alpha_i\in C_c^1(\Omega_i,\mathbb{R}^n)$. Note that $|\phi_i|=\alpha_i|\phi|\le |\phi|\le f_{\X}^{2d}$. It follows that 
\begin{align*}
	\int_{\Omega}g\div(\phi)\,\d\x&= \sum_{i=0}^m \int_{\Omega}g\div(\phi_i)\,\d\x=\sum_{i=0}^m \int_{\Omega_i}g\div(\phi_i)\,\d\x\le \sum_{i=0}^{\infty}\PEACE_d(\X|_{\Omega_i}\to Y).
\end{align*}
Therefore, we have that 
\[\PEACE_d(\X\to Y)\le \sum_{i=0}^{\infty}\PEACE_d(\X|_{\Omega_i}\to Y).\]
\subsection*{Proof of Proposition \ref{monoton}}
Let $\phi\in C_c^1(\Gamma,\mathbb{R}^n)$ with $|\phi|\le f_{\X}^{2d}$. Then, we define $\psi:\Omega\to\mathbb{R}^n$ with $\psi|_{\Gamma}=\phi$ and $\psi|_{\Omega\backslash\Gamma}\equiv 0$. Then, $\psi\in C_c^1(\Omega,\mathbb{R}^n)$, and we have that
\[\int_{\Gamma}g\div(\phi)\,\d\x=\int_{\Omega}g\div(\psi)\,\d\x\le \PEACE_d(\X\to Y),\]
which implies that
\[\PEACE_d(\X|_{\Gamma}\to Y)\le\PEACE_d(\X\to Y).\]
\subsection*{Proof of Theorem \ref{outermeasure}}
	Clearly, $\mu(\emptyset)=0$. Now, we show that $\mu$ is monotone. Let $E,F\in\mathcal{P}(\Omega)$ with $E\subseteq F$. Let $\epsilon>0$ be arbitrary. Then, there exists $ V\in\tau$ with $F\subseteq V$ in such a way that $\mu(F)+\epsilon>\lambda(V)$. We have that $E\subseteq V$, since $E\subseteq F$ and $F\subseteq V$. It follows that $\lambda(V)\ge \mu(E)$. Therefore, $\mu(F)+\epsilon>\mu(E)$, and hence $\mu(F)\ge\mu(E)$. 

Now, we show the countable subadditivity of $\mu$. Let $\{E_i\}_{i=0}^{\infty}$ be a countable family in $\mathcal{P}(\Omega)$ and $E=\bigcup_{i=0}^{\infty}E_i$.  If there exists $i_0$ with $\mu(E_{i_0})=\infty$, then there is nothing to prove. Otherwise, let $\mu(E_i)<\infty$ for any $i$.  Let $\epsilon>0$. Then, for any $i$, there exists $V_i\in\tau$  containing $E_i$ in such a way that $\mu(E_i)+\epsilon/2^{i+1}>\lambda(V_i)$. It follows that $\epsilon+\sum_{i=0}^{\infty}\mu(E_i)>\sum_{i=0}^{\infty}\lambda(V_i)$, while by Proposition \ref{subadditivityofopensets},  $\sum_{i=0}^{\infty}\lambda(V_i)>\lambda(V)$, where $V=\bigcup_{i=0}^{\infty}V_i$. Hence, $\epsilon+\sum_{i=0}^{\infty}\mu(E_i)>\lambda(V)$.  We note that $E\subseteq V\in\tau$, which implies that $\mu(E)\le \lambda(V)$. Therefore, we have that $\mu(E)<\epsilon+\sum_{i=0}^{\infty}\mu(E_i)$, and hence $\mu(E)\le\sum_{i=0}^{\infty}\mu(E_i)$.

Now, we show that $\mu$ is a Borel measure. 
By Caratheodory's criterion (see \cite[Section 1.1]{evans1992studies}), it is enough to show that $\mu(A\cup B)=\mu(A)+\mu(B)$ for any $A,B\in\mathcal{P}(\Omega)$ with $\delta=\mathrm{dist}(A,B)>0$.  First, we note that $\mu(A\cup B)\le \mu(A)+\mu(B)$ by the subadditivity of $\mu$. Now, first, assume that $\mu(A),\mu(B)<\infty$, and let $\epsilon>0$. Then, there exists $\Gamma\in\tau$ with $A\cup B\subseteq \Gamma$  that $\mu(A\cup B)+\epsilon>\lambda(\Gamma)$. For any $x\in A$, there exists $r_{\x}>0$ with $B(\x,r_{\x})\subseteq \Gamma$. Set $r'_{\x}=\min\{r_{\x}, \delta\}$. Consider similar notations for points in $B$. Now, assume that $U=\bigcup_{\x\in A}B(\x;r'_{\x})$ and $V=\bigcup_{\x\in B}B(\x;r'_{\x})$. Then, $U,V\in\tau$ with $A\subseteq U$ and $B\subseteq V$, while $U\cap V=\emptyset$ and $U\cap V\subseteq \Gamma$. Thus, $\mu(A\cup B)+\epsilon>\lambda(U\cup V)$. It follows from the latter and Proposition \ref{subadditivityofopensets} that 
\[\mu(A\cup B)+\epsilon>\lambda(U\cup V)=\lambda(U)+\lambda(V)\ge \mu(A)+\mu(B),\]
and hence 	$\mu(A\cup B)\ge \mu(A)+\mu(B) $.  Now, assume that one of $\mu(A)$ and $\mu(B)$ is $\infty$. Then, it follows from the monotonicity of $\mu$ that $\mu(A\cup B)=\infty$, and hence again the equality $\mu(A)+\mu(B)= \mu(A\cup B)$ holds. 

Finally, we show that $\mu$ is Borel regular. Let $E\in\mathcal{P}(\Omega)$. First, assume that $\mu(E)<\infty$. For any positive integer $k$, there exists $V_k\in\tau$ with $E\subseteq V_k$ and $\mu(E)+1/k>\lambda(V_k)$. Set $U_k=\bigcap_{i=0}^kV_i$ for any positive integer $k$. Then, $\cdots \subseteq U_3\subseteq U_2\subseteq U_1$ are in $\tau$, and $\lambda(U_k)\le \lambda(V_k)$ for any $k$. Since, $\mu$ is a Borel measure, we have that (see Remark~\ref{mu=lambda}):
\[  \mu\left(\bigcap_{k=0}^{\infty}U_k\right)-\mu(E)=\lim_{k\to\infty}\left(\mu\left(U_k\right)-\mu(E)\right)= \lim_{k\to\infty}\left(\lambda\left(U_k\right)-\mu(E)\right)=0,\]
which implies that $\mu(E)=\mu\left(\bigcap_{k=0}^{\infty}U_k\right)$. Now, if $\mu(E)=\infty$, then $\mu(E)=\mu(\Omega)=\infty$.
Therefore, $\mu$ is Borel regular. 
\subsection*{Proof of Proposition \ref{iso}}
(1) Let $\Gamma=\pi(\Omega)$, where $\pi:\mathbb{R}^{n+m}\to\mathbb{R}^n$ defined by setting $\pi(\x,\z)=\x$. Then, $\oo{\Gamma}=\pi_1(\oo{\Omega})$. Assume that $h:\oo{\Gamma}_{*}\to \oo{\Gamma}$ is a diffeomorphism. Denote the probability density functions of $\X$ given $\Z$, and $\W$ given $\Z$ by $f$ and $\widetilde{f}$, respectively.  Define $\widetilde{g}(\mathbf{W},\Z)=g(h(\W),\Z)$. One could see that 
\[\frac{\partial\widetilde{g}_{in}}{\partial \w}(\w,\z)=\mathrm{Jac}(h)(\w)\frac{\partial g_{in}}{\partial \x}(h(\w),\z).\]
By the change of variable formula for probability density functions \cite[Section 5.4]{devore2012modern}, we have that 
\begin{align*}
	\widetilde{f}(\w|\z)&=\left|\det\left(\mathrm{Jac}(h)(\w)\right)\right|f(h(\mathbf{w})|\z).
\end{align*}
We note that $\PIEV_d^{\z}(\mathbf{W}\to Y)=\int_{\Gamma_*}|(\partial \widetilde{g}_{in}/\partial \w)(\w,\z)|\widetilde{f}(\mathbf{w}|\z)^{2d}\,\d\mathbf{w}$. 
Thus, by the change of variable formula, we have that 
\begin{small}	\begin{align*}
		\PIEV_d^{\z}(\W\to Y)&=\int_{\Gamma_*}\left|\mathrm{Jac}(h)(\w)\frac{\partial g_{in}}{\partial \x}(h(\w),\z) \right|\left(\left|\det\left(\mathrm{Jac}(h)(\w)\right)\right|f(h(\mathbf{w})|\z)\right)^{2d}\,\d\w\\
		&=\int_{\Gamma}\left|\mathrm{Jac}(h)(h^{-1}(\x))\frac{\partial g_{in}}{\partial \x}(\x,\z) \right|\left(\left|\det\left(\mathrm{Jac}(h)(h^{-1}(\x))\right)\right|f(\x|\z)\right)^{2d}\,\d\w
\end{align*}\end{small}

(2) Now, assume that $h$ is an affine map of the form of $h(\w)=A\w+\mathbf{a}$. Then, $\mathrm{Jac}(h)(h^{-1}(\x))=A^T$ for any $\x$, which implies the claimed equality in the proposition. 

(3) As we explained in the preliminaries, each onto isometry of $\mathbb{R}^n$ is of the form of an affine map $h(\w)=A\w+\mathbf{a}$, where $A$ is an orthogonal matrix. Thus, $|A^T\mathbf{v}|=|\mathbf{v}|$ for any vector $\mathbf{v}\in\mathbb{R}^n$ and $|\det(A)|=1$. It follows that
\[	\PIEV_d^{\z}(\mathbf{W}\to Y)=\int_{\Gamma}\left|\frac{\partial g_{in}}{\partial \x}(\x,\z) \right| f(\x|\z)^{2d}\,\d\w=\PIEV_d^{\z}(\X\to Y),\]
and hence $\PEACE_d(\mathbf{W}\to Y)=\PEACE_d(\mathbf{X}\to Y)$. 
\subsection*{Proof of Proposition \ref{totalvariation-discrete}}
	Let
\[
\psi^{(i)}_{j_1,\ldots,j_n}:=\frac{1}{2^{n-1}}\left(\sum_{\substack{\x\in F(x_{1j_1},\ldots,x_{nj_n};x_{ij_i})\\x'_i=x_{i,j_i-1},\;\forall\,k\neq i\;x'_k=x_k}}\frac{g(\x)-g(\x')}{\Delta x_{ij_i}}\right),\quad i=1,\ldots,n.\]
Let $\phi^{(i)}_{j_1,\ldots,j_n}:=\psi^{(i)}_{j_1,\ldots,j_n}/|\psi_{j_1,\ldots,j_n}|$ for $\psi_{j_1,\ldots,j_n}\neq 0$, and $\phi^{(i)}_{j_1,\ldots,j_n}:=0$ for $\psi_{j_1,\ldots,j_n}=0$, where $\psi_{j_1,\ldots,j_n}=\left(\psi^{(1)}_{j_1,\ldots,j_n},\ldots,\psi^{(n)}_{j_1,\ldots,j_n}\right)$. Then, for $\psi_{j_1,\ldots,j_n}\neq 0$, we have that
\[\mathrm{Flux}_{\phi}^{(i)}(x_{1j_1},\ldots,x_{nj_n})=\frac{\left(\psi^{(i)}_{j_1,\ldots,j_n}\right)^2}{|\psi_{j_1,\ldots,j_n}|}\mathrm{Vol}(C(x_{1j_1},\ldots,x_{nj_n})),\quad i=1,\ldots,n,\]
and for $\psi_{j_1,\ldots,j_n}= 0$, we have that $\mathrm{Flux}_{\phi}^{(i)}(x_{1j_1},\ldots,x_{nj_n})=0$.
Hence,
\begin{align*}
	\mathrm{Flux}_{\phi}(x_{1j_1},\ldots,x_{nj_n})&=|\psi_{j_1,\ldots,j_n}|\mathrm{Vol}(C(x_{1j_1},\ldots,x_{nj_n}))\\
	&=\sqrt{\mathrm{Flux}^{(1)}(x_{1j_1},\ldots,x_{nj_n})^2+\cdots+\mathrm{Flux}^{(n)}(x_{1j_1},\ldots,x_{nj_n})^2}\\
	&=\mathrm{Flux}(x_{1j_1},\ldots,x_{nj_n}).
\end{align*}
Now, assume that $\eta:\Gamma\to\mathbb{R}^n$ with $|\eta|\le 1$ is arbitrary. Then, it follows from the Cauchy–Schwarz inequality that
\begin{align*}
	\mathrm{Flux}_{\eta}(x_{1j_1},\ldots,x_{nj_n})&=\sum_{i=1}^n \mathrm{Flux}^{(i)}(x_{1j_1},\ldots,x_{nj_n})\eta_{j_1,\ldots,j_n}^{(i)}\\
	&\le \mathrm{Flux}_{\phi}(x_{1j_1},\ldots,x_{nj_n})|\eta_{j_1,\ldots,j_n}|\le \mathrm{Flux}_{\phi}(x_{1j_1},\ldots,x_{nj_n}). 
\end{align*}
It follows that $\TV(g)=\TV_{\phi}(g)$, and consequently, the proof is complete. 
\subsection*{Proof of Theorem \ref{dis-con}}
	Let $\phi\in C_c^1(\Omega,\mathbb{R}^n)$ with $|\phi|\le f_{\X}^{2d}$, and let  $\epsilon>0$. Also, let $\{C_j\}_{j=0}^{\infty}$ be an admissible compact cover for $\Omega$. There exists a positive number $N_0$ for which $j\ge N_0$ yields that $\Supp(\phi)\subseteq C_j\subseteq\mathrm{In}(C_{j+1})$. Let $j\ge N_0+1$. It follows from $g\in C^1(\oo{\Omega},\mathbb{R})$ and the continuity of $\phi$ that there exists $0<\delta<\mathrm{dist}(\Supp(\phi),\bd(C_j))$ in such a way that $|\x-\x'|<\delta$ implies that $|\phi(\x)-\phi(\x')|<\epsilon$ and $|\rond g/\rond x_i(\x)-\rond g/\rond x_i(\x')|<\epsilon$ for any $1\le i\le n$. Now, let  $C_j\subseteq C$, where $C=\prod_{i=1}^n [a_i,b_i]$.   Let $\widetilde{g}:C\to\mathbb{R}$ defined by setting $\widetilde{g}|_{\Omega}=g$ and $\widetilde{g}|_{C\backslash\Omega}\equiv 0$. Similarly, we can define $\widetilde{\phi}$. Let $P\in\mathcal{P}(C)$  with $\lVert P\rVert=\max\{x_{ij}^{(P)}-x_{i,j-1}^{(P)}: 1\le i\le n,\;2\le j\le n_i^{(P)}\}<\delta/\sqrt{n}$. Let $\x_{j_1,\ldots,j_n}^{(P)}$ be a fixed point in $C(x_{1j_1}^{(P)},\ldots,x_{nj_n}^{(P)})$. 
First, assume that  $C(x_{1j_1}^{(P)},\ldots,x_{nj_n}^{(P)})\subseteq C_j$.
By the mean value theorem, for any $\x\in F(x_{1j_1}^{(P)},\ldots,x_{nj_n}^{(P)};x_{ij_i}^{(P)})$ and $\x'\in F(x_{1j_1}^{(P)},\ldots,x_{nj_n}^{(P)};x_{i,j_i-1}^{(P)})$ with $x'_i=x_{i,j_i-1}^{(P)}$ and $x'_k=x_k$ for $k\neq i$, there exsits $\theta_{ij_i}^{(P)}\in(x_{i,j_i-1}^{(P)},x_{i,j_i}^{(P)})$ in such a way that $g(\x)-g(\x')=(\rond g/\rond x_i)(x_{1j_1}^{(P)},\ldots, x_{i-1,j_{i-1}}^{(P)}, \theta_{ij_i}^{(P)},x_{i+1,j_{i+1}}^{(P)},\ldots,x_{nj_n}^{(P)})\Delta x_{ij_i}^{(P)}$. It follows that 
\[((\rond g/\rond x_i)(\x_{j_1,\ldots,j_n}^{(P)})-\epsilon)\Delta x_{ij_i}^{(P)}\le g(\x)-g(\x')\le ((\rond g/\rond x_i)(\x_{j_1,\ldots,j_n}^{(P)})+\epsilon)\Delta x_{ij_i}^{(P)},\]
which implies that 
\begin{align*}
	\mathrm{PFlux}_{\phi}^{(i)}(x_{1j_1}^{(P)},\ldots,x_{nj_n}^{(P)};\x_{j_1,\ldots,j_n}^{(P)})&\le\frac{\rond g}{\rond x_i}(\x_{j_1,\ldots,j_n}^{(P)})\phi^{(i)}(\x_{j_1,\ldots,j_n}^{(P)})\mathrm{Vol}(C(x_{1j_1}^{(P)},\ldots,x_{nj_n}^{(P)}))\\
	&+\epsilon\left|\phi^{(i)}(\x_{j_1,\ldots,j_n}^{(P)})\right|\mathrm{Vol}(C(x_{1j_1}^{(P)},\ldots,x_{nj_n}^{(P)})),\\
	\mathrm{PFlux}_{\phi}^{(i)}(x_{1j_1}^{(P)},\ldots,x_{nj_n}^{(P)};\x_{j_1,\ldots,j_n}^{(P)})&\ge\frac{\rond g}{\rond x_i}(\x_{j_1,\ldots,j_n}^{(P)})\phi^{(i)}(\x_{j_1,\ldots,j_n}^{(P)})\mathrm{Vol}(C(x_{1j_1}^{(P)},\ldots,x_{nj_n}^{(P)}))\\
	&-\epsilon\left|\phi^{(i)}(\x_{j_1,\ldots,j_n}^{(P)})\right|\mathrm{Vol}(C(x_{1j_1}^{(P)},\ldots,x_{nj_n}^{(P)})).
\end{align*}
Let $M_f$  be a bound for $f_{\X}^{2d}$. Now, set $M=M_f\int_{\Omega}\,\d\x$. Then, we have that
\small{	\begin{align*}
		\mathrm{PFlux}_{\phi}^{(i)}(x_{1j_1}^{(P)},\ldots,x_{nj_n}^{(P)};\x_{j_1,\ldots,j_n}^{(P)})&\le\frac{\rond g}{\rond x_i}(\x_{j_1,\ldots,j_n}^{(P)})\phi^{(i)}(\x_{j_1,\ldots,j_n}^{(P)})\mathrm{Vol}(C(x_{1j_1}^{(P)},\ldots,x_{nj_n}^{(P)}))+M\epsilon.
\end{align*}}
Similarly, we have that
\small{	\begin{align*}
		\mathrm{PFlux}_{\phi}^{(i)}(x_{1j_1}^{(P)},\ldots,x_{nj_n}^{(P)};\x_{j_1,\ldots,j_n}^{(P)})&\ge\frac{\rond g}{\rond x_i}(\x_{j_1,\ldots,j_n}^{(P)})\phi^{(i)}(\x_{j_1,\ldots,j_n}^{(P)})\mathrm{Vol}(C(x_{1j_1}^{(P)},\ldots,x_{nj_n}^{(P)}))-M\epsilon.
\end{align*}}
It follows that 
\begin{align*} \mathrm{PFlux}_{\phi}(x_{1j_1}^{(P)},\ldots,x_{nj_n}^{(P)};\x_{j_1,\ldots,j_n}^{(P)})&\le (\nabla g\cdot\phi)(\x_{j_1,\ldots,j_n}^{(P)})\mathrm{Vol}(C(x_{1j_1}^{(P)},\ldots,x_{nj_n}^{(P)}))+nM\epsilon,\\
	\mathrm{PFlux}_{\phi}(x_{1j_1}^{(P)},\ldots,x_{nj_n}^{(P)};\x_{j_1,\ldots,j_n}^{(P)})&\ge (\nabla g\cdot\phi)(\x_{j_1,\ldots,j_n}^{(P)})\mathrm{Vol}(C(x_{1j_1}^{(P)},\ldots,x_{nj_n}^{(P)}))-nM\epsilon.
\end{align*}
Now, assume that $ C(x_{1j_1}^{(P)},\ldots,x_{nj_n}^{(P)})\nsubseteq C_j $. Then, $\x_{j_1,\ldots,j_n}^{(P)}\notin \Supp(\phi)$. Otherwise, for a point $\mathbf{a}\in C(x_{1j_1}^{(P)},\ldots,x_{nj_n}^{(P)})\backslash C_j$, we have that
\[\mathrm{dist}(\Supp(\phi),\bd(C_j))\le\mathrm{dist}(\x_{j_1,\ldots,j_n}^{(P)},\mathrm{Bd}(C_j))\le|\x_{j_1,\ldots,j_n}^{(P)}-\mathbf{a}|\le\delta,\]
that contradicts with $\mathrm{dist}(\Supp(\phi),\bd(\Omega))>\delta$.
Thus, in this case, for each $1\le i\le n$, $\mathrm{PFlux}_{\phi}^{(i)}(x_{1j_1}^{(P)},\ldots,x_{nj_n}^{(P)})=0$,  which implies that $\mathrm{PFlux}_{\phi}(x_{1j_1}^{(P)},\ldots,x_{nj_n}^{(P)})=0$. Therefore, by taking a summation over all discrete-like $n$-cubes of $\mathcal{P}(C)$ and $\lVert P\rVert\to 0$, we have that
\[\int_{C}\nabla \widetilde{g}\cdot \widetilde{\phi}\,\d\x-nM\epsilon\le \PEACE_d^{\phi,\mathrm{dis}}(\X\to Y)\le \int_{C}\nabla \widetilde{g}\cdot \widetilde{\phi}\,\d\x+nM\epsilon.\]
Since, $\epsilon>0$ is arbitrary, we have that 
\[ \PEACE_d^{\phi,\mathrm{dis}}(\X\to Y)= \int_{C}\nabla \widetilde{g}\cdot \widetilde{\phi}\,\d\x=\int_{C_j}\nabla g\cdot \phi\,\d\x=\int_{\Omega}\nabla g\cdot \phi\,\d\x.\]
Now, by Equation \ref{multivariateequation}, 
\[ \PEACE_d^{\phi,\mathrm{dis}}(\X\to Y)= -\int_{\Omega} g\div(\phi)\,\d\x=\int_{\Omega} g\div(-\phi)\,\d\x.\]
Finally, the theorem is a consequence of Theorem \ref{maintheorem}.

\subsection*{Proof of Theorem \ref{NPEV}}
We prove the theorem only for $\bm{\epsilon}=\bm{+}$. The case 	 $\bm{\epsilon}=\bm{-}$ is similar. We note that it follows from the continuity of $(\rond g_{in}/\rond x)(\,\cdot\,,\z)$ and $r^{\bm{+}}=(|r|+r)/2$ that $\left((\rond g_{in}/\rond x)(\,\cdot\,,\z)\right)^{\bm{+}}$ is continuous. Since $f(\,\cdot\,|\mathbf{z})$ is continuous on $[a,b]$, it is uniformly continuous as well. Let us assume that $0<\epsilon<1$ is arbitrary. Then, there exists $\delta_1>0$ for which 
\[\forall\alpha,\beta\in[a,b]\;(|\alpha-\beta|<\delta_1\implies |f(\alpha|\mathbf{z})-f(\beta|\mathbf{z})|<\epsilon).\]
Now, assume that $\lVert P\rVert<\delta_1$. For any $1\le i\le n_P$, it follows from the mean value theorem that there exists $\alpha_i^{(P)}\in(x_i^{(P)}, x_{i-1}^{(P)})$ with \[g_{in}(x_{i}^{(P)},\mathbf{z})-g_{in}(x_{i-1}^{(P)},\mathbf{z})=\Delta x_i^{(P)}\frac{\partial{g_{in}}}{\partial{x}}(\alpha_i^{(P)},\mathbf{z}).\]
Also, for any $1\le i\le n_P$, we have that
\[\left|f(x_{i}^{(P)}|\mathbf{z})-f(\alpha_i^{(P)}|\mathbf{z})\right|<\epsilon,\quad \left|f(x_{i-1}^{(P)}|\mathbf{z})-f(\alpha_i^{(P)}|\mathbf{z})\right|<\epsilon.\]
It follows that
\begin{align*}
	L_{P,d}^{\mathbf{z}}(X\to Y)^{\bm{+}}&=\sum_{i=1}^{n_P}\Delta x_i^{(P)}\left(\frac{\partial{g_{in}}}{\partial{x}}(\alpha_i^{(P)},\mathbf{z})\right)^{\bm{+}}f(x_{i}^{(P)}|\mathbf{z})^df(x_{i-1}^{(P)}|\mathbf{z})^d\\
	&<\sum_{i=1}^{n_P}\Delta x_i^{(P)}\left(\frac{\partial{g_{in}}}{\partial{x}}(\alpha_i^{(P)},\mathbf{z})\right)^{\bm{+}}(f(\alpha_{i}^{(P)}|\mathbf{z})+\epsilon)^{2d}.
\end{align*}
By the mean value theorem for the function $u\mapsto u^{2d}$, for any $1\le i\le n_P$, there exists $0<\eta<\epsilon$ with 
\[(f(\alpha_{i}^{(P)}|\mathbf{z})+\epsilon)^{2d}-f(\alpha_{i}^{(P)}|\mathbf{z})^{2d}=2d\epsilon(f(\alpha_{i}^{(P)}|\mathbf{z})+\eta)^{2d-1}.\]
Now, assume that $M_{f}$ is an upper bound for $f(\cdot|\mathbf{z})$. Then, for any $1\le i\le n_P$, 
\[(f(\alpha_{i}^{(P)}|\mathbf{z})+\epsilon)^{2d}<f(\alpha_{i}^{(P)}|\mathbf{z})^{2d}+2d\epsilon(M_{f}+1)^{2d-1}.\]
Thus,  we have that
\begin{align*}
	L_{P,d}^{\mathbf{z}}(X\to Y)^{\bm{+}}&	<\sum_{i=1}^{n_P}\Delta x_i^{(P)}\left(\frac{\partial{g_{in}}}{\partial{x}}(\alpha_i^{(P)},\mathbf{z})\right)^{\bm{+}}f^2(\alpha_i^{(P)}|\mathbf{z})\\
	&+2d\epsilon(M_{f}+1)^{2d-1}\sum_{i=1}^{n_P}\Delta x_i^{(P)}\left(\frac{\partial{g_{in}}}{\partial{x}}(\alpha_i^{(P)},\mathbf{z})\right)^{\bm{+}}.
\end{align*}
Now, if $M_{g'}^{\mathbf{z}}$ is also an upper bound for $\left|\partial g_{in}/\partial{x}(\cdot,\mathbf{z})\right|$, then we have that
\[\sum_{i=1}^{n_P}\Delta x_i^{(P)}\left(\frac{\partial{g_{in}}}{\partial{x}}(\alpha_i^{(P)},\mathbf{z})\right)^{\bm{+}}<M_{g'}^{\mathbf{z}}\sum_{i=1}^{n_P}\Delta x_i^{(P)}=M_{g'}^{\mathbf{z}}(b-a),\]
which implies that
\[L_{P,d}^{\mathbf{z}}(X\to Y)^{\bm{+}}	<\sum_{i=1}^{n_P}\Delta x_i^{(P)}\left(\frac{\partial{g_{in}}}{\partial{x}}(\alpha_i^{(P)},\mathbf{z})\right)^{\bm{+}}f^2(\alpha_i^{(P)}|\mathbf{z})+C_1\epsilon,\]
where $C_1=2d(b-a)M_{g'}^{\mathbf{z}}(M_{f}+1)^{2d-1}$. 
Similarly, we have that
\[L_{P,d}^{\mathbf{z}}(X\to Y)^{\bm{+}}>-\epsilon C_2+\sum_{i=1}^{n_P}\Delta x_i^{(P)}\left(\frac{\partial{g_{in}}}{\partial{x}}(\alpha_i^{(P)},\mathbf{z})\right)^{\bm{+}}f^2(\alpha_i^{(P)}|\mathbf{z}),\]
where $C_2$ is a non-negative constant. Let $C$ be the maximum of $C_1$ and $C_2$. Then, we have that
\[\left|L_{P,d}^{\mathbf{z}}(X\to Y)^{\bm{+}}-\sum_{i=1}^{n_P}\Delta x_i^{(P)}\left(\frac{\partial{g_{in}}}{\partial{x}}(\alpha_i^{(P)},\mathbf{z})\right)^{\bm{+}}f^2(\alpha_i^{(P)}|\mathbf{z})\right|<\epsilon C.\]
We note that $\left(\partial{g_{in}}/\partial{x}(\cdot,\mathbf{z})\right)^{\bm{+}}f^2(\cdot|\mathbf{z})$ is a continuous function on $[a,b]$, and hence it is Riemann integrable. Thus, there exists $\delta_2>0$ such that for any $P\in\mathcal{P}$ with $\lVert P\rVert<\delta_2$, we have that
\[\left|\int_a^b\left(\frac{\partial{g_{in}}}{\partial{x}}(t,\mathbf{z})\right)^{\bm{+}}f^2(t|\mathbf{z})\,\mathrm{d}t-\sum_{i=1}^{n_P}\Delta x_i^{(P)}\left(\frac{\partial{g_{in}}}{\partial{x}}(\alpha_i^{(P)},\mathbf{z})\right)^{\bm{+}}f^2(\alpha_i^{(P)}|\mathbf{z})\right|<\epsilon.\]
Now, set $\delta$ to be the minimum of $\delta_1$ and $\delta_2$. It follows that for any $P\in\mathcal{P}$ with $\lVert P\rVert<\delta$, if 
\[L=\left|L_{P,d}^{\mathbf{z}}(X\to Y)-\int_a^b\left(\frac{\partial{g_{in}}}{\partial{x}}(t,\mathbf{z})\right)^{\bm{+}}f^2(t|\mathbf{z})\,\mathrm{d}t\right|,\]
then we have that
\begin{align*}
	L&\le \left|L_{P,d}^{\mathbf{z}}(X\to Y)^{\bm{+}}-\sum_{i=1}^{n_P}\Delta x_i^{(P)}\left(\frac{\partial{g_{in}}}{\partial{x}}(\alpha_i^{(P)},\mathbf{z})\right)^{\bm{+}}f^2(\alpha_i^{(P)}|\mathbf{z})\right|\\
	&+\left|\int_a^b\left(\frac{\partial{g_{in}}}{\partial{x}}(t,\mathbf{z})\right)^{\bm{+}}f^2(t|\mathbf{z})\,\mathrm{d}t-\sum_{i=1}^{n_P}\Delta x_i^{(P)}\left(\frac{\partial{g_{in}}}{\partial{x}}(\alpha_i^{(P)},\mathbf{z})\right)^{\bm{+}}f^2(\alpha_i^{(P)}|\mathbf{z})\right|\\
	&< \epsilon C+\epsilon =\epsilon(C+1).
\end{align*}
Therefore, it follows from the arbitrariness of $\epsilon>0$ that 
\[\PIEV_d^{\mathbf{z}}(X\to Y)=\lim_{\lVert P\rVert\to 0}L_P(X\to Y)^{\bm{+}}=\int_a^b\left(\frac{\partial{g_{in}}}{\partial{x}}(t,\mathbf{z})\right)^{\bm{+}}f^2(t|\mathbf{z})\,\mathrm{d}t.\]

\end{document}